\documentclass[12pt, letterpaper]{article}

\usepackage[utf8]{inputenc}
\usepackage[T1]{fontenc}
\usepackage[margin=1in]{geometry}
\usepackage{newtxtext}
\usepackage{microtype} % Better typography
\usepackage{amsmath}
\usepackage{graphicx}
\usepackage{booktabs} % Professional tables
\usepackage{multirow}
\usepackage{xcolor}
\usepackage{titlesec}
\usepackage{caption}
\usepackage{enumitem}
\usepackage{authblk}
\usepackage{natbib}

\usepackage{minitoc}
\usepackage{titletoc}
\usepackage{threeparttable}
\usepackage{cancel}
\usepackage{wrapfig}
\usepackage{rotating}

\usepackage{amsthm}%
\usepackage{mathrsfs}%
\usepackage[title]{appendix}%
\usepackage{textcomp}%
\usepackage{manyfoot}%
\usepackage{algpseudocode}%
\usepackage{listings}%
\usepackage{lineno}
\usepackage{etoolbox}
\usepackage[ruled, vlined, linesnumbered]{algorithm2e}

\usepackage[labelfont=bf, font=small]{caption}
\usepackage[textsize=tiny, color=blue!40!green!50!white, textwidth=3cm]{todonotes}

\titlespacing*{\section}{0pt}{2ex plus 1ex minus .2ex}{1ex plus 1ex minus .2ex}

\theoremstyle{thmstyleone}%
\theoremstyle{thmstyletwo}%

\theoremstyle{thmstylethree}%
\algrenewcommand\algorithmicensure{\textbf{Output:}}

\newcommand\blfootnote[1]{%
  \begingroup
  \renewcommand\thefootnote{}\footnote{#1}%
  \addtocounter{footnote}{-1}%
  \endgroup
}

\definecolor{linkblue}{RGB}{80, 100, 255}

\usepackage{hyperref}
\hypersetup{
    colorlinks=true,
    linkcolor=linkblue,
    filecolor=linkblue,      
    urlcolor=linkblue,
    citecolor=linkblue,
}
\usepackage{cleveref}

\usepackage{amsmath,amsfonts,bm}

\def\eqref#1{equation~\ref{#1}}
\def\1{\bm{1}}

\def\eps{{\epsilon}}

\def\rvf{{\mathbf{f}}}

\def\rvp{{\mathbf{p}}}

\def\rvv{{\mathbf{v}}}
\def\rvw{{\mathbf{w}}}
\def\rvx{{\mathbf{x}}}
\def\rvy{{\mathbf{y}}}
\def\rvz{{\mathbf{z}}}

\DeclareMathAlphabet{\mathsfit}{\encodingdefault}{\sfdefault}{m}{sl}
\SetMathAlphabet{\mathsfit}{bold}{\encodingdefault}{\sfdefault}{bx}{n}

\DeclareMathOperator*{\argmin}{arg\,min}

\Crefformat{figure}{Fig.~(#2#1#3)}
\crefformat{figure}{Fig.~(#2#1#3)}
\Crefformat{section}{Sec.~(#2#1#3)}
\crefformat{section}{Sec.~(#2#1#3)}
\Crefformat{equation}{Eq.~(#2#1#3)}
\crefformat{equation}{Eq.~(#2#1#3)}
\Crefformat{table}{Tab.~(#2#1#3)}
\crefformat{table}{Tab.~(#2#1#3)}
\Crefformat{appendix}{Appx.~(#2#1#3)}
\crefformat{appendix}{Appx.~(#2#1#3)}
\Crefformat{algorithm}{Alg.~(#2#1#3)}
\crefformat{algorithm}{Alg.~(#2#1#3)}

\titleformat{\section}
  {\Large\bfseries}{\thesection}{1em}{}
\titleformat{\subsection}
  {\large\bfseries}{\thesubsection}{1em}{}
\titleformat{\subsubsection}
  {\normalsize\bfseries}{\thesubsubsection}{1em}{}

\renewenvironment{abstract}{%
  \vspace{-1em}
  \begin{center}%
    {\bfseries\large Abstract}%
  \end{center}%
  \par\noindent\ignorespaces
}{\vspace{2em}}

\title{\vspace{-0.5em}\textbf{\huge Memorization to Generalization: \\ Emergence of Diffusion Models from Associative Memory}
}

\author[1]{Bao Pham$^*$}

\author[2]{Gabriel Raya$^*$}

\author[3]{Matteo Negri}

\author[1]{Mohammed J. Zaki}

\author[4]{Luca Ambrogioni}

\author[5]{Dmitry Krotov}

\affil[1]{\text{Department of Computer Science, Rensselaer Polytechnic Institute (\texttt{RPI})}}

\affil[2]{\text{Jheronimus Academy of Data Science, Tilburg University}}

\affil[3]{\text{Department of Physics, University of Rome Sapienza}}

\affil[4]{\text{Donders Institute for Brain, Cognition, and Behaviour, Radboud University}}

\affil[5]{IBM Research}
\date{}

\begin{document}

\maketitle
\begin{abstract}
Dense Associative Memories (DenseAMs) are generalizations of Hopfield networks, which have superior information storage capacity and can store training data points (memories) at local minima of the energy landscape. When the amount of training data exceeds the critical memory storage capacity of these models, new local minima, which are different from the training data, emerge. In Associative Memory these emergent local minima are called \textbf{\textit{spurious states}}, which hinder memory retrieval. In this work, we examine diffusion models (DMs) through the DenseAM lens, viewing their generative process as an attempt of a memory retrieval. In the small data regimes, DMs create distinct attractors for each training sample, akin to DenseAMs below the critical memory storage. As the training data size increases, they transition from memorization to generalization. We identify a critical intermediate phase, predicted by DenseAM theory -- the spurious states. In generative modeling, these states are no longer negative artifacts but rather are the first signs of generative capabilities. We characterize the basins of attraction, energy landscape curvature, and computational properties of these previously overlooked states. Their existence is demonstrated across a wide range of architectures and datasets.
\blfootnote{Corresponding email: \hyperlink{phamb@rpi.edu}{phamb@rpi.edu}. The code is available from this \hyperlink{https://github.com/Lemon-cmd/Diffusion-Models-and-Associative-Memory/tree/main}{link}.}
\blfootnote{$^*$ denotes equal contribution.}
\end{abstract}

%\keywords{Dense Associative Memory, Hopfield networks, Diffusion models, and explainability.}

\clearpage 

\section{Introduction} \label{sec:intro}

Hopfield networks are energy-based Associative Memory (AM)  models, which conceptualize memories as \textit{attractor states} corresponding to local minima of their energy function, and the memory retrieval as dynamical convergence towards such attractors \cite{hopfield1982, hopfield1984neurons, amari1972learning}. Modern variants of such networks, called Dense Associative Memories (DenseAMs) \cite{DenseAssociative,krotov2018dense}, have helped revitalize  scientific interest in these ideas and paved the way for more sophisticated AM systems \cite{Demircigil_2017, agliari2020neural, agliari2020tolerance, albanese2022replica, millidge2022universal, sahaCLAM, krotov2023new, et}, driven primarily by their connection to the attention mechanism in Transformers \cite{ramsauer2021hopfield, et} and neurobiology \cite{krotovlarge, kozachkov2025neuron}.

Simultaneously, diffusion models (DMs) \cite{sohl2015deep} have gained popularity, due to their flexibility and accuracy in modeling various high-dimensional distributions \cite{ho2020denoising, song2019generative, song2021scorebased, rombach2022high}. However, despite their effectiveness, DMs pose challenges related to privacy and security, as concerns grow about their tendency to replicate training data alongside a lack of understanding of how generalization arises in these models \cite{somepalli2023diffusion, somepalli2023understanding, carlini2023extracting, wen2024detecting, jeon2024understanding, webster2023reproducible}. Hence, such matters emphasize the need for further understanding of memorization and generalization behaviors of DMs.

Currently, much of the recent works on the memorization-to-generalization transition of DMs tackle it in a generalization-centric fashion \cite{meehan2020non, burg2021on, somepalli2023diffusion, somepalli2023understanding, kadkhodaie2023generalization, ventura2024manifolds, achilli2024losing, kamb2024analytic, ross2024geometric, biroli2024dynamical,  cui2025precise,   achilli2025memorization}. Namely, the memorization phenomenon is typically viewed as a ``small side effect,'' alleviated by drastically increasing the training data size \cite{yoon2023diffusion, gu2023memorization}. Although these studies shed some light on memorization and generalization, they do not fully explore the \textit{intermediate transition regime} between these two phases. Our work adopts a complimentary approach which further ties memorization and generalization together. From the beginning, we cast the diffusion modeling pipeline into the AM framework. The training phase of diffusion modeling is conceptualized as the operation of {\em writing the training data into the memory}. The generation phase is viewed as {\em the attempt of memory recall}. This recall can be successful, resulting in the retrieval of training samples (memorization), or unsuccessful, resulting in the generation of new previously unseen samples (generalization). Consequently, this point of view allows us to apply the theory developed for AMs to the memorization-to-generalization transition in DMs. 

Recent works \cite{hoover2023memory, ambrogioni2023search, raya2024spontaneous} have also begun establishing theoretical connections between DenseAMs and DMs, illustrating that the score function of typical DMs can be interpreted as the gradient of a DenseAM's energy function. These prior works, however, do not discuss a cornerstone phenomenon of AMs -- \textit{the emergence of spurious states} -- which can further bridge both DenseAMs and DMs. Historically considered as detrimental to reliable memory recall \cite{hopfield1982, Hopfield1983UnlearningHA, hopfield1984neurons, AmitHopfield, AbuHopfieldCapacity}, spurious states were viewed as negative artifacts of AM networks. In contrast, for generative modelling these same spurious states serve as the first signatures of generalization capabilities, thus becoming a desirable emergent phenomenon \cite{kalaj2024random}.

In accord with previous literature, we empirically find that DMs undergo a transition from memorization to generalization as the number of training points increases on realistic datasets. Unlike previous works, at the onset of generalization we detect the emergence of spurious patterns, which have sizable basins of attraction, but do not correspond to memorized data points. No longer negative artifacts, spurious states are now observed to be positive indicators of early generalization in DMs and they may also exist even during the full generalization stage of DMs, including Stable Diffusion \cite{rombach2022high}. Lastly, we provide theoretical descriptions distinguishing these peculiar states from memorized and generalized patterns, in terms of energy landscapes. Our findings suggest a close parallel between two seemingly different systems, DMs and DenseAMs, studied by two disjoint communities. Specifically, the DenseAM's perspective on DMs can illuminate some of the computational properties of DMs at the boundary of the memorization-generalization transition. 

\begin{figure*}[!t]
    \vspace{-10mm}
    \centering
    \setlength{\abovecaptionskip}{10pt}
    \setlength{\belowcaptionskip}{0pt} 
    \includegraphics[keepaspectratio, width=1.0\textwidth, height=1\textheight]{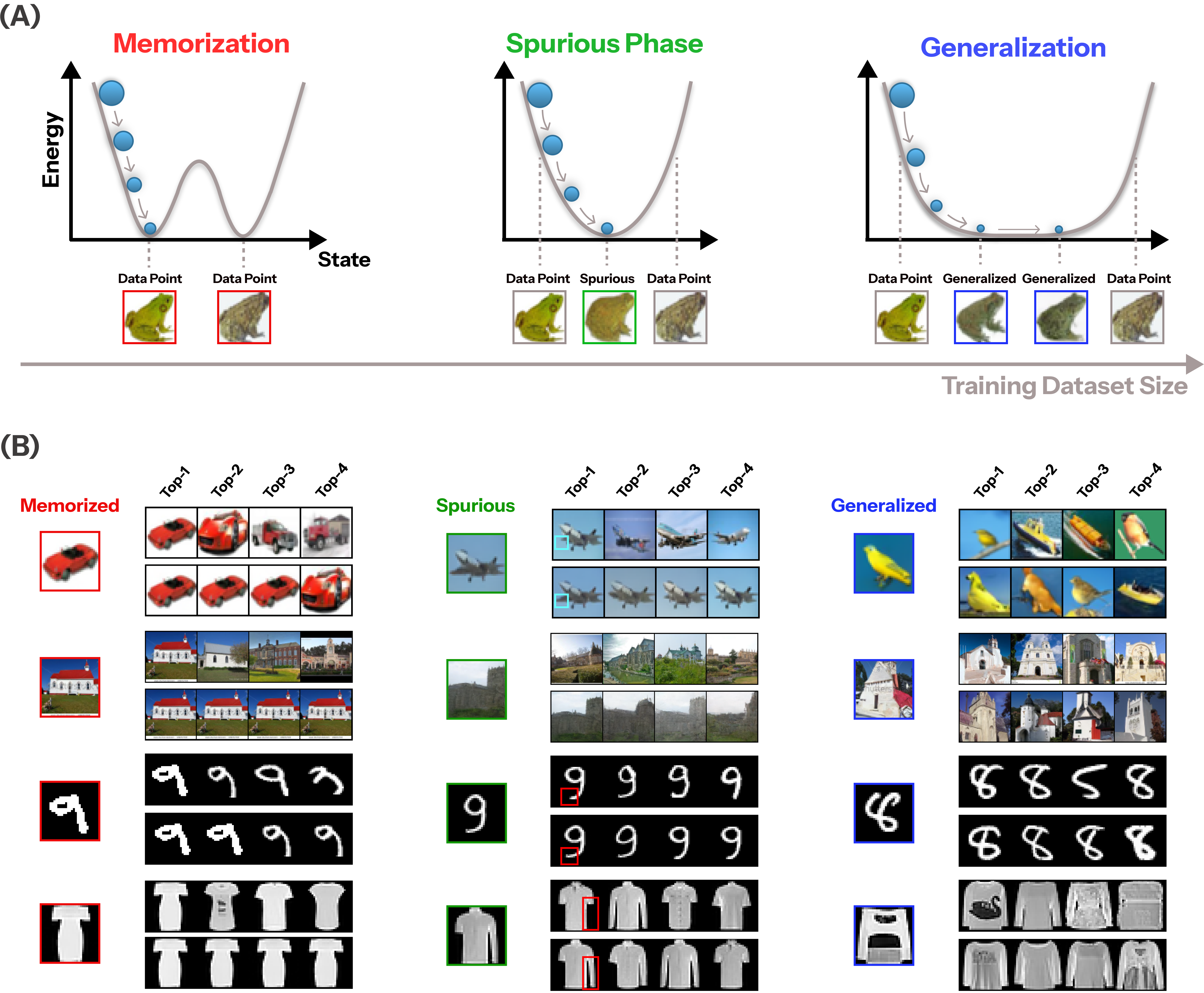}
    \caption{Panel \textsf(A) schematically illustrates the change in the energy landscape as the size of the training dataset is increased. In the small data regime, the model stores the training data points as local minima of the energy. When the amount of training data exceeds the model's memory capacity, spurious patterns are formed and training data points are no longer energy minima. Subsequent increase of the training data size leads to the generalization phase, defined by the formation of continuous manifold of the low energy states. Examples of \textcolor{red}{memorized}, \textcolor[rgb]{0, 0.6, 0}{spurious}, and \textcolor{blue}{generalized} samples in their respective columns for four datasets (MNIST, FASHION-MNIST, CIFAR10, and LSUN-CHURCH) are provided in Panel \textsf{(B)}, see Sec. (\ref{sec:transition}) for our definitions of these three sample types. For each target image (shown on the left), its {\textbf{top-4}} nearest neighbors from the training set {(\textbf{top row})} and the synthetic set {(\textbf{bottom row})} are shown to highlight the novelty and commonality of the target image with respect to its training and synthetic sets. To help highlight the novelty of spurious patterns, we provide rectangular markers guiding where the features differ in their corresponding nearest neighbors. 
    }
    \label{fig:general-energy-transition}
\end{figure*}
\newpage 

\section{Diffusion Models and Dense Associative Memories} \label{sec:results}
Motivated by \cite{hoover2023memory, ambrogioni2023search, raya2024spontaneous}, we first provide a theoretical link between DMs and DenseAMs through the perspective of energy. Given a dataset of samples $\rvy \in \mathbb{R}^N$ drawn from a target data distribution $\rvy \overset{\text{i.i.d}}{\sim} p(\rvy)$, we rely on the following stochastic differential equation (SDE) which describes the forward process \cite{song2021scorebased}: 
\begin{equation}
    \mathrm{d}\rvx_t = \mathbf{f}(\rvx_t, t) \mathrm{d}t + {g}(t) \mathrm{d} \rvw_t, 
     \label{eqn:forward_sde} 
\end{equation}
which transforms $p(\rvy)$, where $\rvx_{0} = \rvy$, into a simpler distribution, e.g., isotropic Gaussian. Here, $\rvw_t$ is the standard Wiener process and $\mathbf{f}(\rvx_t, t)$ denotes the drift term that guides the diffusion process, which we assume to be zero for most of the paper. Meanwhile, $g(t)$ denotes the diffusion coefficient controlling the noise at each time step $t \rightarrow T$. The reverse, denoising, or generative process is described as 
\begin{align}
    \mathrm{d} \rvx_t = [\mathbf{f}(\rvx_t, t) - g(t)^2 \nabla_{\rvx_t} \log p_t (\rvx_t)] \mathrm{d}t + g(t) \mathrm{d} \bar{\rvw}_t, 
    \label{eqn:backward_sde}
\end{align}
where $\bar{\rvw}_t$ is the standard Wiener process. To effectively solve this equation, one must reliably estimate the score $\nabla_{\rvx_t} \log p_t(\rvx_t) = -\nabla_{\rvx_t} E(\rvx_t, t)$ via training a neural network $s_\theta(\rvx_t, t)$, where $\theta$ denotes its set of trainable parameters, using methods for denoising score matching across multiple time steps \cite{ho2020denoising, song2021scorebased, hyvarinen05a, vincent2011connection}. 

Consider the training data distribution in the variance-exploding setting of $f(\rvx_t, t) = 0$ and $g(t) = \sigma$. In this case, the marginal probability distribution of noisy samples at time $t$ can be computed exactly as
\begin{equation}
    p(\rvx_t, t) = \underset{\rvy \sim \text{data}}{\mathbb{E}} \Bigg[\frac{1}{(2\pi\sigma^2 t)^{\frac{N}{2}}}\exp\Big(- \frac{\lVert \rvx_t - \mathbf{y} \rVert^2_2}{2 \sigma^2 t}\Big) \Bigg] .
\end{equation}
Assuming the empirical distribution of the data $p(\mathbf{y}) = \frac{1}{K}\sum\limits_{\mu=1}^K \delta^{(N)}(\mathbf{y} - \boldsymbol{\xi}^\mu)$, where $\boldsymbol{\xi}^\mu$ represents an individual data point (belonging to a dataset with training data size $K$), this marginal distribution can be written as 
\begin{equation}
    p(\rvx_t, t) = \frac{1}{K} \sum \limits_{\mu=1}^K \frac{1}{(2\pi\sigma^2 t)^{\frac{N}{2}}}  \exp\Big({ - \frac{\lVert \rvx_t - \boldsymbol{\xi}^\mu \rVert^2_2}{2 \sigma^2 t}}\Big)
    \overset{\text{def}}{\equiv} \exp\Bigg (- {\frac{E^\text{DM}(\rvx_t, t)}{2 \sigma^2 t}}\Bigg ),
    \label{eq:marginal}
\end{equation}
where we also defined the energy $E^\mathrm{DM}$ of the DM, up to $\rvx$--independent terms, to be 
\begin{equation}
    E^\text{DM}(\rvx_t, t) = -2 \sigma^2 t \log\bigg[\sum\limits_{\mu=1}^K \exp \Big(- \frac{\lVert \rvx_t - \boldsymbol{\xi}^\mu \rVert^2_2}{2 \sigma^2 t}\Big) \bigg] .
    \label{eq:energy-diffusion}
\end{equation}
As already observed in \cite{ambrogioni2023search}, Eq.~(\ref{eq:energy-diffusion}) is closely related to a commonly studied DenseAM's model, see for example \cite{sahaCLAM}:
\begin{align}
    E^\text{AM}(\rvx) = -\beta^{-1} \log\bigg[\sum\limits_{\mu=1}^K \exp\Big(- \beta \lVert \rvx - \boldsymbol{\xi}^\mu \rVert^2_2\Big) \bigg].
    \label{eq:DenseAM}
\end{align}
A direct comparison of Eqs. (\ref{eq:energy-diffusion}) and (\ref{eq:DenseAM}) highlights a close connection between these two seemingly different frameworks. Specifically, the data points $\boldsymbol{\xi}^\mu$ in the DM framework play the role of memories in the AM formulation. The variance of the noise added during the forward process (\ref{eqn:forward_sde}) plays the role of the effective temperature $\beta^{-1}$ (which controls the separation of memories), while the reverse process (\ref{eqn:backward_sde}) corresponds to memory retrieval dynamics \cite{hoover2023memory,ambrogioni2023search}. 

For a single data point $\boldsymbol{\xi}^1$ or $K = 1$, Eq. (\ref{eq:DenseAM}) has a single local minimum at that stored pattern, where the shape of this energy landscape is independent of the value of $\beta$. This is an example of a memorized state. In contrast, for $K = 2$, there are several possibilities. The energy landscape has two local minima when $\beta\rightarrow\infty$ is large. But, for finite values of $\beta$, there exist configurations of patterns such that Eq. (\ref{eq:DenseAM}) has only one local minimum:
\begin{equation}
    \boldsymbol{\eta} = \argmin_{\rvx} E^\text{AM}(\rvx), 
\end{equation}
such that $\boldsymbol{\eta}\neq\boldsymbol{\xi}^1$ and $\boldsymbol{\eta}\neq\boldsymbol{\xi}^2$. This is an example of a \textit{spurious state}. Such states are ubiquitous in AM models, when the amount of training data (memories) crosses the critical memory storage capacity of the model (which depends on the correlations between the training data points). Thus, the following question can be posed: \textit{Is it possible to observe spurious states in DMs trained using score matching objective when the energy landscape is modeled by the score-predicting neural network?}

\section{Discussion}

We argue below that the answer to the above question is positive, and the spurious states indeed emerge, as DMs progress from memorization to generalization as the training data size $K$ increases. As we observed numerically in Fig.~(\ref{fig:transition-plots}A), when $K$ is small, DMs initially memorize their training data points and store them as local minima of their energy function. In the large data regime, a different phase appears where a sufficient increase of $K$ fosters the creation of new attractor states corresponding to the manifold of the generated samples. Spurious states appear at the boundary of this transition and correspond to emergent attractor states of the energy function, absent in the training set, but at the same time, having distinct basins of attraction around them. Subsequent increases of $K$ leads to a decrease of the number of memorized and spurious states, and the creation of generalized states, points that occupy the continuous manifold. This sequence of transitions is illustrated in Fig.~(\ref{fig:general-energy-transition}A).

In order to empirically establish this sequence of transitions, we have trained a set of conventional diffusion models on datasets of gradually increasing sizes. Each of the trained models is characterized by its training set $\mathsf{S}$, and a synthetic set of generated samples drawn from this model $\mathsf{S}'$ (the synthetic set is assumed to be very large). A sample $\hat{\rvx} \in \mathsf{S}'$ is memorized if it has duplicates in both training and synthetic sets. This is clear from the three universality classes shown in Fig.~(\ref{fig:general-energy-transition}A). Since memorized samples have a basin of attraction around them, the same sample should appear in the synthetic set more than once, and that generated sample should be a copy of the training data point. A sample $\hat{\rvx} \in \mathsf{S}'$ is spurious if it has duplicates in the synthetic set, but not in the training set. Again, this is clear from Fig.~(\ref{fig:general-energy-transition}A): spurious samples have a well-defined basin of attraction around them, but the bottom of that basin corresponds to a newly generated sample -- not the training example. Finally, the generalized sample $\hat{\rvx} \in \mathsf{S}'$ has no duplicates in either the training or synthetic sets. Generalized samples occupy flat manifolds in the configuration space. For this reason, every generation produces a slightly different sample resulting in absence of duplicates in $\mathsf{S}'$. Additional details pertaining to these detection metrics are given in \autoref{sec:methods}.

Examples of generated samples belonging to these three categories are shown in Fig.~(\ref{fig:general-energy-transition}B). For each sample, the top-4 nearest neighbors from the training set (top row) and synthetic set (bottom row) are shown. It is instructive to visually examine some of the samples classified as spurious. For instance, the first spurious sample in Fig.~(\ref{fig:general-energy-transition}B) (the model was trained on CIFAR-10 dataset \cite{cifar10}) is a plane that has a large wing (highlighted by the cyan frame), which is absent in the training set, but has an exact duplicate in the synthetic set. Another interesting example is shown in the 4-th row (the model was trained on FASHION-MNIST dataset \cite{fmnist}). There are many examples of shirts in the training set with either two sleeves, or no sleeves. The generated sample shows only one sleeve. This sample has exact duplicates in the synthetic set, indicating the existence of a well-defined basin of attraction around it, but does not appear in the training set. These examples illustrate the general phenomenon, which is predicted by AM perspective and is clearly observed empirically for conventional diffusion models trained with the score matching objective. 

The abundance of memorized, spurious, and generalized samples is quantified as a function of training data set size in Fig.~(\ref{fig:transition-plots}A). As training set gets larger, the amount of memorized samples decreases, while the amount of generalized samples increases. The spurious states appear as a peak at the boundary of this transition.

The distinct nature of these three different states is further substantiated by the hierarchy of the log-volume of their basins of attraction in Fig.~(\ref{fig:transition-plots}B) and their energy curvature spectra in Fig.~(\ref{fig:energy-curvature}A). Regarding the basin volume, memorized samples consistently exhibit the largest log-volume, followed by spurious samples, while generalized samples have near-zero volume. This is in accord with the intuitive expectation for the energy profiles in Fig.~(\ref{fig:general-energy-transition}A). Memorized and spurious samples should have substantial basins of attraction around them, while generalized samples should have almost vanishing basins, since they occupy an almost flat energy manifold. The energy curvature spectral analysis reveals a similar finding: as the training data size $K$ increases: the memorized and spurious samples tend to have higher curvature (indicated by their fewer near-zero singular values) compared to the generalized samples. 

Initially, our studies focused on relatively small scale datasets: MNIST \cite{mnist}, FASHION-MNIST \cite{fmnist}, CIFAR10 \cite{cifar10}, and LSUN-CHURCH \cite{lsun}. Focusing on these settings allowed us to train a dense grid of models with varying size of the training set and carefully examine the existence of duplicates in the training and synthetic sets. Such a careful analysis would be intractable for large-scale real world scenarios such as Stable Diffusion \cite{rombach2022high}: training these models requires substantial compute resources,  the training set is large, and the synthetic set needs to be even larger in order to reliably perform the analysis similar to what was described above. At the same time, the analysis based on the curvature of the energy landscape around generated samples is more feasible. 

Thus, we have identified several candidate samples from Stable Diffusion that fall into three categories: sharp energy curvature, intermediate energy curvature, and almost flat energy. Examples of these samples together with their singular values of the score function Jacobian are shown in Fig.~(\ref{fig:energy-curvature}B). For each sample a language prompt was given to the model and the generated samples are shown together with the training image corresponding to that prompt. High curvature samples (candidate memorized states) are exact duplicates of the training image. Intermediate curvature samples (candidate spurious states) are somewhat similar to the training image, but deviate from it in certain aspects (color and pattern of the rug, color of the curtains). Finally, small curvature samples (candidate generalized states) demonstrate a greater degree of variability and larger deviations from the training image. Qualitatively, these results align with our primary findings on the carefully studied small dataset models.

\begin{figure*}[!t]
    \vspace{-10mm}
    \centering  
    \setlength{\abovecaptionskip}{10pt}
    \setlength{\belowcaptionskip}{0pt} 
    \centering
    \includegraphics[width=1.05\linewidth]{nature-submission/image-files/Toy-Example.png} 
    \caption{
    Energy landscape evolution for the 2D toy model as training data size $K$ increases. Models trained at $K \in \{2, 9, 1000\}$, using the VE-SDE based diffusion pipeline from \cite{song2021scorebased}, with training data sampled from the unit circle (shown in white). Generated samples are shown alongside the learned score field or neural network $s_{\theta}(\rvx_t, t)$, aligned with the negative gradient of the energy (\ref{eq:energy-diffusion}). Hierarchical clustering identifies structure within the generations, with cluster centroid energies visualized by $\textcolor{cyan}{\boldsymbol{\times}}$ and numerical value. The right-most panel shows the exact solution as $K \rightarrow \infty$ derived in Eq.~(\ref{eqn:toy-model-energy}). As $K$ grows, the model initially memorizes individual data points, forming isolated basins. Around $K = 9$, \textit{spurious patterns}, distinct low-energy attractors not present in the data, emerge and signal the onset of generalization. At large $K$, the model enters a fully generalized regime, where low-energy states lie on a flat continuous manifold.}
    \label{fig:toy-example}
\end{figure*}

\section{Results} \label{sec:methods}

The main result of our work is the empirical demonstration that DMs trained with standard methods generate spurious samples at the onset of memorization-to-generalization transition. The existence of these samples is a cornerstone of energy-based AM models. By casting DMs' generation step into DenseAM setting, one can theoretically expect the emergence of these states at the transition boundary. We have characterized computational properties of memorized, spurious, and generalized samples by studying their abundance, volume of the basins of attraction, and curvature of the energy landscape as a function of the DM training set size. Spurious samples in DMs have well-defined basins of attraction around them and have not been discussed in prior literature.  

Our theoretical arguments pertain to the idealistic setting in which the neural network $s_\theta (\rvx_t, t)$ is capable of learning the true empirical distribution obtained by adding noise to the training data. This limiting case is often adopted in theoretical studies of DMs, see \cite{biroli2024dynamical, biroli2023generative, achilli2025memorization, vastola2025generalization}. From a practical perspective, this is achieved if $s_\theta (\rvx_t, t)$ is over-parametrized and trained for a long time. This simplification allows us to make a theoretical prediction that DMs must generate spurious states. In practical settings, the size of the score modeling neural network is finite. It is non-trivial that the theoretical prediction obtained in the regime of an infinitely large neural network can be extrapolated to the regime in which the size of $s_\theta (\rvx_t, t)$ is finite and the amount of training data is large. A possible perspective on our contribution is the empirical demonstration that this extrapolation is valid, and spurious states indeed exist in practically used settings. 

In the AM literature, specifically on Hopfield networks, the concept of spurious states is often entangled with the phenomenon of mixture states, which are linear combinations of a certain number of memory vectors. While in certain kinds of AM models these two notions can be closely related, it is important to emphasize that they are different concepts. Spurious states are defined by the existence of a basin of attraction around them. They may or may not be mixture states. The main message of our work is to establish that spurious states exist in DMs. We do not claim that these spurious states are necessarily mixture states. It is up to the model, defined by the score modeling neural network, how the spurious states are constructed from the training data. 

Careful characterization of the memorization-generalization transition and the computational aspects of spurious states that define the onset of this transition may offer future insights for guiding the development of methods that can mitigate DMs' memorization.

\subsection{Memorization-to-Generalization Transition}
\label{sec:transition}
\textbf{Toy Model.} An instructive example, which can reinforce the energy connection between Eqs.~(\ref{eq:energy-diffusion}) and (\ref{eq:DenseAM}), can be shown via a simple 2-dimensional toy model exhibiting many aspects of the memorization-to-generalization transition. Imagine that the training data lies on a unit circle, see Fig.~(\ref{fig:toy-example}). We are interested in exploring how the shape of $E^\text{DM}$ (\ref{eq:DenseAM}) changes as the number of training data points increases. Consider the case when the number of training data points is infinite or $K \rightarrow \infty$. Using Eq. (\ref{eq:DenseAM}), the energy of this toy model can be derived as 
\begin{equation}
\begin{split}
     E^\text{AM}(R, \phi) & =  R^2+1 - \frac{1}{\beta} \log\big[I_0(2\beta R)\big] \underset{\beta\rightarrow\infty}{\approx} (R - 1)^2,
\end{split}
    \label{eqn:toy-model-energy}
\end{equation}
where $R$ is the radius of the unit circle, $\phi$ is the polar angle, and $I_0(\cdot)$ is a modified Bessel function of the first kind. The dependence on $\phi$ in Eq.~(\ref{eqn:toy-model-energy}) completely disappears from the final result, since the local minima of the resulting energy function form a {\em continuous manifold}, described by a parabola centered around $R=1$ as the inverse temperature $\beta \rightarrow \infty$. This behavior describes the fully generalized phase of this toy model. Please refer to the Supplementary for Eq.~(\ref{eqn:toy-model-energy}) derivation.

To test the resemblance of $E^\text{DM}$ and $E^\text{AM}$, we trained a set of DMs in this precise setting. Each model is trained on a dataset of size $K$. The resulting energy landscapes are shown in Fig.~(\ref{fig:toy-example}). For a small data size of $K = 2$, the DM exhibits memorization. The local minima of the energy correspond to the training data points. Importantly, at $K=9$ we can observe the first signs of \textit{spurious states}. At this stage, the model begins to learn emergent (different from the training data) local minima of the energy. Subsequent increase of the size of the training set leads to fully generalized behavior, illustrated at $K=1000$. At this stage, all of the samples from the model live in close proximity of the exact data manifold. The right panel shows the analytical expression of the energy landscape defined by Eq.~(\ref{eqn:toy-model-energy}). Thus, the conventional diffusion modeling pipeline agrees very well with the theoretical prediction of the empirical energy. \\

\begin{figure}[!t]
    \vspace{-10mm}
    \centering
    \setlength{\abovecaptionskip}{10pt}
    \setlength{\belowcaptionskip}{0pt} 
    \includegraphics[keepaspectratio, width=1.01\textwidth, height=1\textheight]{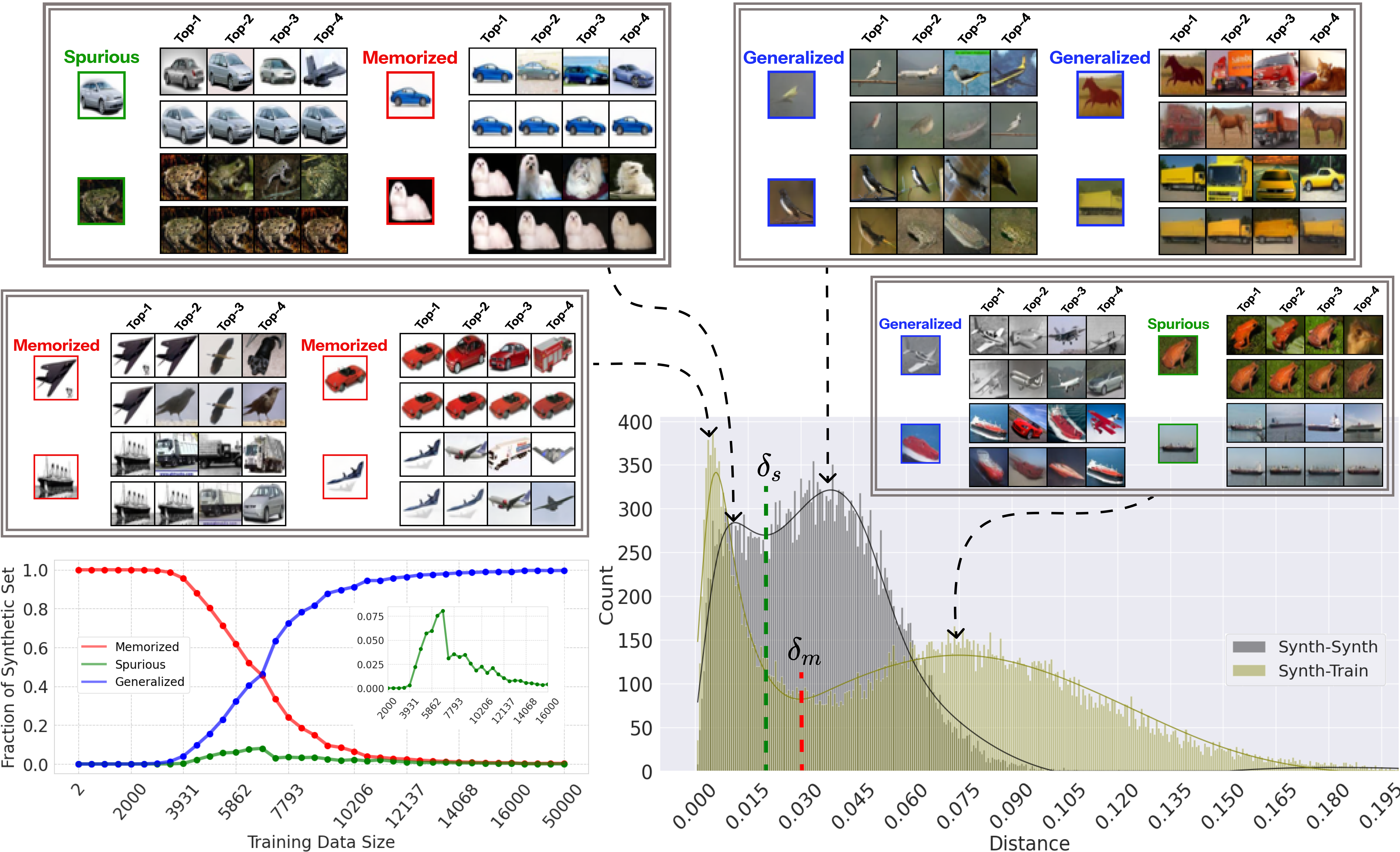}
    \caption{Different sample types across the memorization-to-generalization transition for CIFAR10. The \textcolor[gray]{0.4}{grey histogram} shows the distances between synthetic samples and their nearest neighbors from the synthetic set $\mathsf{S}'$. The threshold $\delta_s$ defines a boundary between the two peaks. The \textcolor{olive}{olive histogram} depicts the distances from the synthetic samples to their closest neighbor from the training set $\mathsf{S}$, with threshold $\delta_m$ separating the two peaks. \textcolor{red}{Memorized} samples are located in the left peak of the olive histogram, below $\delta_m$. In contrast, \textcolor{blue}{generalized} and \textcolor[rgb]{0, 0.6, 0}{spurious} samples appear to the right of $\delta_m$ in the olive histogram. Examples of the generated samples forming each of the four peaks of the histograms are shown in the inset frames. For each generated sample, {\textbf{top-4}} nearest neighbors from the training set are shown in the {\textbf{top row}}, and those from the synthetic set are shown in the {\textbf{bottom row}}. Training set size $K=7310$ is used in this figure, but the discussed phenomena are general and largely independent of this specific value. The fraction of the memorized, spurious, and generalized samples in the pool of all generated samples is shown in the bottom left panel as a function of the training set size. The inset shows amplified spurious fraction (\textcolor[rgb]{0, 0.6, 0}{green curve}). 
    }
    \label{fig:cifar10-hist}
\end{figure}

\noindent \textbf{Natural Datasets.} However, the main goal of our work is to also establish the existence of spurious states in DMs trained on natural high-dimensional datasets. With this goal in mind, we design detection metrics that can classify any generated sample into one of three categories -- \textit{memorized}, \textit{spurious}, and \textit{generalized} samples -- to aid in our analyses. Fundamentally, these metrics defined below rely on two datasets: the training dataset $\mathsf{S}$ used to train the DM, and the synthetic dataset $\mathsf{S}'$ generated from the fully trained DM. The size of $\mathsf{S}'$ is assumed to be much bigger than $\mathsf{S}$. 

Our \textbf{core intuition} is rooted in the energy landscape, where its topography dictates the nature of the generated samples, as shown in Fig.~(\ref{fig:general-energy-transition}A). Each sample type corresponds to a distinct feature of this landscape. Memorized samples originate from the basins of attraction the model creates around individual data points from the training set $\mathsf{S}$. Because these are strong (having large basins), the generation process frequently converges to exact memorized samples at small data sizes, producing outputs in the synthetic set $\mathsf{S}'$ that are perfect duplicates of training data. In contrast, spurious samples arise from new emergent basins of attraction, formed during training, which do not correspond to any specific sample in $\mathsf{S}$. These emergent minima are also stable attractors, which explains why they appear as duplicates within $\mathsf{S}'$ but are absent from $\mathsf{S}$. Finally, generalized samples are drawn from flatter regions of the energy landscape. Since these areas support a continuum of low-energy states rather than a single sharp minimum, the probability of landing on the exact same point twice is negligible, resulting in unique novel creations.

With this intuition in mind we examine two histograms, see Fig. (\ref{fig:cifar10-hist}). First, for every element of  $\mathsf{S}'$, their distance to their closest nearest neighbor from $\mathsf{S}$ is shown by the \textcolor{olive}{olive histogram}. Second, for every element of $\mathsf{S'}$, their distance to their closest nearest neighbor in $\mathsf{S}'$ (excluding itself) is shown by  the \textcolor[gray]{0.4}{gray histogram}. Both histograms exhibit a clear bimodal shape, indicating that at least \textit{two groups} of samples are present in $\mathsf{S}'$. This bimodality makes it possible to select the memorized  $\delta_m$ and spurious $\delta_s$ thresholds to separate the samples forming the two peaks in each histogram. The left peak of the olive histogram identifies memorized samples by their small distance to the training set $\mathsf{S}$, leaving spurious and generalized samples in the right peak. The gray histogram then distinguishes truly novel patterns: its left peak contains high-frequency duplicates (both memorized and spurious), while its right peak consists of unique generalized samples. This bimodality provides an empirical basis for defining our detection metrics below.

Following \cite{yoon2023diffusion}, we define the memorization detection metric $\mathcal{M}$. A sample $\hat{\rvx} \in \mathsf{S}'$ is considered memorized if its distance to its nearest neighbor $\rvx_1$, from the training set $\mathsf{S}$, falls below a threshold $\delta_m \in \mathbb{R}$:
\begin{equation}
    \mathcal{M}(\hat{\rvx}, \mathsf{S}) = \mathbb{I} \bigg{(} d(\hat{\rvx}, \rvx_1) \leq \delta_m \bigg{)} , 
    \label{eqn:mem-detection}
\end{equation}
where $\mathbb{I}$ represents the indicator function and $\delta_m$ is a threshold derived from the olive distance histograms, see Fig. (\ref{fig:cifar10-hist}). Second, a sample $\hat{\rvx} \in \mathsf{S}'$ is spurious if it is not a memorized state but is a duplicate of the synthetic set $\mathsf{S}'$. This is detected by the metric $\mathcal{S}$, which identifies non-memorized samples that are close to their respective nearest neighbor $\rvx'_1$ within the synthetic set $\mathsf{S}'$:
\begin{equation}
    \mathcal{S} (\hat{\rvx}, \mathsf{S}, \mathsf{S}') = \mathbb{I} \bigg{(} d(\hat{\rvx}, \rvx'_1) \leq \delta_s \bigg{)} \land \neg \mathcal{M}(\hat{\rvx}, \mathsf{S}), 
    \label{eqn:spur-detection}
\end{equation}
where $\delta_s \in \mathbb{R}$ is a threshold value chosen based on the gray distance histogram. Finally, a sample $\hat{\rvx}$ is generalized if it is neither memorized nor spurious:
\begin{equation}
    \mathcal{G}(\hat{\rvx}, \mathsf{S}, \mathsf{S}') = \neg \mathcal{M}(\hat{\rvx}, \mathsf{S}) \land \neg\mathcal{S}(\hat{\rvx}, \mathsf{S}, \mathsf{S}') . 
    \label{eqn:gen-detection}
\end{equation}

\begin{figure}[!t]
    \vspace{-10mm}
    \centering  
    \setlength{\abovecaptionskip}{10pt}
    \setlength{\belowcaptionskip}{0pt} 
    \includegraphics[keepaspectratio, width=1.01\textwidth, height=1\textheight]{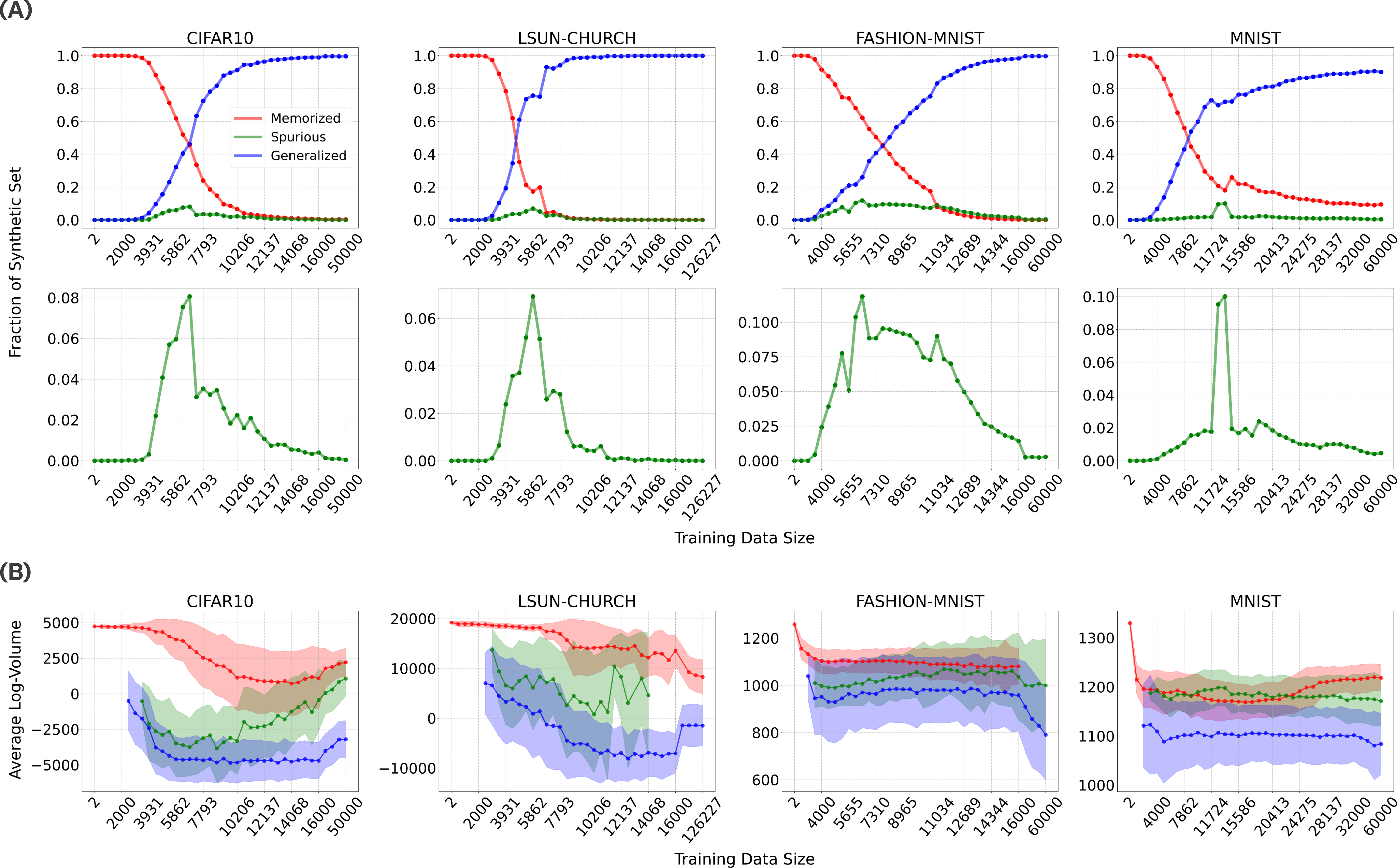}
    \caption{Fractions of \textcolor{red}{memorized}, \textcolor[rgb]{0, 0.6, 0}{spurious}, and \textcolor{blue}{generalized} samples in synthetic sets across training sizes and datasets. As the training data size $K$ increases, memorization decreases and the fraction of generalized samples steadily increases, see \textbf{top row} of Panel (\textsf{A}). The fraction of spurious patterns rises and decreases at the boundary between the memorization and generalization phases, see \textbf{bottom row} of Panel (\textsf{A}). Panel (\textsf{B}) illustrates the average log-volume of the basins of attraction for memorized and spurious samples being statistically larger than that of generalized samples across all datasets. The shaded regions indicate standard deviation of the log-volume.
}
\label{fig:transition-plots}
\end{figure}

With these metrics, we computed the fractions of memorized, spurious, and generalized samples in the synthetic sets for various training data sizes for different datasets, see Fig. (\ref{fig:transition-plots}A). These datasets include MNIST \cite{mnist}, FASHION-MNIST \cite{fmnist}, CIFAR10 \cite{cifar10}, and down-sampled LSUN-CHURCH \cite{lsun}. For each dataset, we trained DDPM-based DMs \cite{ho2020denoising} for $M=38$ different data sizes. For each model $\alpha=1,...,M$, trained on the training set $\mathsf{S}_\alpha$, a synthetic set $\mathsf{S}'_\alpha$ was generated. To account for duplication, we ensured that each $\mathsf{S}'_\alpha$ is four times the size of its corresponding $\mathsf{S}_\alpha$. We experimented with up to 8$\times$ ratio to the training set size and did not notice qualitative differences in our conclusions. Each sample $\hat{\rvx} \in \mathsf{S}'_\alpha$ is classified as either memorized, spurious, or generalized using Eqs.~(\ref{eqn:mem-detection})-(\ref{eqn:gen-detection}). The fractions of these three pattern types with respect to $|\mathsf{S}'_\alpha|$ are plotted in Fig.~(\ref{fig:transition-plots}). For each dataset, our smallest model was trained on the training set of $|\mathsf{S}_1|=2$ data points while the largest model was trained on the entire original training set $\mathsf{S}_M = \mathsf{S}$. 

Furthermore, motivated by \cite{biroli2024dynamical, raya2024spontaneous, li2024critical}, for these three distinct states we measure the critical time $t_c$, where the reverse process (\ref{eqn:backward_sde}) can still recover a target pattern $\hat{\rvx}$ from its perturbation $\rvx_{t_c}$ within an error margin. The time $t_c$ helps define the radius $R(\hat{\rvx}, \rvx_{t_c}) = \lVert \hat{\rvx} - \rvx_{t_c} \rVert$ of the basin of the attractions of a sample $\hat{\rvx}$, which is the Euclidean norm of the difference between $\hat{\rvx}$ and its recoverable perturbation $\rvx_{t_c}$. The volume of this basin is simply defined by the $N$-ball volume:
\begin{equation}
    V(\hat{\rvx}, \rvx_{t_c}) = \frac{\pi^\frac{N}{2}}{\Gamma(\frac{N}{2} + 1)} R(\hat{\rvx}, \rvx_{t_c})^N,
    \label{eqn:volume-hypersphere}
\end{equation}
where $\Gamma(\cdot)$ is the Gamma function. For numerical stability, we show the results of the average logarithmic of Eq.~(\ref{eqn:volume-hypersphere}) in Fig.~(\ref{fig:transition-plots}B) computed from samples belonging to each distinct state.

Confirming the phenomenon observed in our toy model and core intuition, our experiments on these datasets demonstrate a clear three-phase transition from memorization to generalization. As shown in Fig.~(\ref{fig:transition-plots}A), models trained in the small data regime predominantly replicate training samples. As the training data size $K$ surpasses a critical capacity, memorization sharply declines, and spurious samples emerge, signaling the onset of generalization. Further increase of $K$ diminishes the fraction of both memorized and spurious samples, and the model enters the full generalization regime, frequently producing unique novel patterns, see Fig.~(\ref{fig:general-energy-transition}B) for the visualization of these three distinct patterns. %This characteristic trend holds true not only as $K$ increases, but is also consistently observed when varying the model's network size while not affected by having much larger synthetic sets. 

Additionally, our results reveal a distinct hierarchy in the basins of attraction for the three sample types. Memorized samples consistently exhibit the largest log-volume, followed by spurious samples, while generalized samples have a much smaller volume, consistent with theoretical expectations. %Crucially, the basin volumes for all sample types systematically decrease as the training data size increases. This is an expected consequence of packing more data into the configuration space, forcing the model to shift its function from denoising toward more generative behaviors. 
Please refer to Supplementary for more details of this section and additional results.

\begin{figure*}[t]
    \vspace{-10mm}
    \centering
    \setlength{\abovecaptionskip}{10pt}
    \setlength{\belowcaptionskip}{0pt} 
    \includegraphics[keepaspectratio, width=1.01\textwidth, height=1\textheight]{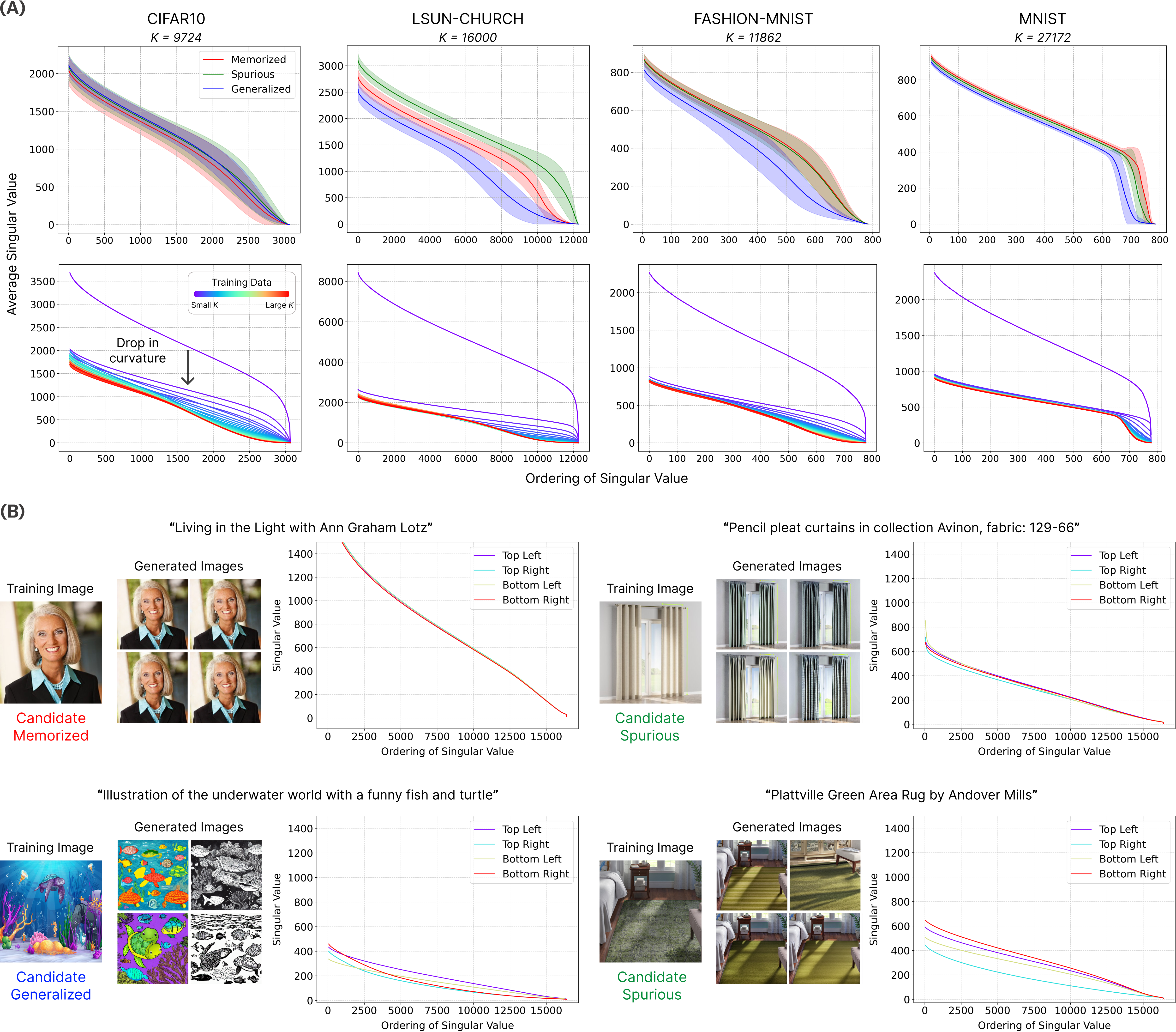}
    \caption{Panel \textsf{(A)} depicts the average singular values of the energy curvature for different sample types and datasets and the shaded region is the standard deviation of the singular values (a particular training set size $K$ is shown). Memorized and spurious samples generally exhibit higher energy curvature, characterized by fewer near-zero singular values than  generalized samples. Bottom row of \textsf{(A)} illustrates the average spectra computed for the training data points, where the changing color scheme denotes small to large data sizes, suggesting a drop in curvature (indicated by the decrease in singular values) as $K$ increases. Panel \textsf{(B)} shows candidate examples of memorized, generalized, and spurious samples from a Stable Diffusion model \cite{rombach2022high} trained on the LAION dataset \cite{schuhmann2022laion}, each corresponding to a distinct curvature signature. The candidate memorized sample has much larger singular values, while the candidate generalized sample has much smaller ones. The selected possible spurious samples have larger singular values than the candidate generalized sample, but smaller than the candidate memorized sample. The common trait between candidate memorized and spurious samples is that both are stable attractors, as demonstrated by their repeated similar generations given different initial points (or noise vectors) and the conditioning on text-prompt. The y-axis is clipped at the value of 1500 to better contrast the shown examples' spectra.
    }
    \label{fig:energy-curvature}
\end{figure*}

\clearpage 

\subsection{Energy Curvature}

While the changing fractions of memorized, spurious, and generalized samples constitute a good signature of the memorization-to-generalization transition, the underlying mechanism is also rooted in the geometry of the model's learned energy landscape, as shown in Figs.~(\ref{fig:general-energy-transition}A) and (\ref{fig:toy-example}). To elucidate this mechanism, we quantify the energy curvature at the minima corresponding to each sample type obtained at a small time $t_0$. 

To measure this curvature, we rely on the relationship $\nabla_{\rvx_t} \log p_t (\rvx_t) = - \nabla_{\rvx_t} E_t(\rvx_t)$ to analyze the Hessian of the energy function or the Jacobian of the score, evaluated at the location of each generated sample $\hat{\rvx} \in \mathbb{R}^N$. Specifically, following \cite{stanczuk2022your, achilli2024losing, ventura2024manifolds}, we numerically approximate the spectra of the energy Hessian by exploiting the tangent space of the model's learned manifold at a given pattern $\hat{\rvx}$. First, we generate a set of $4N$ perturbations around $\hat{\rvx}$ using a small time $t_0$ in Eq.~(\ref{eqn:forward_sde}). The learned score $s_\theta(\hat{\rvx}_{t_0}, t_0)$ is then evaluated at each perturbed point $\hat{\rvx}_{t_0}$ and the resulting $4N$ score vectors form a matrix $\mathsf{M} \in \mathbb{R}^{N \times 4N}$. Singular value decomposition is then performed on $\mathsf{M}$ to extract its singular values, serving as our measure of the local curvature of $\hat{\rvx}$ in all directions. The curvature spectra for the three sample types for different datasets are illustrated in the top row of Fig.~(\ref{fig:energy-curvature}A). Please refer to the Supplementary for further experimental details.

While a consistent separation in the curvature spectra of memorized, spurious, and generalized samples is not readily apparent for all training data sizes $K$, likely due to the geometric complexity of the transitional phases, we are still able to observe very well the separation of the curvature spectra between memorized and generalized samples at many sizes of the training dataset (prominently at or after the peak of spurious states). In addition, we repeated a similar analysis for training data points. There is a consistent decrease in the singular values measured at the locations of the training data points as their number $K$ increases, see bottom row of Fig. (\ref{fig:energy-curvature}A). This provides a direct geometric evidence of the mechanism behind generalization: the model transitions from creating sharp, high-curvature minima for individual data points in the memorization regime to learning a smoother, lower-curvature manifold that encompasses the entire dataset in the generalization regime. 

To validate these geometric findings on a large-scale model, we performed a case study using Stable Diffusion \cite{rombach2022high}, trained on the LAION dataset \cite{schuhmann2022laion}, shown in Fig.~(\ref{fig:energy-curvature}B). The analysis of handpicked candidate examples of memorized, spurious, and generalized samples from LAION reveals a similar hierarchy of energy curvatures. As expected, a heavily memorized sample occupies a high-curvature minimum, evidenced by large singular values. In contrast, a more generalized sample exhibits a significantly lower-curvature signature, while spurious samples occupy an intermediate geometric space. Notably, the spectra for all generated samples remain relatively sharp, with few near-zero singular values, as similarly found in \cite{jeon2024understanding}. We argue this is a consequence of text conditioning, which constrains the generative manifold, making Stable Diffusion more prone to operate in high-curvature modes, increasing the risk of memorization.

\newpage 
\subsection{Relative Energy Across the Transition}
\label{sec:relative-energy}
\begin{wrapfigure}{r}{0.50\textwidth}
\vskip -0.2in
\includegraphics[keepaspectratio, width=0.49\textwidth, height=1\textheight]{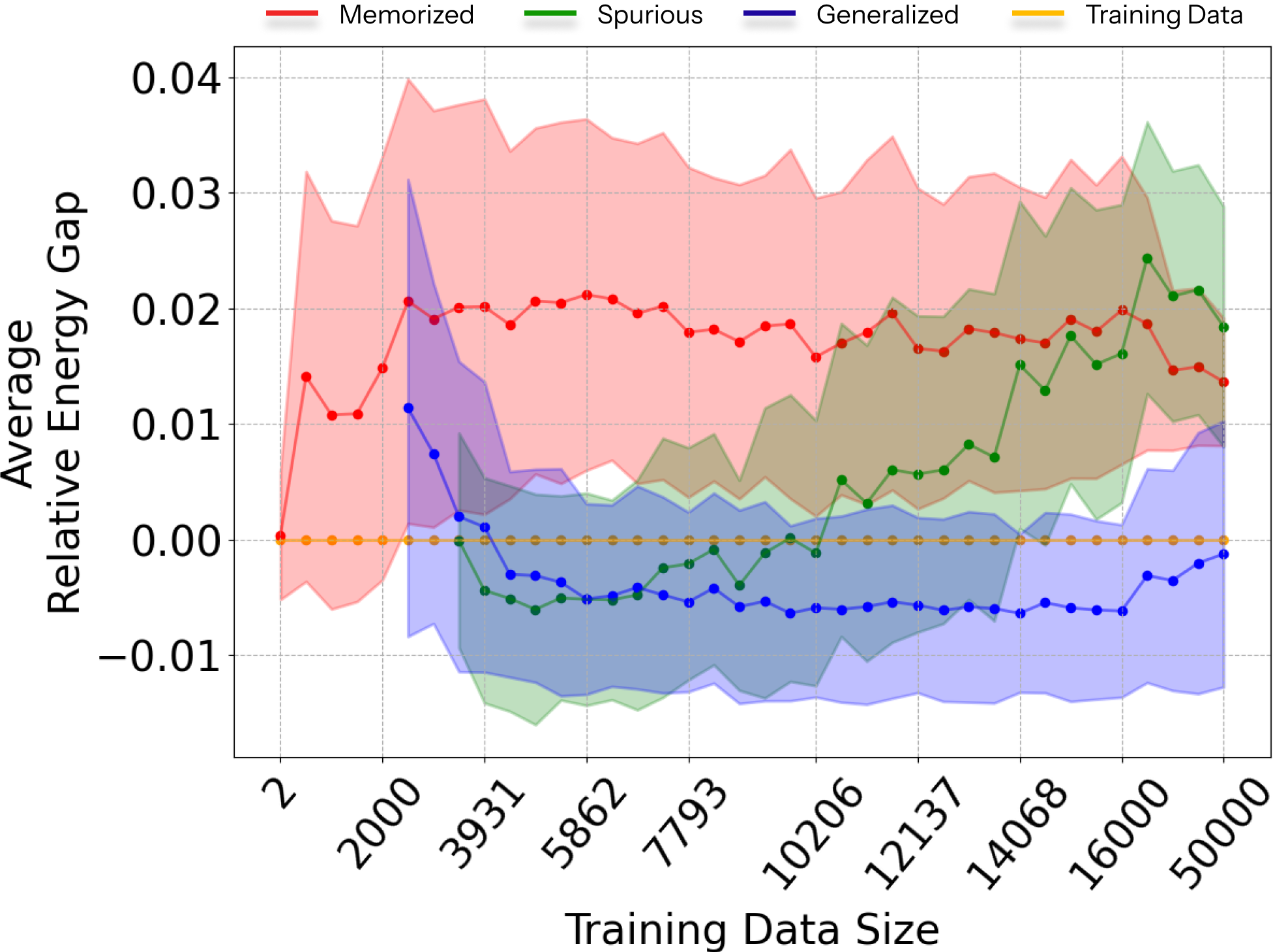}
\
\caption{
Average relative energy gap of each sample type and the training samples in CIFAR10 \cite{cifar10} as training data size $K$ grows. The gaps are measured relative to training samples, whose energy is set to zero (energy is defined up to an additive constant). Shaded regions show standard deviation of the gap values.
}
\label{fig:cifar10-rel-energy}
\end{wrapfigure}

Alongside our energy curvature analyses, we study how the energy of these three distinct sample types changes as the training data size increases. Following the efficient relative energy computation from \cite{raya2024spontaneous}, we measure the relative energy gap of the three sample types with respect to that of the training data samples. The results are recorded in \Cref{fig:cifar10-rel-energy}. Please refer to \Cref{sec:relative_energy_details} for full details on the experiment and computation of the relative energy gap for different sample types, alongside additional results.

For small training data sizes, the energy of the memorized samples is low and of the generalized samples is high. As $K$ increases, the energy of the memorized samples increases, while the energy of the generalized samples decreases. The energy of the spurious states is the lowest among these three distinct sample types during the memorization-generalization transition.

\section{Conclusion}

In conclusion, we established a novel connection between DMs and DenseAMs by identifying the emergence of creativity (generalization) in DMs with the failure of successful memory recalls via an energy-based associative memory. Using this theory, we empirically identified a novel intermediate phase, {\em spurious states}, previously overlooked in related DM literature, which marks the onset of generalization in these models as the training data size increases.  Additional analyses of the volume of the basins of attraction on samples generated from DMs and various aspects about their energy help validate our hypothesis delineated in Fig. (\ref{fig:general-energy-transition}). Our findings on Stable Diffusion suggest that these novel, stable attractors persist even in large models trained on massive datasets. Thus, our framework provides a new tool for understanding the computational mechanisms of DMs, opening new avenues for characterizing their behaviors.

\section*{Acknowledgments}
The results presented here were obtained while Dmitry Krotov was employed by IBM Research. At the time of the submission Dmitry Krotov is no longer employed by IBM Research. Matteo Negri acknowledges the support of PNRR MUR project PE0000013-FAIR. Bao Pham and Mohammed Zaki acknowledge the support of RPI-IBM Future of Computing Research Collaboration program (FCRC).

\newpage

\bibliography{nature-submission/sn-bibliography}
\bibliographystyle{unsrt}

\clearpage 
\appendix
\startlist{toc}
\printlist{toc}{}{\section*{\Large \textbf{Appendix}}}
\clearpage 

\section{Toy Example}
\subsection{Overview}
\label{sec:toy-model-energy-details}
For the 2D toy example, our training data points lie on a unit circle. In the case of infinite training data size or $K \rightarrow \infty$, the distribution of these data points can be described by a continuous density of states: 
\begin{equation}
    p(\rvy) = \frac{1}{\pi} \delta\big(y_1^2+y_2^2-1\big) .
    \label{2D density}
\end{equation}
The probability of the generated data is proportional (up to terms independent of the state $\rvx$) to 
\begin{equation}
    p(\rvx) \sim \int\limits_{-\infty}^{+\infty} dy_1 dy_2\ p(\rvy)\ e^{- \beta \lVert \rvx - \rvy \rVert^2_2} = e^{-\beta (R^2+1)} I_0(2\beta R)  ,
    \label{eq: continous circle}
\end{equation}
where $I_0(\cdot)$ is a modified Bessel function of the first kind. Thus, the energy of the model is given by 
\begin{equation}
\begin{split}
     E^\text{AM}(R, \phi) & =  R^2+1 - \frac{1}{\beta} \log\big[I_0(2\beta R)\big] \underset{\beta\rightarrow\infty}{\approx} (R - 1)^2,
\end{split}
    \label{eqn:toy-model-energy}
\end{equation}
where $R$ is the radius of the unit circle and $\phi$ is the polar angle. The dependence on the polar angle $\phi$ in Eq. (\ref{eqn:toy-model-energy}) completely disappears from the final result, since the local minima of the resulting energy function form a {\em continuous manifold}, described by a parabola centered around $R=1$ as $\beta \rightarrow \infty$. This behavior describes the fully generalized phase. 

In order to obtain Eq. (\ref{eq: continous circle}), it is easiest to introduce polar coordinates for both the state vector $\rvx$ and the training data $\rvy$: $$\begin{cases}
    x_1 = R \cos(\phi)\\
    x_2 = R \sin(\phi)
\end{cases} \ \ \ \ \ \ \ \ \ \ \begin{cases}
    y_1 = r \cos(\varphi)\\
    y_2 = r \sin(\varphi)
\end{cases}$$ 
The integral (\ref{eq: continous circle}) can then be written as $$p(\rvx) \sim \int\limits_{0}^{2\pi} d\varphi\int\limits_0^\infty r dr \frac{1}{\pi} \delta(r^2-1) e^{-\beta[R^2+r^2-2Rr\cos(\varphi-\phi)]}
= e^{-\beta (R^2+1)} I_0(2\beta R)
$$ and explicitly computed using the definition of the modified Bessel functions \cite{gradshteyn2014table}.

\subsection{Experimental Details}
\label{sec:toy-model-experiment}

\noindent \textbf{Dataset}. To construct the dataset, we uniformly sampled points on the unit circle. Concretely, we had 
$\rvy \sim p(\rvy) = \frac{1}{2\pi}\delta(r-1)$, where $r = \sqrt{y_1^2 + y_2^2}$ is the radius and $\delta(r - 1)$ is the delta function ensures that all probability mass lies on radius $r = 1$. The factor $\frac{1}{2\pi}$ guarantees that the angles are sampled uniformly, relying on the identity:
\begin{equation}
    \delta(\sqrt{y_1^2 + y_2^2}-1) = 2\delta(y_1^2 + y_2^2-1),
\end{equation}
where $p(\rvy) = \frac{1}{\pi} \delta(y^2_1 + y_2^2 - 1)$. In practice, we sampled polar coordinates $(r, \varphi)$ with $r = 1$ and uniformly sampled the angular coordinate $\varphi \in [0, 2\pi]$. We then converted $(r, \varphi)$ to Cartesian coordinates ($y_1, y_2$) using $y_1 =r \cos \varphi$ and $y_2=r \sin \varphi$.

Then, We created a data set of $60000$ points using a fixed random seed, then built smaller subsets by progressively adding distinct samples without replacement, ensuring that each subset is a strict subset of the next. This approach allowed us to systematically examine how the model behaves as we vary the size of the training set. \\

\noindent \textbf{Exact Score}. Following Eq. (\ref{eqn:toy-model-energy}), the energy can be written in closed form, whose local minima form the continuous data manifold at $R = 1$\footnote{Please note throughout this paper we use $r$ for the polar radius of the training data and $R$ for the polar radius of the generated samples.}. Furthermore, in the case of many data points or $K \rightarrow \infty$, we have $p(\rvy) = \frac{1}{\pi}\delta(y_1^2 + y_2^2-1)$. Since $p(R) \propto \exp^{-E^\text{AM}(R)}$, the score can be described as 
\begin{equation}
    \begin{split}
        -\nabla_R \log p(R) &= \nabla_R E^\text{AM}(R) \\
        & = 2R - \frac{2\beta I_1(2\beta R)}{\beta I_0(2\beta R)} \\
        &=  2R - 2\frac{I_1(2\beta R)}{I_0(2\beta R)} ,
    \end{split}
\end{equation}
where $E^\text{AM}(R)$ is Eq. (\ref{eqn:toy-model-energy}), and $I_1$ and $I_0$ are the first-order and zero-order modified Bessel functions of the first kind, respectively. With further expansion, the score becomes
\begin{equation}
    \nabla_R \log p(R) = 2 \left(\frac{I_1(2\beta R)}{I_0(2\beta R)} - R \right) \quad \Rightarrow \quad \nabla_x \log p(x) =   2 \left(\frac{I_1(2\beta R)}{I_0(2\beta R)} - R \right) \frac{x}{R} .
    \label{eq:toy-model-score}
\end{equation}
We implemented Eq. (\ref{eqn:toy-model-energy}) on a $20 \times 20$  grid to obtain the exact energy and score of the model, overlaying 1000 samples from the training set. For visualization purposes, we set $\beta=20$, ensuring a well-defined energy landscape. 
Additionally, we normalized the energy by subtracting its minimum value to ensure that the lowest energy point remains at zero.

To align with the theoretical framework, we employed the variance-exploding (VE) SDE of the form:
\begin{equation}
    \mathrm{d}\rvx_t = \sigma \,\mathrm{d}\mathbf{w}_t
    \label{eq:toymodel-sde}
\end{equation}
where the diffusion coefficient $g(t) = \sigma$. Assuming $t\in (0, 1]$, the variance of the diffusion kernel is given by $ \int_0^t g^2(s) \,\mathrm{d}s = \sigma^2 t$, enabling us to construct the corresponding Gaussian kernel: 
\begin{equation*}
    p(\rvx_t | \rvy) = \mathcal{N}(\rvx_t; \rvy, \sigma^2 t I)
    \label{eq:kernel-toyexample}
\end{equation*}
However, in practice we used $t \in [\epsilon, 1]$ with $\epsilon = 10^{-5}$ for numerical stability. The generative dynamics are given by   
\begin{equation}
    \mathrm{d}\mathbf{x}_{t} = \big[-\sigma^2 \nabla_{\mathbf{x}_{t}} \log p_t(\mathbf{x}_{t})\big] \mathrm{d}t + \sigma^2\,\mathrm{d}\mathbf{w}_{t}.
    \label{eq:toymodel-backward-sde}
\end{equation}
where this runs backwards in time to match the description of \cite{song2021scorebased}. The deterministic Probability Flow ODE is
\begin{equation}
    \frac{\mathrm{d}\mathbf{x}_{t}}{\mathrm{d}t } = -\frac{1}{2}\sigma^2 \nabla_{\mathbf{x}_{t}} \log p_t(\mathbf{x}_{t})
    \label{eq:toymodel-pf-ode}
\end{equation}
which shares the same marginal distributions as the SDE (\ref{eq:toymodel-backward-sde}) and is useful for likelihood estimation. \\

\noindent \textbf{Empirical distribution}. We represented the data as an empirical distribution 
 \begin{equation}
      p(\mathbf{y})=\frac{1}{K}\sum_{\mu=1}^K \delta^{(N)}(\mathbf{y}-\bm{\xi}^\mu)
 \end{equation}
 where $N$ is the dimensionality of the data point $\rvy$. Since $\rvx_t$ is drawn from the forward process distribution $p(\rvx_t | \rvy)$ that is conditioned on the data point $\rvy$, which can be expressed as a Gaussian kernel $\mathcal{N}(\rvx_t; \mathbf{y}, \sigma^2 t I)$. We obtained the distribution $p(\mathbf{x}_t, t)$ as
\begin{equation}
    \begin{split}
    p(\mathbf{x}_t, t) &= \int p(\mathbf{x}_t \mid \mathbf{y}) \,p(\mathbf{y}) \,\mathrm{d}\mathbf{y} \\
    &=\frac{1}{K}\sum_{\mu=1}^K 
    \int \mathcal{N}\!\bigl(\mathbf{x}_t; \mathbf{y}, \sigma^2 t I \bigr)\,
    \delta^{(N)}\!\bigl(\mathbf{y}-\bm{\xi}^\mu\bigr)\,\mathrm{d}\mathbf{y}\\
    & = \frac{1}{K}\sum_{\mu=1}^K
\mathcal{N}\!\bigl(
   \mathbf{x}_t;\,\bm{\xi}^\mu,\,\sigma^2 t\, I
\bigr)\\
    &=\frac{1}{K}\sum_{\mu=1}^K \frac{1}{(2\pi\,\sigma^2t)^{\frac{N}{2}}}\exp\!\Bigl(
-\,\frac{\|\mathbf{x}_t - \bm{\xi}^\mu\|_2^2
    }{2\,\sigma^2 t}\Bigr) 
    \end{split}
\end{equation}
which is the derived $E^\text{DM}$ in the main text, when the variance of the Gaussian kernel is $\sigma^2 t$. \\

\noindent \textbf{Energy Computation}. Following \cite{song2021scorebased} we computed the log-likelihood $\log p_\theta(\rvx_0)$ given by a diffusion model, with the instantaneous change of variable formula \cite{NeuralODE}, where $\rvx_0$ are the generated samples from the model. Replacing Eq. (\ref{eq:toymodel-backward-sde}) into the log-likelihood equation, 
\begin{align}  
    \label{eqn:loglikelihood-dm}
    \log p_0(\rvx_0;\mathbf{\theta}) = \log p_T(\rvx_T) + \int_0^T \nabla \cdot \tilde{\rvf}(\rvx_t, t)  \mathrm{d}t
\end{align}
with $\tilde{\rvf}(\rvx_t, t) =-\frac{1}{2}\sigma^2 \nabla_{\mathbf{x}_{t}} \log p_t(\mathbf{x}_{t})$. The function $\tilde{\rvf}(\rvx_t, t)$ comes from the above Probability Flow ODE (\ref{eq:toymodel-pf-ode}), and $\nabla \cdot \tilde{\rvf}(\rvx_t, t)$ denotes the divergence of the function (or the trace of its Jacobian). 

To estimate the likelihood of the model, we integrated $\nabla \cdot \tilde{\rvf}(\rvx_t, t)$, from a small time $\eps=10^{-5}$ to $T=1$, using a numerical integrator and added the prior logarithmic likelihood to it following Eq. (\ref{eqn:loglikelihood-dm}). The divergence term $\nabla \cdot \tilde{\rvf}(\rvx_t, t)$ is computed using the Laplacian, which is computationally feasible for this toy example, instead of using the Hutchinson trace estimator \cite{hutchinson1989stochastic} done in \cite{song2021scorebased}. We use the RK45 method \cite{DORMAND198019} implemented in the ODE solver \textit{scipy.integrate.solve\_ivp} from Scipy \cite{SciPy} to solve the above integral. 

To compute the energy, we used the equation of $E^\text{DM}$ in the main text, which can be also computed using Eq. (\ref{eqn:loglikelihood-dm}). From the Boltzmann distribution $p(\rvx) = \frac{1}{Z}\exp\bigl[-\beta E(\mathbf{x})\bigl]$, we obtained $\log p(\rvx) \propto -\beta E(\mathbf{x})$. In our case, the inverse temperature is $\beta = \frac{1}{2 \sigma^2 t}$. The equation for our VE-SDE thus becomes 
\begin{equation}
    E^\text{DM}(\rvx_t, t) = -2 \sigma^2 t \log p(\rvx_t, t)  
\end{equation}
To visualize the energy landscape, we observed the energy at $t=0.15$ which corresponds to $\beta=3.3$. \\

\noindent \textbf{Model and Training Details}. For each training data size $K$, we used an MLP-based network to estimate the score $s_\theta(\rvx_t, t)$. The general architecture includes \textbf{(1)} a Fourier random feature (FRF) timestep embedding layer and a linear layer which projects the concatenation of FRF timestep embedding and input into a latent dimension of 256; \textbf{(2)} an encoder which consists of four non-convolutional residual blocks (using the same latent dimension) with Swish \cite{Swish} activation in between; \textbf{(3)} a decoder which consists of the same number of residual blocks; and \textbf{(4)} a linear block which projects the latent variable back to the 2D space.

 The network was trained in continuous time with $\sigma=1$ using the objective function from \cite{ho2020denoising, song2021scorebased}. Optimization is performed using the Adam \cite{Adam} optimizer with a learning rate $lr = 10^{-4}$. The batch size is set to $\textbf{min}(K,500)$. All models were trained for $800000$ iterations with the maximum batch size of $500$. Meanwhile, we used Euler-Maruyama discretization method \cite{kloeden1992stochastic}, with 1000 discretized steps, to solve the reverse SDE (\ref{eq:toymodel-backward-sde}) to generate samples. \\

\noindent \textbf{Sample~Clustering}. To cluster the generated samples, we used the \texttt{AgglomerativeClustering} algorithm from scikit-learn, a part of SciPy \cite{SciPy}.

\subsection{Log-Likelihood of Diffusion Models}
\label{sec:likelihood-dm-details}

For concreteness, we provide the likelihood computation of Eq. (\ref{eqn:loglikelihood-dm}) within this section to highlight the formulation of Eq. (\ref{eqn:loglikelihood-dm}) done by \cite{song2021scorebased, NeuralODE}. \\

\noindent \textbf{Probability Flow ODE}. Suppose we have the following forward process:
\begin{equation}
    \mathrm{d}\rvx_t = \mathbf{f}(\rvx_t, t) \mathrm{d}t + \mathbf{g}(\rvx_t, t) \mathrm{d} \mathbf{w}_t
\end{equation}
where $\mathbf{f}(\cdot, t) : \mathbb{R}^{N} \rightarrow \mathbb{R}^N$, $\mathbf{g}(\cdot, t): \mathbb{R}^{N} \rightarrow \mathbb{R}^{N \times N}$ and $N$ denotes the dimensionality of $\rvx_t$. Using the derivations done by \cite{song2021scorebased} in their Appendix D, the evolution of the marginal probability density $p_t(\rvx_t)$ is
\begin{equation}
    \frac{\partial p_t(\rvx_t)}{\partial t} = - \sum^N_{i = 1} \frac{\partial}{\partial x_i} \big [  f_i (\rvx_t, t) p_t (\rvx_t)\big ] + \frac{1}{2} \sum^N_{i = 1} \sum^N_{j = 1} \frac{\partial^2}{\partial x_i \partial x_j} \bigg [\sum^N_{k = 1} g_{ij} (\rvx_t, t) g_{jk} (\rvx_t, t) p_t(\rvx_t) \bigg] 
\end{equation}
which corresponds to the Fokker-Planck equation \cite{OksendalSDE}: 
\begin{equation}
\begin{split}
    \frac{\partial p_t(\rvx_t)}{\partial t} & = - \sum^N_{i = 1} \frac{\partial}{\partial x_i} \big [  f_i (\rvx_t, t) p_t (\rvx_t)\big ] + \frac{1}{2} \sum^N_{i = 1} \frac{\partial}{\partial x_i} \bigg [ \sum^N_{j = 1} \frac{\partial }{\partial x_j} \bigg [\sum^N_{k = 1} g_{ij} (\rvx_t, t) g_{jk} (\rvx_t, t) p_t(\rvx_t) \bigg] \bigg ] \\
    & = - \sum^N_{i = 1} \frac{\partial}{\partial x_i} \big [  f_i (\rvx_t, t) p_t (\rvx_t)\big ] \\
    & \quad \,\, + \frac{1}{2} \sum^N_{i = 1} \frac{\partial}{\partial x_i} \bigg [ p_t (\rvx_t) \, \nabla \cdot \big[\mathbf{g} (\rvx_t, t) \mathbf{g} (\rvx_t, t)^\top \big] + p_t(\rvx_t) \big [ \mathbf{g} (\rvx_t, t) \mathbf{g} (\rvx_t, t)^\top \big ] \nabla_{\rvx_t} \log p_t(\rvx_t) \bigg] \\ 
    & = - \sum^N_{i = 1} \frac{\partial}{\partial x_i} \bigg \{  f_i (\rvx_t, t) p_t (\rvx_t) \\
    &\quad \,\, - \frac{1}{2} \bigg [ \nabla \cdot \big[\mathbf{g} (\rvx_t, t) \mathbf{g} (\rvx_t, t)^\top \big] + \big [ \mathbf{g} (\rvx_t, t) \mathbf{g} (\rvx_t, t)^\top \big ] \nabla_{\rvx_t} \log p_t(\rvx_t) \bigg ] p_t(\rvx_t) \bigg \} \\ 
    & = - \sum^N_{i = 1} \frac{\partial}{\partial x_i} \big [ \tilde{f}_i(\rvx_t, t) \big ] p_t(\rvx_t)
\end{split}    
\label{eqn:diffusion-fokker-placnk}
\end{equation}
where $\tilde{\rvf}(\rvx_t, t) = \rvf (\rvx_t, t) - \frac{1}{2} \nabla \cdot \big[\mathbf{g} (\rvx_t, t) \mathbf{g} (\rvx_t, t)^\top \big] - \frac{1}{2}  \big [ \mathbf{g} (\rvx_t, t) \mathbf{g} (\rvx_t, t)^\top \big ] \nabla_{\rvx_t} \log p_t(\rvx_t) $. With careful inspection of Eq. (\ref{eqn:diffusion-fokker-placnk}), it is equal to the Liouville equation if the diffusion term $\tilde{\mathbf{g}}(\rvx, t) = 0$ and essentially, it is the probability flow ODE where
\begin{equation}
\begin{split}
\mathrm{d} \rvx_t & = \tilde{\rvf}(\rvx_t, t)\mathrm{d} t +\underset{0}{\underbrace{\tilde{\mathbf{g}} (\rvx_t, t) \mathrm{d} \rvw_t}} \\ 
& = \bigg \{\rvf (\rvx_t, t) - \frac{1}{2} \nabla \cdot \big[\mathbf{g} (\rvx_t, t) \mathbf{g} (\rvx_t, t)^\top \big] -\frac{1}{2} \big [ \mathbf{g} (\rvx_t, t) \mathbf{g} (\rvx_t, t)^\top \big ] \nabla_{\rvx_t} \log p_t(\rvx_t) \bigg \} \mathrm{d} t
\end{split}
\label{eqn:general-probability-flow-ode}
\end{equation}
Using the ODE (\ref{eqn:general-probability-flow-ode}), we can derive an appropriate probability flow ODE from the forward process detailed in the main text. For example, we have the following equation
\begin{equation}
    \mathrm{d} \rvx_t = \underset{:= \, \tilde{\mathbf{f}}(\rvx_t, t)}{\underbrace{\bigg \{ \mathbf{f} (\rvx_t, t) - \frac{1}{2} g(t)^2 \nabla_{\rvx_t} \log p_t (\rvx_t)\bigg \}}} \mathrm{d} t
\end{equation}
for the toy model where $\mathbf{f} (\rvx_t, t) = 0$. \\

\noindent \textbf{Log-likelihood}. Furthermore, if we first take the logarithm of Eq. (\ref{eqn:diffusion-fokker-placnk}) where 
\begin{equation}
    \begin{split}
        \frac{\partial \log p_t(\rvx_t)}{\partial t} = \frac{1}{p_t(\rvx_t)} \frac{\partial p_t(\rvx_t)}{\partial t} = - \sum^N_{i = 1} \frac{\partial \tilde{f}_i(\rvx_t, t)}{\partial x_i} = \nabla \cdot \tilde{\rvf} (\rvx_t, t)
    \end{split}
\end{equation}
we can compute the log-likelihood of $p_0(\rvx_0)$ using the following equation
\begin{align} 
    \log p_0(\rvx_0) = \log p_T(\rvx_T) + \int_0^T \nabla \cdot \tilde{\rvf}(\rvx_t, t)  \mathrm{d}t
    \label{eqn:loglikelihood-dm-general}
\end{align}
where $\nabla \cdot \tilde{\rvf}(\rvx_t, t)$ is parameterized as $\nabla \cdot \tilde{\rvf}_\theta (\rvx_t, t)$ since $s_\theta(\rvx_t, t) = \nabla_{\rvx_t} \log p_t(\rvx_t; \theta)$. Moreover, in general the term $\nabla \cdot \tilde{\rvf}_\theta (\rvx_t, t)$ is computed via the Hutchinson trace estimator \cite{hutchinson1989stochastic}, 
\begin{equation}
    \nabla \cdot \tilde{\rvf}_\theta (\rvx_t, t) = \mathbb{E}_{p(\epsilon)} \big [ \epsilon^\top \nabla_{\rvx} \tilde{\rvf}_\theta (\rvx_t, t) \epsilon\big] 
\end{equation}
where $\nabla_{\rvx} \tilde{\rvf}_\theta (\rvx_t, t)$ denotes the Jacobian of $\tilde{\rvf}_\theta (\rvx_t, t)$ and the random variable $\epsilon$ satisfies $\mathbb{E}_{p(\epsilon)} [\epsilon]= 0$ and $\text{Cov}_{p(\epsilon)} [\epsilon] = I$. 

However, for the toy example, due to its low dimensionality we found that it is possible to not utilize the Hutchinson trace estimator at all, and instead opted to compute the trace of the Jacobian without using the noise estimators.

\section{Memorization-to-Generalization Transition}
\label{sec:transition-details}
\subsection{Selection of Points}
For the computation of \cref{fig:transition-plots} in the main text, we followed the experiment setup of \cite{yoon2023diffusion} and conducted an initial sparse search of the transition, starting at data size $500$ and doubling it all the way to the total data size $|\mathsf{S}|$, i.e., $K \in \{500, 1000, 2000, \dots, |\mathsf{S}|\}$. We then identified two transitional critical points, $A$ and $B$, see Table (\ref{tab:transitions}), to conduct a more thorough search to elucidate the memorization-to-generalization transition. 

Here, point $A$ indicates the initial drop in memorization while $B$ signals the plateauing of memorization. To capture the finer details of this transition, we performed a search of $30$ inclusive linearly spaced points, from point $A$ to point $B$. For regions outside of the transition, using linear spacing, we sampled $5$ points from $|\mathsf{S}| = 2$ to point $A$, and another $5$ points from $B$ to the total dataset size $|\mathsf{S}|$, inclusively. In other words, we trained a separate model for each selected data size and generated the corresponding evaluation and synthetic sets while keeping the same configuration for all models, see Table (\ref{tab:models}). Then, we computed the memorization, spurious, and generalization fractions using the detection metrics in Sec. (4). The selection process for the values $\delta_m$ and $\delta_s$ is explained below.

\begin{table}[h]
\caption{A table showing the critical points $A$ and $B$ of the memorization-generalization transition for each dataset.}
\label{tab:transitions}
\begin{tabular*}{\textwidth}{@{\extracolsep\fill}lccc}
\toprule%
& \multicolumn{3}{@{}c@{}}{Training Data Size} \\ \cmidrule{2-4}%
Dataset & Point $A$ & Point $B$ & Total \\
\midrule
MNIST & 4000 & 32000 & 60000 \\ 
CIFAR10 & 2000 & 16000 & 50000 \\ 
LSUN-CHURCH & 2000 & 16000 & 126227 \\
FASHION-MNIST&  4000 & 16000 & 60000 \\ 
\bottomrule
\end{tabular*}
\footnotetext{Note: We use 30 linearly spaced points between $A$ and $B$ to finely characterize the memorization-to-generalization transition.}
\end{table}

\subsection{Selection of Distance Metric}
For high-dimensional datasets, CIFAR10 \cite{cifar10} and LSUN-CHURCH \cite{lsun}, we utilized LPIPS \cite{lpips} with the AlexNet \cite{AlexNet} backbone, as the function $d(\cdot, \cdot)$ for both memorization and spurious detection metrics. We selected this approach since it is a commonly used perceptual metric that compares the similarity between two images based on their feature representations. Moreover, it has been shown to better align with human judgment of visual similarity, making it ideal for assessing the quality and diversity of generated samples in these high-dimensional image datasets \cite{lpips}. For simpler datasets like MNIST \cite{mnist} and FASHION-MNIST \cite{fmnist}, where images are single-channel and less complex, we found that $L_2$-distance suffices for both memorization and spurious detection metrics.

\subsection{Selection of Threshold Values}
The detection thresholds, $\delta_m$ and $\delta_s$, were set based on the chosen distance metric $d(\cdot, \cdot)$ and visual inspection of the distance histograms for bimodality as mentioned in Sec. (4). Moreover, these thresholds were chosen to reflect the varying visual complexity and feature richness of the datasets, and their values for each dataset are shown in Table (\ref{tab:thresholds}). Specifically, for each training data size $K$ of each dataset, we manually inspected the bimodality of the olive and gray histograms to select $\delta_m$ and $\delta_s$. Moreover, to better tune these thresholds at each $K$, we manually inspected the least $5$ spurious samples, ensuring that they do not resemble memorized samples, while doing the same for memorized samples. If the resemblance of one set to another is too strong, we decreased the corresponding threshold. Please refer to Figs.~(\ref{fig:church-hist-figure})-(\ref{fig:mnist-hist-figure}) below and \cref{fig:cifar10-hist} in the main text for examples of the histograms. Please note that the bimodality of the gray histograms in FASHION-MNIST might not be significant, but it is enough to select their respective $\delta_s$.

%For CIFAR10, we utilized the histograms obtained at $K = 7310$, which demonstrate strong bimodality shape, to tune our threshold values to minimize the number of spurious samples classified as memorized samples and vice versa, via manual inspection. For manual inspection, a general rule we followed --- it is better to have more false positives in the generalized set than the memorized set since spurious samples can be seen as early-generalized samples. Thus, we observe the least-$5$ spurious samples and ensure that they do not resemble memorized samples. We performed the same process for memorized samples. The entire process is then repeated for the other three datasets. We utilized $K = 4896$ for LSUN-CHURCH, $K = 21379$ for MNIST, and $K = 7724$ for FASHION-MNIST to select $\delta_m$ and $\delta_s$. For more details on the selected histograms, please refer to Figs. (\ref{fig:cifar10-hist}), (\ref{fig:mnist-hist-figure}), (\ref{fig:church-hist-figure}), and (\ref{fig:fmnist-hist-figure}).

\subsection{Model and Training Details}
For each point in our transition plots of \cref{fig:transition-plots}, we trained a DDPM-based diffusion model, where the score model is a PixelCNN++ based U-Net \cite{van2016conditional, salimans2017pixelcnn++}. We kept the variances, $\beta_\text{min} = 10^{-4}$ and $\beta_\text{max}$ = $2 \times 10^{-2}$, timesteps $T = 1000$, and learning rate $lr =$ $2 \times 10^{-4}$ for all models and datasets. Each model has $2$ residual blocks \cite{he2016deep} for each down-sampling and up-sampling layer, while an attention block is placed at $16\times$ resolution. We only modified the channel multipliers for each model based on the complexity of the dataset, see Table (\ref{tab:models}). If the training data size $K$ is smaller than the specified batch size, the batch size is set to be equal to $K$. For generation or inference, we used the exponential moving average of each trained model, as delineated in \cite{ho2020denoising}, which is obtained with the decay value set as $0.9999$ during training. Please note that we did not use random flipping in the training of our models, since we want our measurements to reflect the training data size at best as random flipping implicitly increases the number of patterns that the models see during training. However, we did use dropout (of value $0.1$) for the training of CIFAR10, MNIST, and FASHION-MNIST models. For LSUN-CHURCH dataset, the images were center-cropped and down-sampled to $64 \times 64$ resolution. Lastly, for each of the training set $\mathsf{S}_\alpha$, where $\alpha = 1, \dots, M$, they are split from the original dataset given a specific size, using the same random seed value of $3407$.

\subsection{Additional Results}

\begin{figure}[t]
    \centering
    \includegraphics[keepaspectratio, width=1.0\linewidth]{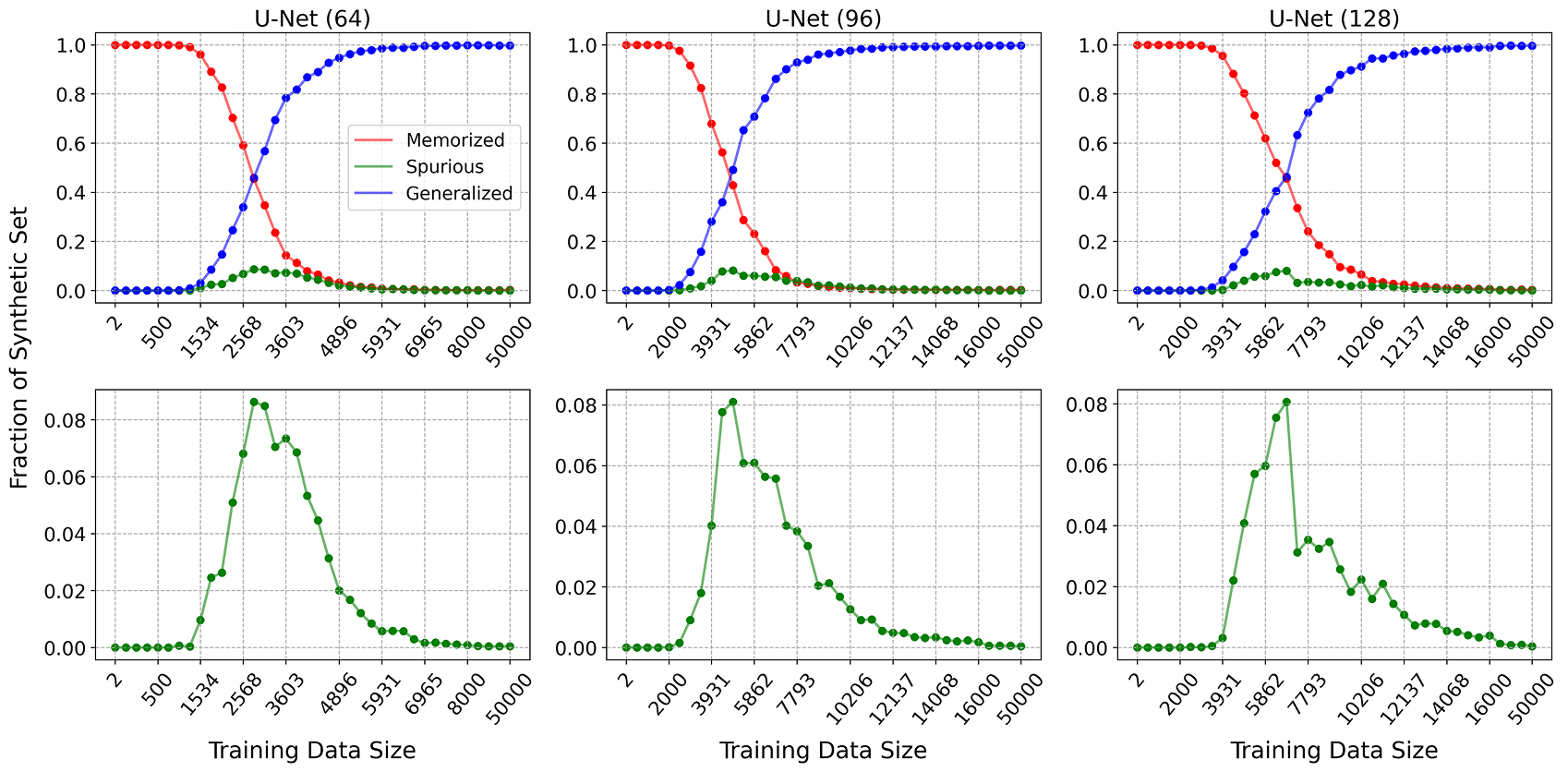}
    \caption{
    The fractions of \textcolor{red}{memorized}, \textcolor[rgb]{0, 0.6, 0}{spurious}, and \textcolor{blue}{generalized} samples in synthetic sets across different training sizes for CIFAR10 \cite{cifar10} and various model's widths. As the training data size $K$ increases, memorization decreases while the emergence of spurious patterns follows and the fraction of generalized patterns steadily increase (see \textsf{top row}). The fraction of spurious patterns rises and decreases at the boundary between the memorization and generalization phases (see \textsf{bottom row}). Additionally, with a smaller number of parameters, the memorization-generalization transition happens at an earlier stage (or smaller $K$), since the model has a memorization capacity. In contrast, as the model's size increases, the transition happens at an later stage (or larger $K$), where it is able to memorize more training data. The numbering, e.g., $64$, $96$, and $128$, denotes the initial latent dimension of the U-Net, see Table~(\ref{tab:models}).
    }
    \label{fig:cifar10-transitions-extra}
\end{figure}

\begin{figure}[t]
    \centering
    \includegraphics[keepaspectratio, width=1.0\linewidth]{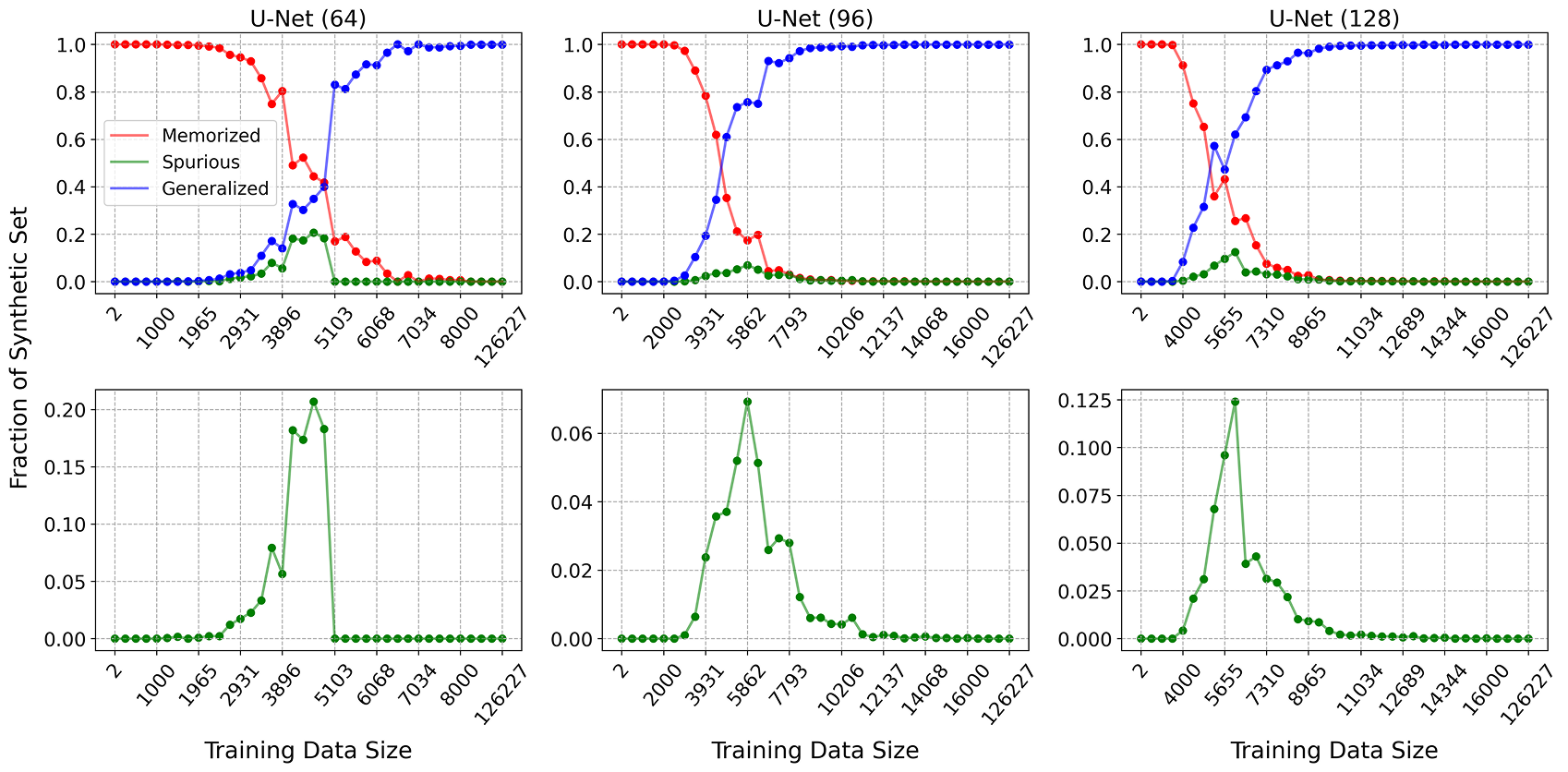}
    \caption{
    The fractions of \textcolor{red}{memorized}, \textcolor[rgb]{0, 0.6, 0}{spurious}, and \textcolor{blue}{generalized} samples in synthetic sets across different training sizes for LSUN-CHURCH \cite{lsun} and various model's widths. As the training data size $K$ increases, memorization decreases while the emergence of spurious patterns follows and the fraction of generalized patterns steadily increase (see \textsf{top row}). The fraction of spurious patterns rises and decreases at the boundary between the memorization and generalization phases (see \textsf{bottom row}). Additionally, with a smaller number of parameters, the memorization-generalization transition happens at an earlier stage (or smaller $K$), since the model has a memorization capacity. In contrast, as the model's size increases, the transition happens at an later stage (or larger $K$), where it is able to memorize more training data. The numbering, e.g., $64$, $96$, and $128$, denotes the initial latent dimension of the U-Net, see Table~(\ref{tab:models}).
    }
    \label{fig:church-transitions-extra}
\end{figure}

Inspired by \cite{yoon2023diffusion, gu2023memorization}, we extend our investigation on the memorization-generalization transition by studying the effects of varying the parameter size of the diffusion model for CIFAR10 \cite{cifar10} and LSUN-CHURCH \cite{lsun}. The results of this experiment are recorded in Figs.~(\ref{fig:cifar10-transitions-extra})-(\ref{fig:church-transitions-extra}) for these two datasets respectively. Specifically, we trained two additional sets of 38 diffusion models, with the initial latent dimensions of 64 and 96 respectively, and repeated the same experimental setup delineated above, see also Table~(\ref{tab:models}). 

For CIFAR10 and the U-Net (64) models, we identified the critical points, $A$ and $B$, to be $500$ and $8000$ respectively. Meanwhile, for the U-Net (96) models, these critical points are $2000$ and $16000$. To select $\delta_m$ and $\delta_s$, we followed the same process detailed above by observing the distance histograms to see where bimodality occurs with respect to the distance value, and manually inspecting the images to ensure there is a minimal amount of false positives in memorized and spurious samples. Similarly, for the LSUN-CHURCH models, we identified the critical points, $A$ and $B$, to be $1000$ and $8000$ for the U-Net (64) models; and for the larger U-Net (128) models, $4000$ and $16000$ as the critical points. Please refer to Table (\ref{tab:thresholds}) for the threshold values. 

%For U-Net (64) models, the threshold values $\delta_m$ and $\delta_s$ are both selected to be $0.02$ based on the bimodality of the distance histograms observed at $K = 2568$ --- meanwhile, for U-Net (96), the threshold values $\delta_m$ and $\delta_s$ are chosen as $0.03$ and $0.02$ according to distance histograms observed at $K = 3931$.

%Similarly, for the LSUN-CHURCH models, we identified the critical points, $A$ and $B$, to be $1000$ and $8000$ for the U-Net (64) models; and for the larger U-Net (128) models, $4000$ and $16000$ as the critical points. The threshold values, $\delta_m$ and $\delta_s$, are both selected to be $0.04$ after observing the bimodality of the distances at $K=3413$ --- meanwhile, for U-Net (128), these values are selected to be $\delta_s = 0.04$ and $\delta_m = 0.1$ after observing the distance histograms at $K = 5241$. 

With the smaller U-Net (64) models, we observe the memorization-generalization transition happens at an earlier stage with respect to the larger U-Net (128) models, where the peak of spurious samples appears at $K = 2568$ instead of $K = 5862$ for CIFAR10, and $K = 4862$ instead of $K = 6086$ for LSUN-CHURCH.
Overall, these results indicate that as the parameter size of the diffusion model increases, the memorization capacity of the model is increased and thus, generalization is delayed since the model can memorize more of its training data denoted by the appearance of spurious patterns happening at a larger training data size $K$. In contrast, when the memorization capacity is reduced, generalization is initiated at an earlier stage (or smaller training data size) given a larger parameterized model. Most importantly, the emergence of spurious patterns remains consistent despite the variations in the parameter size of the diffusion model: these patterns quickly rise and fall at the boundaries of the memorization-to-generalization transition.

\begin{figure}[!t]
    \centering
    \includegraphics[keepaspectratio, width=0.65\linewidth]{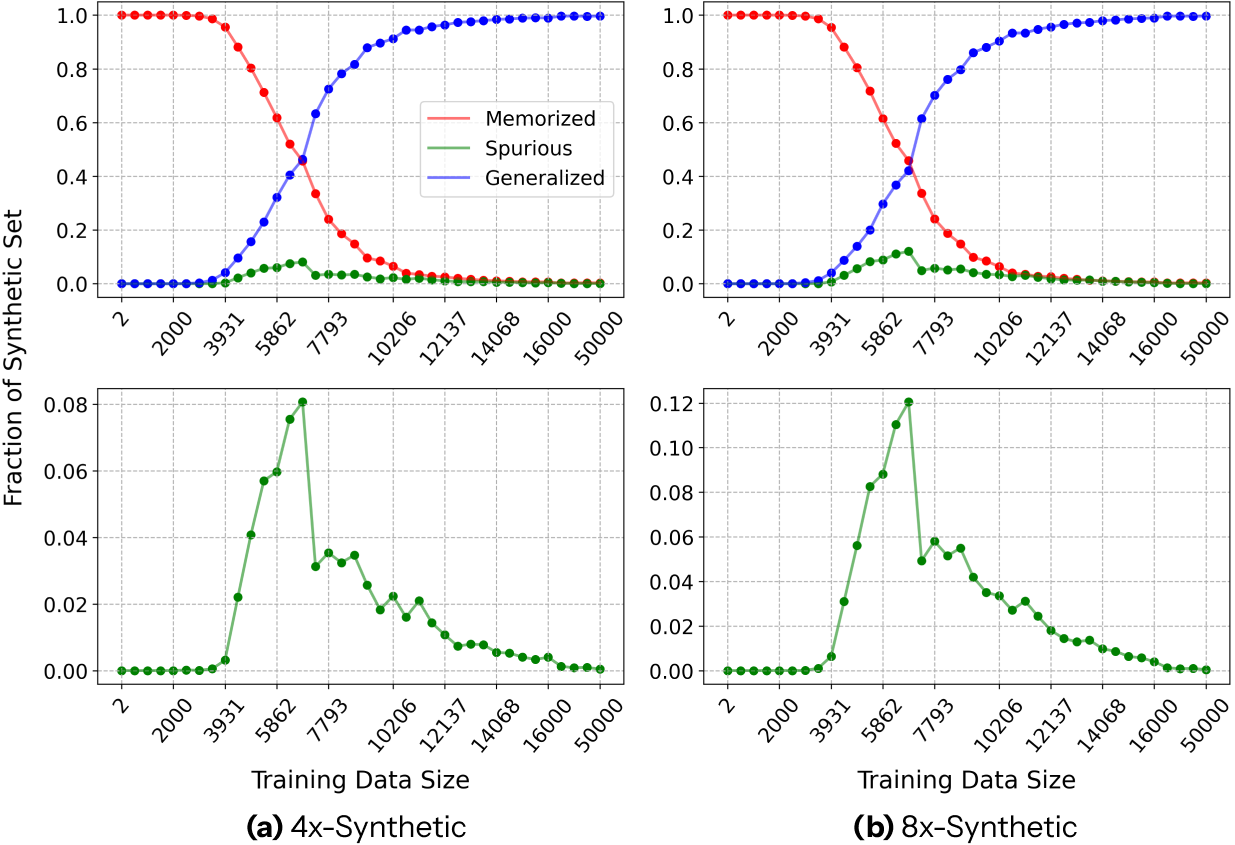}
    \caption{
    Memorization-Generalization transition plots for the CIFAR10 dataset. The column \textsf{(a)} figure depicts the memorization-generalization transition computed with $4\times$ synthetic sets for various training data sizes. Meanwhile, the column \textsf{(b)} figure shows the transition computed using $8\times$ synthetic sets for the same training data sizes. With the larger synthetic sets, more spurious patterns are detected around the peak of the transition. However, the overall trend of these patterns quickly rise, during the co-existence of memorization and generalization, and fall as generalization becomes the dominant phase, is maintained. Both figures use the same memorized and spurious threshold values detailed in Table~(\ref{tab:thresholds}).
    }
    \label{fig:cifar10-4x-8x}
\end{figure}

Additionally, to verify the consistency of the memorization-generalization transition, we explored the influence of larger synthetic sets and recorded our results in Fig.~(\ref{fig:cifar10-4x-8x}). Specifically, we repeated our experiment done in Fig. (4) for CIFAR10 using $8\times$ synthetic sets, which was originally done using $4 \times$ synthetic sets. Based on these results, we observe a slight increase in the fraction of spurious samples when using the larger $8\times$ synthetic sets. Specifically, at $K = 5862$, the fraction increases from $0.04$ to roughly $0.06$ with respect to the synthetic set size. Nonetheless, the trend of memorization-generalization, with the emergence of spurious samples, is still maintained regardless of the multiplier for the synthetic size. This aspect highlights the consistency of the spurious phenomenon in diffusion models in relation to the varying of training data size.

\clearpage 
\begin{sidewaystable}[ht]
\caption{Table displaying the range of the memorized and spurious threshold values in the computation of the transition plots in Fig.~(4) of the main text.}
\label{tab:thresholds}
\begin{tabular*}{\textwidth}{@{\extracolsep\fill}lccc}
\toprule%
& & \multicolumn{2}{c}{Range [min, max]} \\ 
\cmidrule{3-4}
Dataset & Initial Latent Dim. & $\delta_m$ & $\delta_s$ \\ 
\midrule
MNIST & 128 & $[2, 4.5]$ & $[2, 4]$ \\ 
FASHION-MNIST & 128 & $[2, 2.5]$ & $[2.5, 3]$ \\ 
\hline 
& 64 & $[0.01, 0.04]$ & $[0.015, 0.025]$ \\
CIFAR10 & 96 & $[0.015, 0.04]$ & $[0.015, 0.025]$ \\
& 128 & $[0.015, 0.035]$ & $[0.015, 0.025]$ \\
\hline 
& 64 & $[0.06, 0.15]$ & $[0.01, 0.1]$ \\
LSUN-CHURCH & 96 & $[0.1, 0.125]$ & $[0.05, 0.06]$ \\
& 128 & $[0.1, 0.12]$ & $[0.04, 0.06]$ \\
\bottomrule
\end{tabular*}
\footnotetext{The full set of threshold values is provided alongside with the code.}
\end{sidewaystable}

\begin{sidewaystable}[h]
\caption{A table displaying both model and training configurations for each dataset.}
\label{tab:models}
\begin{tabular*}{\textwidth}{@{\extracolsep\fill}lccccc}
\toprule%
& \multicolumn{5}{c}{Configurations} \\ 
\cmidrule{2-6}
Dataset & Initial Latent Dim. & Channel Multipliers & Num. of Param. & Batch Size & Training Iterations \\
\midrule
 MNIST & 128 & (1, 2, 2) & 24.5M & 128 & 400,000 \\ 
FASHION-MNIST & 128 & (1, 2, 2) & 24.5M & 128 & 400,000 \\ 
\hline 
& 64 & (1, 2, 2, 2) & 8.9M & 128 & 500,000 \\
CIFAR10 & 96 & (1, 2, 2, 2) & 20.1M & 128 & 500,000 \\
& 128 & (1, 2, 2, 2) & 35.7M & 128 & 500,000 \\
\hline 
& 64 & (1, 1, 2, 2, 4, 4) & 27.4M & 64 & 800,000 \\
LSUN-CHURCH & 96 & (1, 1, 2, 2, 4, 4) & 61.7M & 64 & 800,000 \\
& 128 & (1, 1, 2, 2, 4, 4) & 109.7M & 64 & 800,000\\
\bottomrule
\end{tabular*}
\end{sidewaystable}

\clearpage

\newpage
\subsection{Additional Visualizations of Distance Histogram}
\begin{figure}[h]
    %\vspace{-5mm}
    %\vskip 0.2in
    \centering
    \setlength{\abovecaptionskip}{5pt}
    \setlength{\belowcaptionskip}{5pt} 
    \includegraphics[keepaspectratio, width=1\textwidth, height=1\textheight]{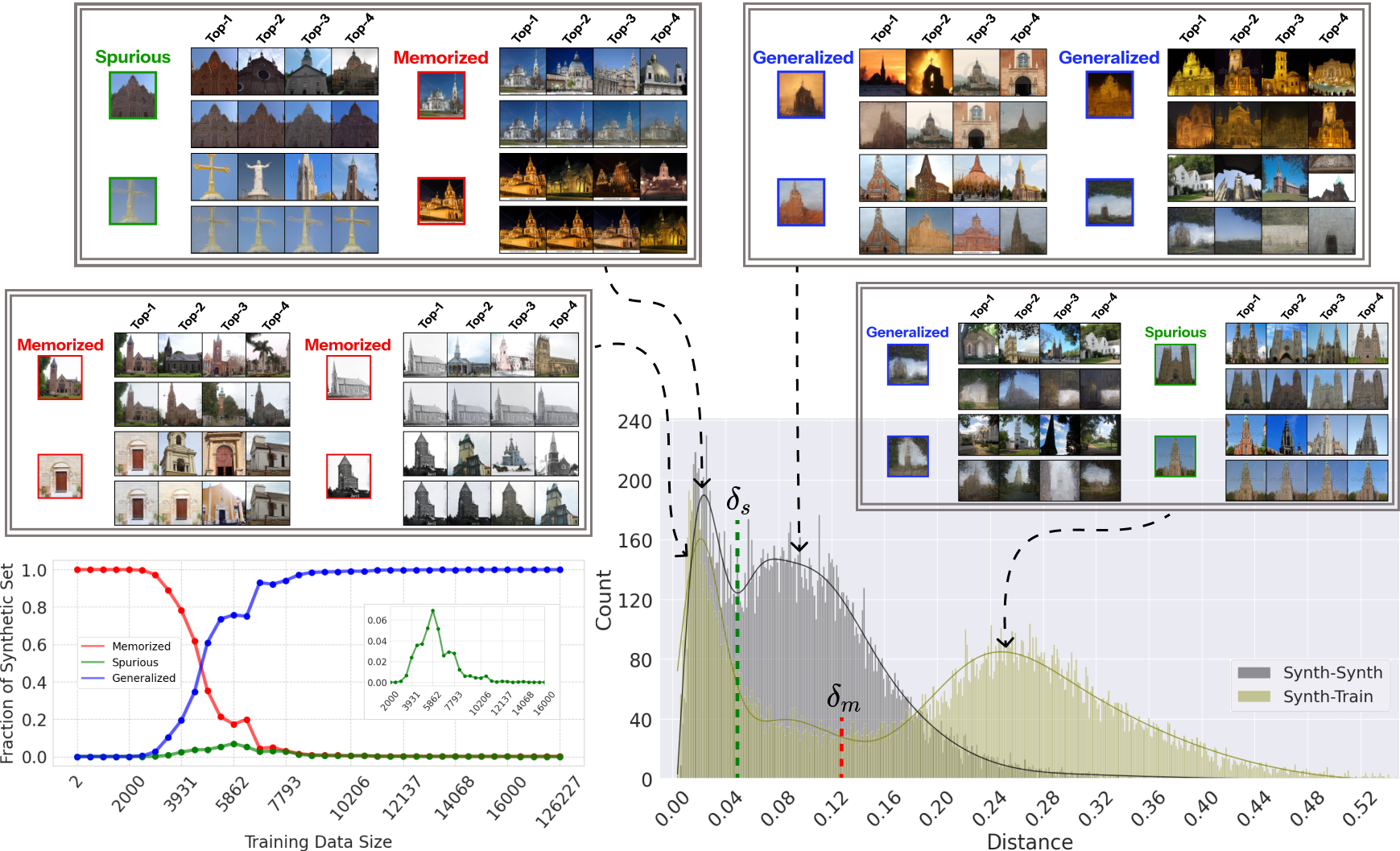}
    \caption{Different sample types across the memorization-to-generalization transition for LSUN-CHURCH \cite{lsun}, defined by the spurious and memorized thresholds, $\delta_s$ and $\delta_m$. The \textcolor[gray]{0.4}{grey histogram} shows the distances between synthetic samples and their nearest neighbors from the synthetic set $\mathsf{S}'$. The threshold $\delta_s$ is defined as a boundary between the two peaks. The \textcolor{olive}{olive histogram} depicts the distances from the synthetic samples to their closest neighbor from the training set $\mathsf{S}$, with threshold $\delta_m$ separating the two peaks. The threshold $\delta_m$ is chosen much stricter here such that it works well for all training dataset sizes via visual inspection. \textcolor{red}{Memorized} samples are located in the left peak of the olive histogram, below $\delta_m$. In contrast, \textcolor{blue}{generalized} and \textcolor[rgb]{0, 0.6, 0}{spurious} samples appear to the right of $\delta_m$ (olive histogram). Examples of the generated samples forming each of the four peaks of the histograms are shown in the inset frames. For each generated sample \textsf{top-4} nearest neighbor images from the training set are shown in the \textsf{top row}, and \textsf{top-4} nearest neighbors from the synthetic set are shown in the \textsf{bottom row}. Training set size $K = 4896$ was used in this figure (at the peak of the frequency of spurious states), but phenomena discussed are generic and largely independent of this specific value. The fraction of the memorized, spurious, and generalized samples in the pool of all generated samples is shown in the bottom left panel as a function of the training set size. The inset shows amplified spurious fraction (\textcolor[rgb]{0, 0.6, 0}{green curve}). 
    }
    \label{fig:church-hist-figure}
\end{figure}
\clearpage
\begin{figure}[h]
    %\vspace{-2mm}
    %\vskip 0.2in
    \centering
    \setlength{\abovecaptionskip}{5pt}
    \setlength{\belowcaptionskip}{5pt} 
    \includegraphics[keepaspectratio, width=1\textwidth, height=1\textheight]{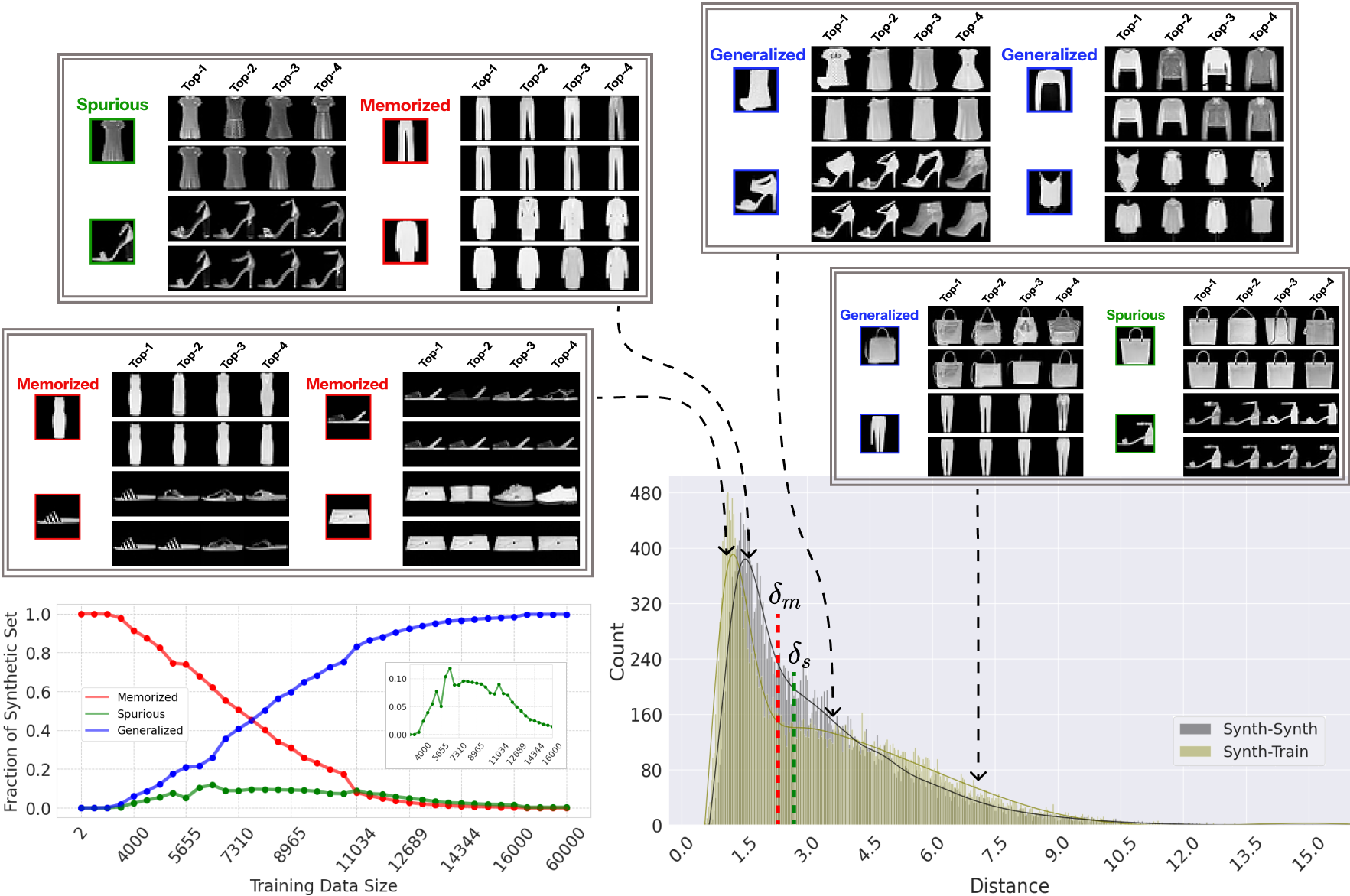}
    \caption{Different sample types across the memorization-to-generalization transition for FASHION-MNIST \cite{fmnist}, defined by the spurious and memorized thresholds, $\delta_s$ and $\delta_m$. The \textcolor[gray]{0.4}{grey histogram} shows the distances between synthetic samples and their nearest neighbors from the synthetic set $\mathsf{S}'$. The threshold $\delta_s$ is defined as a boundary between the two peaks. The \textcolor{olive}{olive histogram} depicts the distances from the synthetic samples to their closest neighbor from the training set $\mathsf{S}$, with threshold $\delta_m$ separating the two peaks. \textcolor{red}{Memorized} samples are located in the left peak of the olive histogram, below $\delta_m$. In contrast, \textcolor{blue}{generalized} and \textcolor[rgb]{0, 0.6, 0}{spurious} samples appear to the right of $\delta_m$ (olive histogram). Examples of the generated samples forming each of the four peaks of the histograms are shown in the inset frames. For each generated sample \textsf{top-4} nearest neighbor images from the training set are shown in the \textsf{top row}, and \textsf{top-4} nearest neighbors from the synthetic set are shown in the \textsf{bottom row}. Training set size $K = 7724$ was used in this figure (at the peak of the frequency of spurious states), but phenomena discussed are generic and largely independent of this specific value. The fraction of the memorized, spurious, and generalized samples in the pool of all generated samples is shown in the bottom left panel as a function of the training set size. The inset shows amplified spurious fraction (\textcolor[rgb]{0, 0.6, 0}{green curve}). 
    }
    \label{fig:fmnist-hist-figure}
\end{figure}
\clearpage 
\begin{figure}[h]
    %\vspace{-50mm}
    %\vskip -3.35in
    \centering
    \setlength{\abovecaptionskip}{5pt}
    \setlength{\belowcaptionskip}{5pt} 
    \includegraphics[keepaspectratio, width=1\textwidth, height=1\textheight]{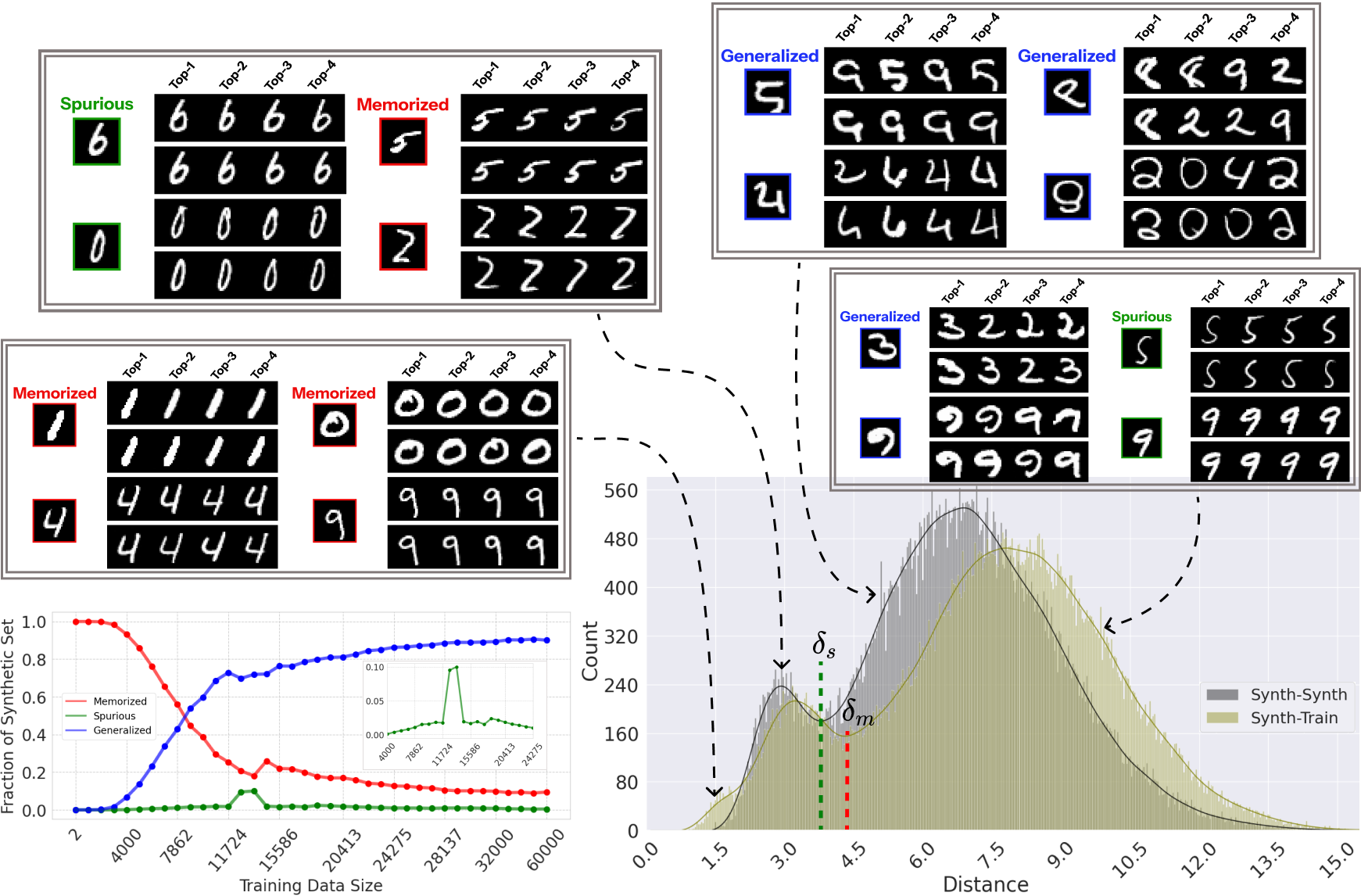}
    \caption{Different sample types across the memorization-to-generalization transition for MNIST \cite{mnist}, defined by the spurious and memorized thresholds, $\delta_s$ and $\delta_m$. The \textcolor[gray]{0.4}{grey histogram} shows the distances between synthetic samples and their nearest neighbors from the synthetic set $\mathsf{S}'$. The threshold $\delta_s$ is defined as a boundary between the two peaks. The \textcolor{olive}{olive histogram} depicts the distances from the synthetic samples to their closest neighbor from the training set $\mathsf{S}$, with threshold $\delta_m$ separating the two peaks. \textcolor{red}{Memorized} samples are located in the left peak of the olive histogram, below $\delta_m$. In contrast, \textcolor{blue}{generalized} and \textcolor[rgb]{0, 0.6, 0}{spurious} samples appear to the right of $\delta_m$ (olive histogram). Examples of the generated samples forming each of the four peaks of the histograms are shown in the inset frames. For each generated sample \textsf{top-4} nearest neighbor images from the training set are shown in the \textsf{top row}, and \textsf{top-4} nearest neighbors from the synthetic set are shown in the \textsf{bottom row}. Training set size $K = 21379$ was used in this figure, which is slightly past the peak of the frequency of spurious states.  The phenomena discussed are generic and largely independent of this specific value. The fraction of the memorized, spurious, and generalized samples in the pool of all generated samples is shown in the bottom left panel as a function of the training set size. The inset shows amplified spurious fraction (\textcolor[rgb]{0, 0.6, 0}{green curve}). 
    }
    \label{fig:mnist-hist-figure}
    %\vspace{-5mm}
    %\vskip -0.2in
\end{figure}
\clearpage

\subsection{Additional Examples of the Three Distinct Sample Types}
\begin{figure}[h]
    \centering
    \includegraphics[width=1.0\textwidth]{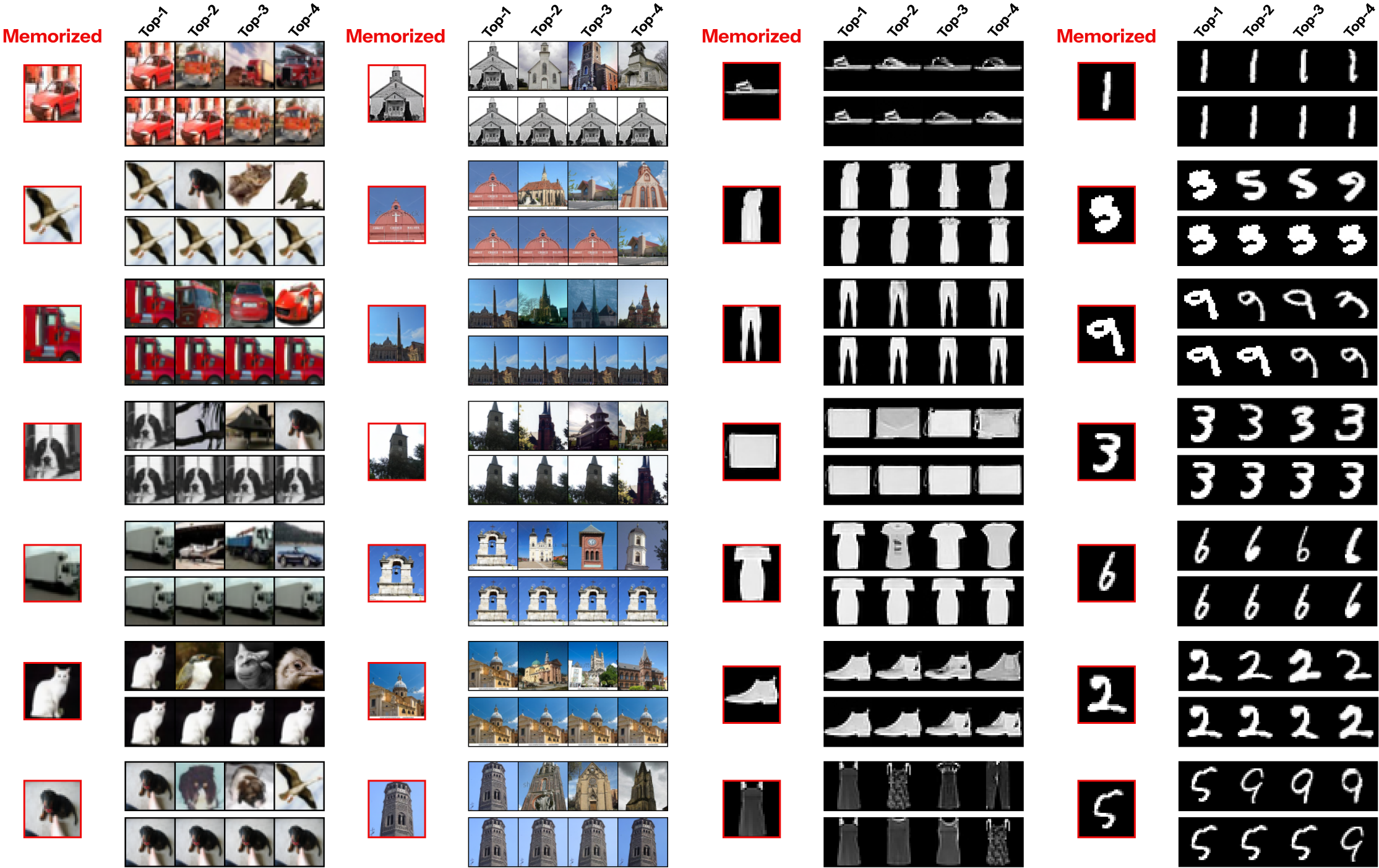}
    \caption{Visualizations of additional \textcolor{red}{memorized} patterns and their {\textsf{top-4}} nearest neighbors for different datasets. The {\textsf{top row}} illustrates nearest neighbors from the training set while the {\textsf{bottom row}} depicts those from the synthetic set. Memorized samples are duplicates of the training set $\mathsf{S}$. During the strong memorization phase, duplicates are also found within the synthetic set $\mathsf{S}'$. Note, even though our memorized detection metric does not utilize the synthetic set, we are showing the nearest neighbors obtained from it, for consistency. }
    \label{fig:memorized}
\end{figure}
\clearpage 

\begin{figure}[h]
    \centering
    \includegraphics[keepaspectratio, width=1.0\textwidth, height=1.0\textheight]{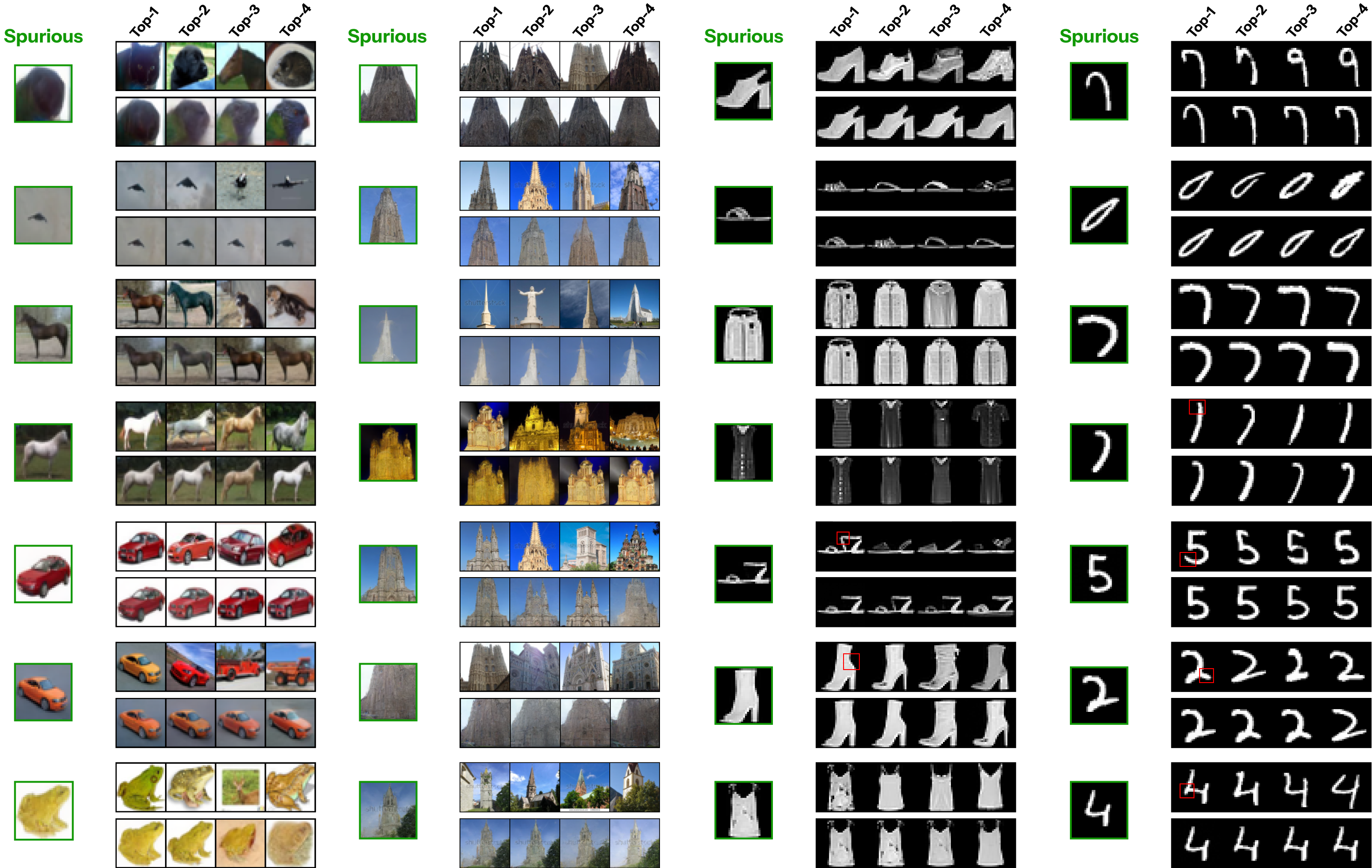}
    \caption{Visualizations of additional \textcolor[rgb]{0, 0.6, 0}{spurious} patterns and their {\textsf{top-4}} nearest neighbors for different datasets. The {\textsf{top row}} depicts nearest neighbors from the training set, while the {\textsf{bottom row}} shows those from the synthetic set. Spurious patterns are demonstrated to arise from the onset of generalization where the mixing of training data points begins. Since the model's generalization is at its infancy, duplicates of spurious patterns sometimes appear several times in the synthetic set $\mathsf{S}'$, much like \textcolor{red}{memorized} patterns. Additionally, these samples lack the uniqueness to be considered as \textcolor{blue}{generalized} samples as the model has yet to fully learn the underlying data distribution.}
    \label{fig:spurious}
\end{figure}
\clearpage 

\begin{figure}[h]
\centering
\includegraphics[keepaspectratio, width=1.0\textwidth, height=1.0\textheight]{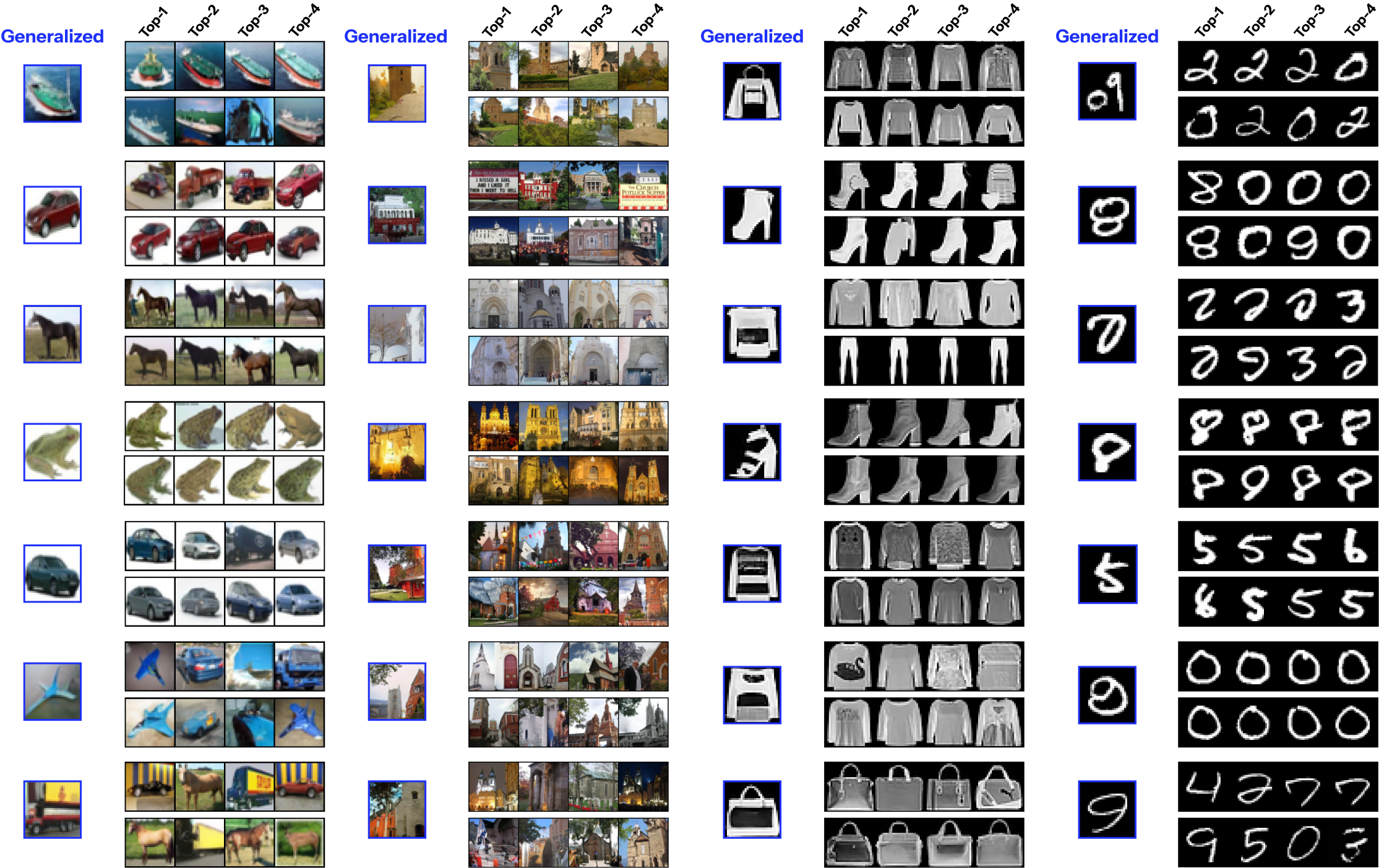}
\caption{Visualization of \textcolor{blue}{generalized} patterns and their {\textsf{top-4}} nearest neighbors for different datasets. The {\textsf{top row}} illustrates nearest neighbors from the training set $\mathsf{S}$ while the {\textsf{bottom row}} depicts those from the synthetic set $\mathsf{S}'$. Generalized samples are novel samples, which have little to no resemblance to their nearest neighbors in training and synthetic sets.}
\label{fig:generalized}
\end{figure}
\clearpage

\newpage 
\section{The Volume of the Basin of Attraction}
\label{sec:basin_of_attraction_extra_details}
\subsection{Algorithm Details}
To obtain the average log-volume results recorded in \cref{fig:transition-plots} of the main text, we first computed the critical time $t_c$ given an image $\hat{\rvx} = \rvx_0$. Specifically, this is done by using Alg.~(\ref{alg:critical_time}), where $t_c$ is obtained using the deterministic DDIM \cite{song2020denoising} sampler to denoise the perturbed version $\rvx_t$ of $\rvx_0$ for each step in $t = 1, \dots, T$, given $N'$ number of denoising steps starting backward from $t$ to $0$. To save computational time, the algorithm utilizes a stride constant $s$ to skip over some of the timesteps $t =1, 1 + s, 1 + 2s, \dots, T - s - 1$. For each timestep $t$, the variance preserving forward process is applied onto the targeted image $\rvx_0$ yielding $\rvx_t$. To account for the variations in noise, the image $\rvx_0$ is perturbed for $M$ number of times at each step $t$. 

To check whether if the critical time $t_c$ is reached, the algorithm utilizes a distance metric $d(\cdot, \cdot)$, which is LPIPS \cite{lpips}, equipped with the \texttt{vgg-16} backbone \cite{VGGsimonyan2014very} for all of the datasets. The distance metric $d(\cdot, \cdot)$ checks whether if a recovered image $\hat{\rvx}_0$, obtained from the DDIM sampler \cite{song2020denoising} with $N'$ denoising steps, is similar to the original pattern $\rvx_0$ within an error threshold $\delta_d \in (0, 1)$. Unlike the experiment on the memorization-generalization transition detailed in Sec.~(\ref{sec:transition-details}), we can afford to trade computational time for better accuracy in this experiment by using a larger backbone for LPIPS.  To keep track of the critical time $t_c$ for all perturbations, we initialized it as a vector of integers $t_c \in \mathbb{N}^{M}$, where $M$ is the number of perturbations or trials for the image $\rvx_0$. To ensure that $t_c$ is correct, we utilized a binary stopping vector $\hat{S} \in \{0, 1\}^M$, which keeps track of whether each perturbation of $\rvx_0$ is no longer recoverable, such that we only update an entry of the vector $t_c$ if and only if the corresponding entry in $\hat{S}$ is zero or not stopped. 

Since the entire process of finding $t_c$ is a rough search over a set of timesteps, we require a `breaking condition' for the search. In Alg.~(\ref{alg:critical_time}), this condition is determined by the value $p \in [0, 1]$, which is a probability corresponding to the recovery probability of the image $\rvx_0$ with respect to its $M$ perturbations. To compute $p$, we calculated the distance vector $\mathbf{d} = d(\hat{\rvx}_0, \rvx_0)$ where $\hat{\rvx}_0$ is the predicted version of $\rvx_0$ obtained from denoising $\rvx_t$. Then, we computed the recovery binary vector $\mathbf{r} \in [0, 1]^{M}$ by applying the indicator function onto $\mathbf{d}$ using the threshold $\delta_d$. Lastly, we calculated the probability value $p = \frac{1}{M} \sum^M_{i = 1} \mathbf{r}_i$. If this 
value $p$ is below the probability threshold $\delta_p$, we stop the search for critical time across all trials.

\begin{algorithm}[H]
\caption{Compute critical time $t_c$ for a pattern}
\label{alg:critical_time}

\DontPrintSemicolon

\SetKwProg{initialize}{Initialize}{}{end}
\SetKwProg{inputs}{Inputs}{}{end}
\SetKwProg{output}{Output}{}{end}
\SetKwProg{params}{Parameters}{}{end}
\SetKwProg{hyperparams}{HyperParameters}{}{end}
\SetKwProg{train}{Train}{}{end}
\SetKwProg{infer}{Compute $t_c$}{}{end}
\SetArgSty{textnormal}
\SetKwInOut{Input}{Input}
\SetKwInOut{Output}{Output}

\inputs{}{\vspace{1.5pt}
Pattern $\hat{\rvx}$\; \vspace{1.5pt}
Time stride $s$\; \vspace{1.5pt}
Score model $s_\theta$\; \vspace{1.5pt}
DDIM sampler $\Phi$\; \vspace{1.5pt}
Error threshold $\delta_d$\; \vspace{1.5pt}
Distance metric $d$\;  \vspace{1.5pt}
Denoising steps $N'$\; \vspace{1.5pt}
Total diffusion time $T$\;  \vspace{1.5pt}
Number of trials $M$\; \vspace{1.5pt}
Probability threshold $\delta_p$\; \vspace{1.5pt}
Forward process $\mathcal{F}$ 
}
\infer{}{%
\vspace{5pt}
$\rvx_0 \gets \text{duplicate}(\hat{\rvx}, M)$ \tcp*{\small Duplicate pattern for $M$ times} \vspace{5pt} 
$t_c \gets \mathbf{0}$ \tcp*{\small Initialize critical time tracker} \vspace{5pt}
$\hat{S} \gets \mathbf{0}$ \tcp*{\small Track trials that have stopped} \vspace{5pt} 
\For{$t = 1, 1+s, 1+2s, \dots, T - s - 1$}{\vspace{5pt}
    $\epsilon \sim \mathcal{N}(0, I)$ \tcp*{\small Sample noise} \vspace{5pt} 
    
    $\rvx_t \gets \mathcal{F}(\rvx_0, \epsilon, t)$ \tcp*{\small Forward process} \vspace{5pt} 
    
    $\hat{\rvx}_0 \gets \Phi(s_\theta, \rvx_t, t, N')$ \tcp*{\small DDIM denoising} \vspace{5pt} 
    
    $\mathbf{d} \gets d(\hat{\rvx}_0, \rvx_0)$ \tcp*{\small Compute distance per trial} \vspace{8pt} 
    
    $\mathbf{r}_i \gets 
        \begin{cases}
            1 & \text{if } \mathbf{d}_i \leq \delta_d \\
            0 & \text{otherwise}
        \end{cases}$ for $i = 1, \dots, M$ \tcp*{\small Binary recovery vector} \vspace{8pt} 
    
    $p \gets \frac{1}{M} \sum_{i=1}^M \mathbf{r}_i$ \tcp*{\small Recovery probability} \vspace{8pt} 
    
    \eIf{$p < \delta_p$}{
        \Return $t_c$ 
    }{
        $\hat{S} \gets \text{update}(\hat{S}, \mathbf{r})$ \;
        $t_c \gets \text{update}(t, t_c, \hat{S})$ 
    }
}
}
\output{}{\vspace{5pt}
Critical time $t_c \in \mathbb{N}^M$}
\end{algorithm}

\subsection{Experimental Details}
\begin{figure}[h]
\vspace{-5mm}
\centering  
\includegraphics[keepaspectratio, width=0.85\textwidth, height=1\textheight]{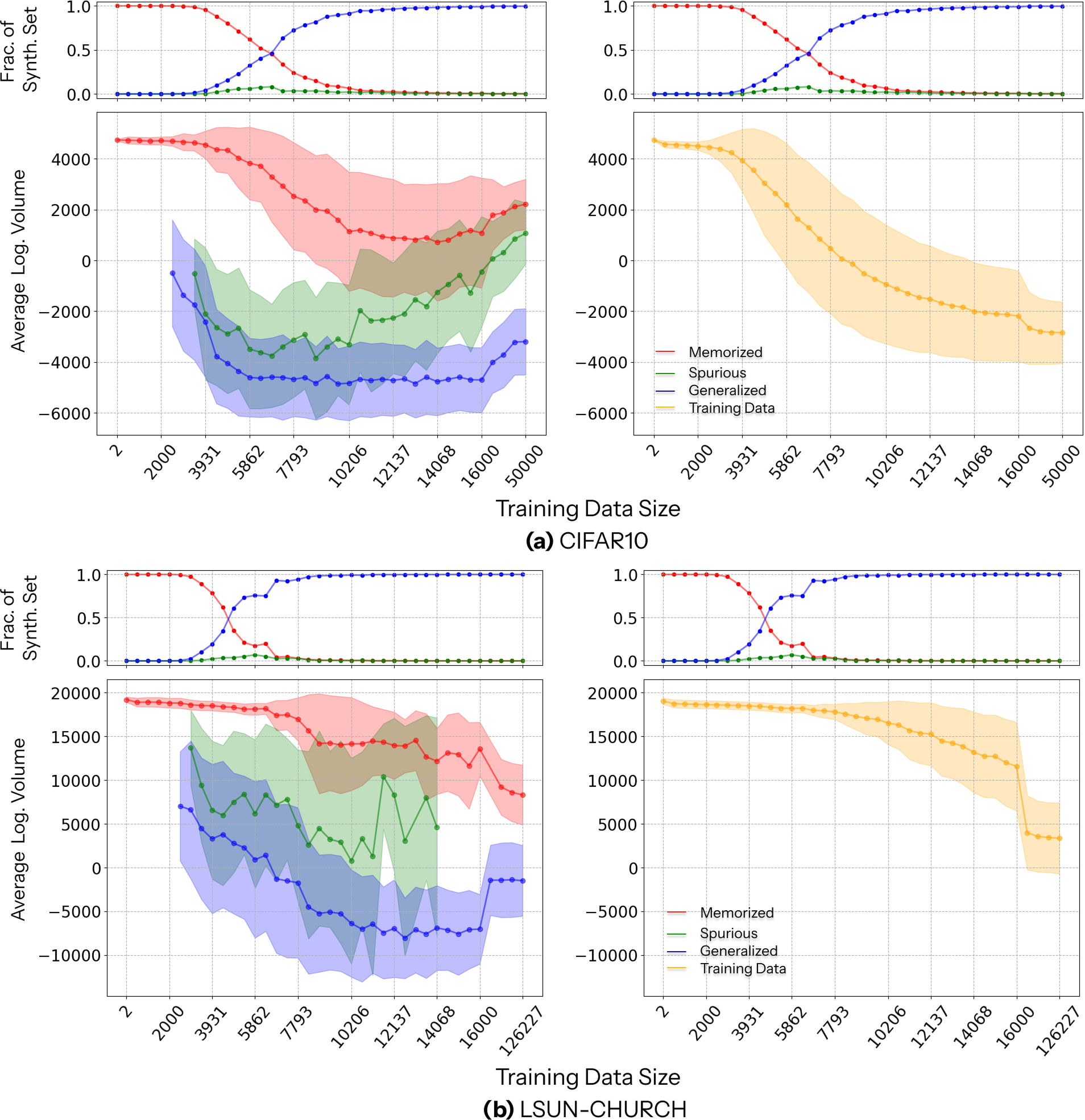}
\caption{
Average log-volume of the hyperspheric basin of attraction of different sample types as the training data size $K$ grows, in CIFAR10 \cite{cifar10} and LSUN-CHURCH \cite{lsun}  using Eq.~(\ref{eqn:volume-hypersphere}) with up to 512 samples per type. When fewer than 512 samples are available, all points are used. The average log-volume of training samples and the transition plot are included to confirm the shrinking basins of attraction in diffusion models. Variations in the critical time $t_c$ during the transition, which affect the radius $R$, lead to distinct log-volume trends across sample types, see the recorded critical times in Fig.~(\ref{fig:all-critical-times-p1}). Shaded regions indicate standard deviation of the log-volume.
}
\label{fig:volumes-1}
\end{figure}

\begin{figure}[!t]
\centering  
\includegraphics[keepaspectratio, width=0.85\textwidth, height=1\textheight]{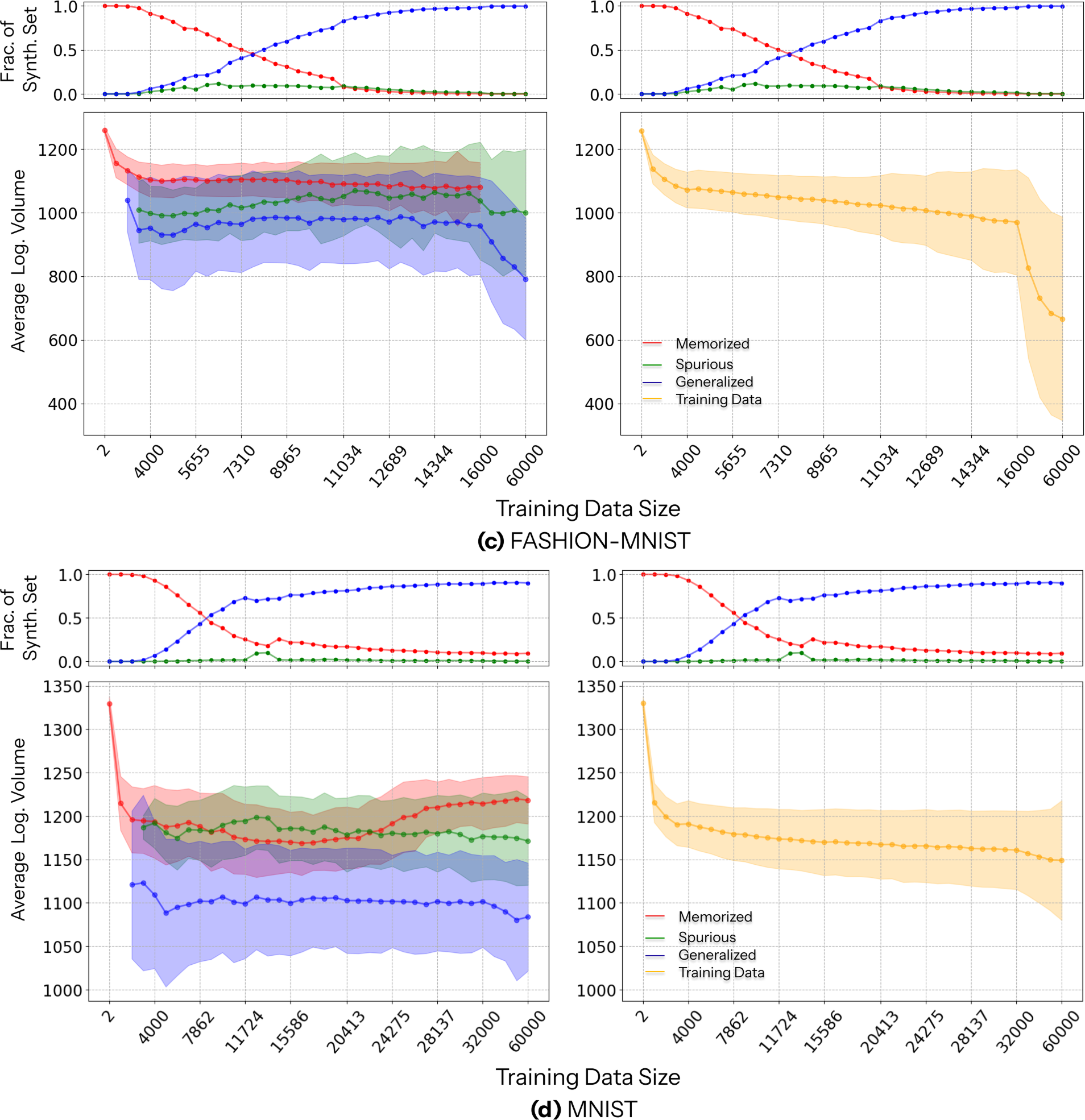} 
\caption{
Average log-volume of the hyperspheric basin of attraction of different sample types as the training data size $K$ grows, in FASHION-MNIST \cite{fmnist} and MNIST \cite{mnist}, computed using Eq.~(\ref{eqn:volume-hypersphere}) with up to 512 samples per type. When fewer than 512 samples are available, all points are used. The average log-volume of training samples and the transition plot are included to confirm the shrinking basins of attraction in diffusion models. Variations in critical time $t_c$ and radius $R$ lead to distinct log-volume trends across sample types, see the recorded critical times in Fig.~(\ref{fig:all-critical-times-p2}). Shaded regions indicate standard deviation of the log-volume.
}
\label{fig:volumes-2}
\end{figure}

For each training data size $K$, we used a maximum of the top 512 samples belonging to each of the sample types: training data, memorized, spurious, and generalized samples. Note, with the exception of the training data points, we sorted samples belonging to each set, identified via our detection metrics, by their distance from least to greatest. We computed the logarithmic version of the following $N$-ball volume equation:
\begin{equation}
    V(\hat{\rvx}, \rvx_{t_c}) = \frac{\pi^\frac{N}{2}}{\Gamma(\frac{N}{2} + 1)} R(\hat{\rvx}, \rvx_{t_c})^N
    \label{eqn:volume-hypersphere}
\end{equation}
using the $t_c$ we computed from Alg.~(\ref{alg:critical_time}) where $\Gamma(\cdot)$ is the Euler's gamma function \cite{sebah2002introductionGammaFunction} and $N$ is the dimensionality of the image. The radius $R(\hat{\rvx}, \rvx_{t_c})$ is the Euclidean norm of the difference between a given image $\hat{\rvx} \in \mathbb{R}^N$ and its recoverable perturbation $\rvx_{t_c}$ computed from the forward process. For Figs.~(\ref{fig:volumes-1})-(\ref{fig:volumes-2}), we plotted the average log-volume value calculated for each training data size $K$ per dataset. 

Meanwhile, for Alg.~(\ref{alg:critical_time}), we fixed the stopping probability $\delta_p$ as $0.8$, the number of perturbations $M$ as $20$, the DDIM steps $N'$ as $10$ and the stride constant $s$ as $10$ for all datasets. At some training data size $K$, in which there are very few samples in either memorized, spurious or generalized set, we opted to not include the statistics of such a set, partly due to its large variance. We remove the statistics of such a set if the number of its samples is less than $0.1\%$ of its corresponding synthetic set for a training data size $K$.

For CIFAR10 and LSUN-CHURCH, we utilized the threshold $\delta_d$ as $0.03$ and $0.12$ respectively. We kept these threshold values similar to the memorized threshold values $\delta_m$ detailed in Table~(\ref{tab:thresholds}), since they have proven to be effective at separating the memorized samples from non-memorized ones in the previous experiment. However, since it is difficult to select an appropriate threshold value for the $L_2$-distance metric (due to its range being between $0$ and $\infty$), we opted to use LPIPS to experiment with MNIST and FASHION-MNIST using the same threshold $\delta_d = 0.1$ for both. To find this optimal threshold value, we selected three data sizes, where each belongs to one of the three phases: memorization, spurious, or generalization. We then observed which of the several threshold values, starting from $0.01$ to $0.15$ with the increment of $0.01$, reflects best of the performance given by LPIPS using \texttt{vgg-16} \cite{VGGsimonyan2014very} as its backbone. Specifically, we evaluated these thresholds using $K \in \{ 4000, 11724, 60000 \}$ for MNIST, and $K \in \{ 4000, 8965, 60000\}$ for FASHION-MNIST. We simply performed Alg.~(\ref{alg:critical_time}) for those three data sizes on a small batch of $32$ samples for the three sample types, and manually checked whether the recovered samples (using the obtained critical time $t_c$ to denoise from) are similar to the original images and if this observation is consistent across those three data sizes. 

\newpage 
\subsection{Average Log-Volume Results}
\clearpage

\clearpage
\subsection{Average Critical Time Results}
\begin{figure}[h]
\vspace{-5mm}
\centering  
\setlength{\belowcaptionskip}{2pt} 
\includegraphics[keepaspectratio, width=0.825\textwidth, height=1\textheight]{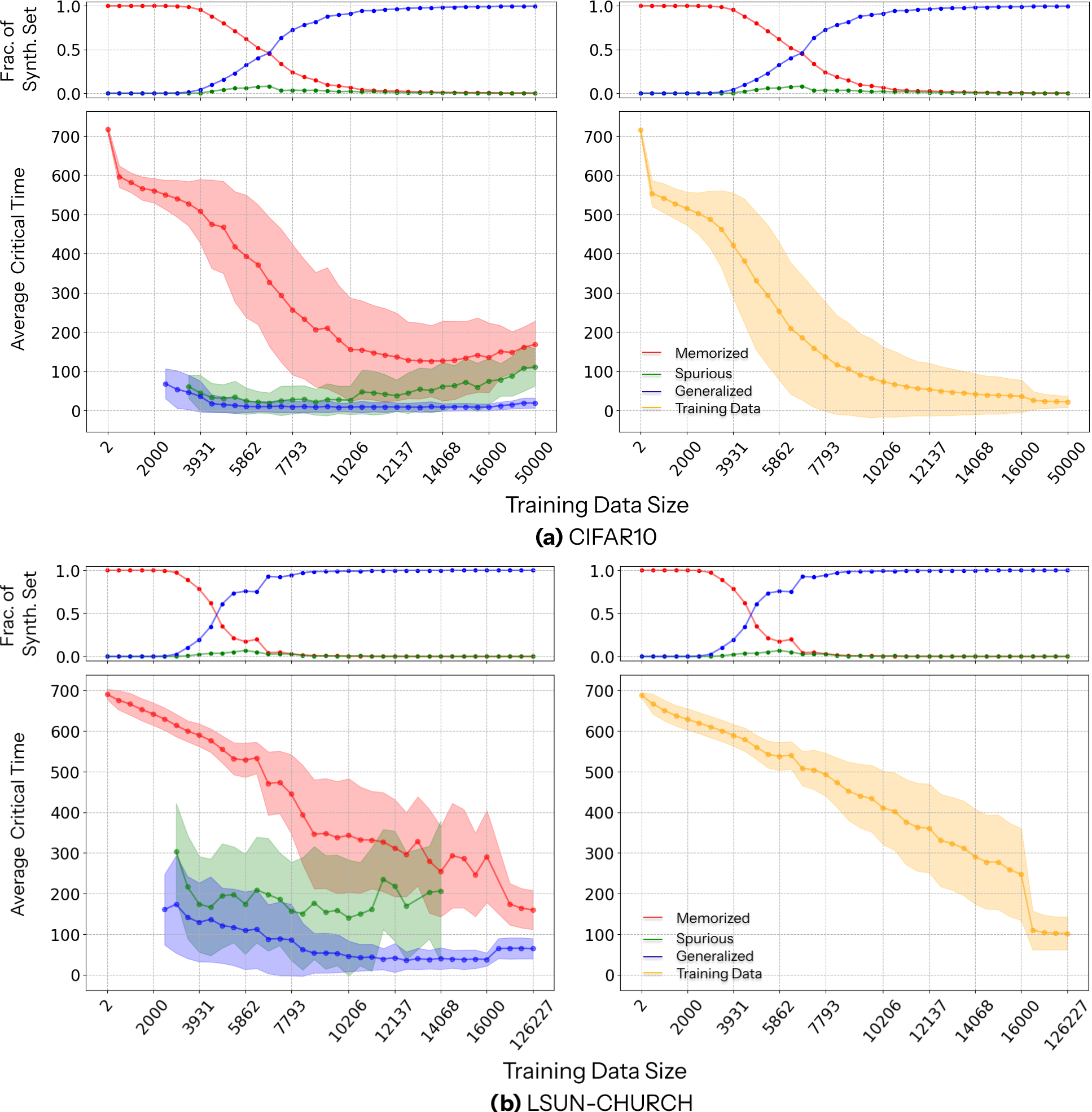}
\caption{Average critical time, obtained using Alg.~(\ref{alg:critical_time}), with a maximum of 512 samples across the memorization-generalization transition for each sample type in CIFAR10 \cite{cifar10} and LSUN-CHURCH \cite{lsun} The average critical times for the training data samples at each training data size $K$ are included to illustrate the overall decrease in the time and also the amount of perturbation in which the diffusion model can effectively recover the original sample within some error. Each sample type exhibits a distinctive critical time at each training data size $K$. The shaded region (or the error bars) shown here is the standard deviation computed from all of the critical times at each $K$ with respect to each type of samples. The transition plot of each dataset is shown here to visually guide where the memorization-generalization transition occurs.}
\label{fig:all-critical-times-p1}
\end{figure}
\begin{figure}[h]
\centering  
\setlength{\belowcaptionskip}{2pt} 
\includegraphics[keepaspectratio, width=0.85\textwidth, height=1\textheight]{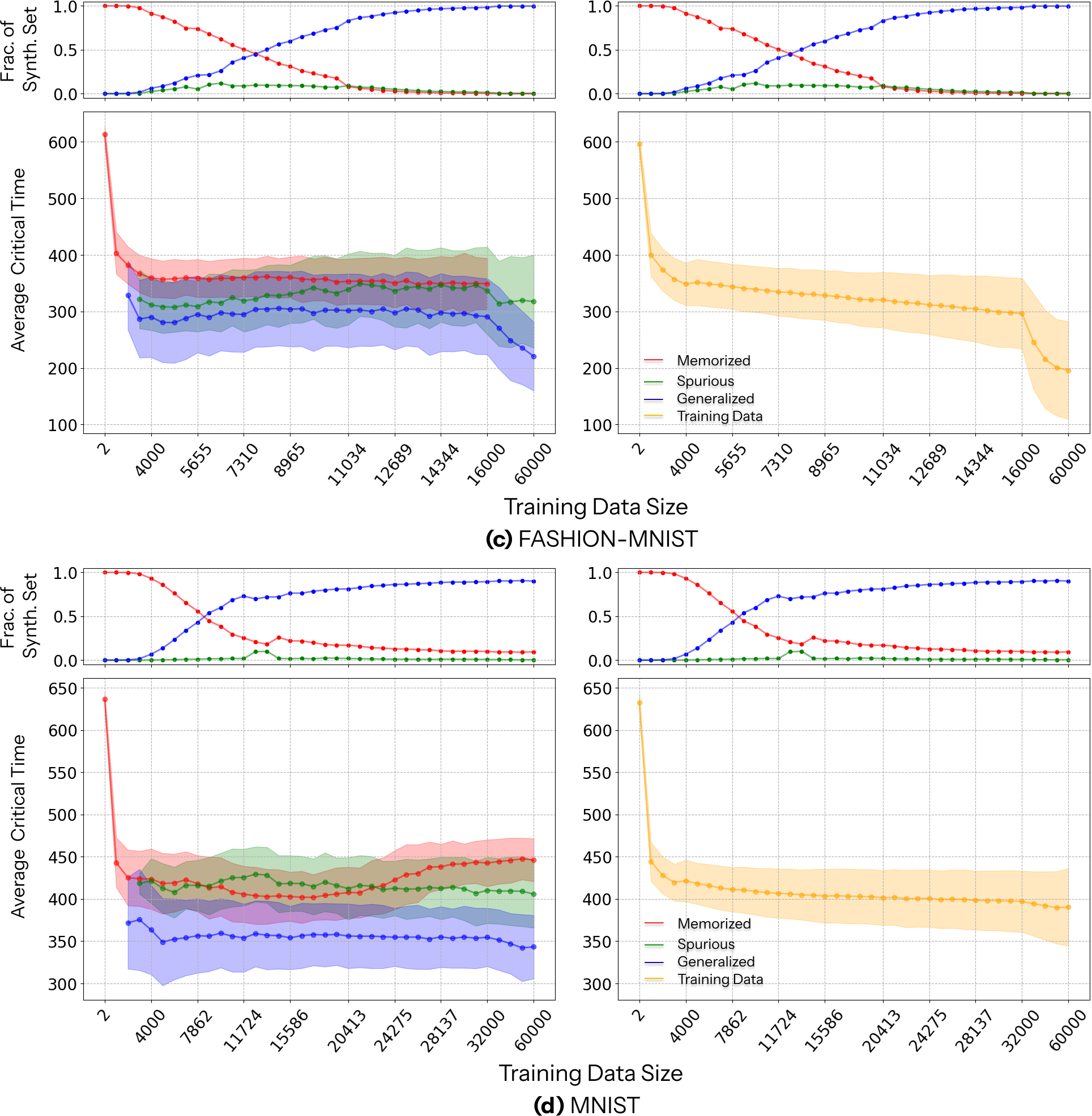}   
\caption{Average critical time, obtained using Alg. (\ref{alg:critical_time}), with a maximum of 512 samples across the memorization-generalization transition for each sample type in FASHION-MNIST \cite{fmnist} and MNIST \cite{mnist} The average critical times for the training data samples at each training data size $K$ are included to illustrate the overall decrease in the time and also the amount of perturbation in which the diffusion model can effectively recover the original sample within some error. Each sample type exhibits a distinctive critical time at each training data size $K$. The shaded region (or the error bars) shown here is the standard deviation computed from all of the critical times at each $K$ with respect to each type of samples. The transition plot of each dataset is shown here to visually guide where the memorization-generalization transition occurs.
}
\label{fig:all-critical-times-p2}
\end{figure}
\clearpage 

\newpage 
\section{The Curvature of the Energy}
\label{sec:energy-curvature}
\subsection{Overview}
Following \cite{achilli2024losing, ventura2024manifolds}, we utilized their stable version of algorithm for computing the singular values of the Jacobian of the score function, detailed in Alg.~(\ref{alg:central-diff}). The motivations for the algorithm are first detailed in \cite{stanczuk2022your} and later expanded by \cite{achilli2024losing, ventura2024manifolds}. Here we recite the motivations laid out in \cite{ventura2024manifolds}.

Consider a generative diffusion model with $p_0(\rvx)$ defined on a $d$-dimensional manifold $\mathcal{M}$, where $\mathcal{M} = \mathcal{M}_0$ denotes the manifold at $t = 0$. We can define a time-dependent set of points:
\begin{equation}
    \mathcal{M}_t = \{ \rvx^* | \tilde{s}_\mathcal{M} (\rvx^*, t) = 0, \,\,\text{with} \,\, J_\mathcal{M} (\rvx^*, t) \,\,\text{n.s.d.} \}
    \label{eqn:stable-latent-set}
\end{equation}
which \cite{ventura2024manifolds} name the \textbf{stable latent set} of the diffusion process. The negative semi-definiteness (n.s.d.) is a stability condition of the Jacobian matrix $J_\mathcal{M} (\rvx^*, t)$ of the support score $\tilde{s}_\mathcal{M} (\rvx^*, t)$, defined as the score function obtained from the uniform data distribution $\tilde{p}_0 (\rvx) = \frac{1}{|\mathcal{M}|} \delta_\mathcal{M} (\rvx)$. 

\begin{figure}[h]
    \centering
    \includegraphics[keepaspectratio, width=1\linewidth]{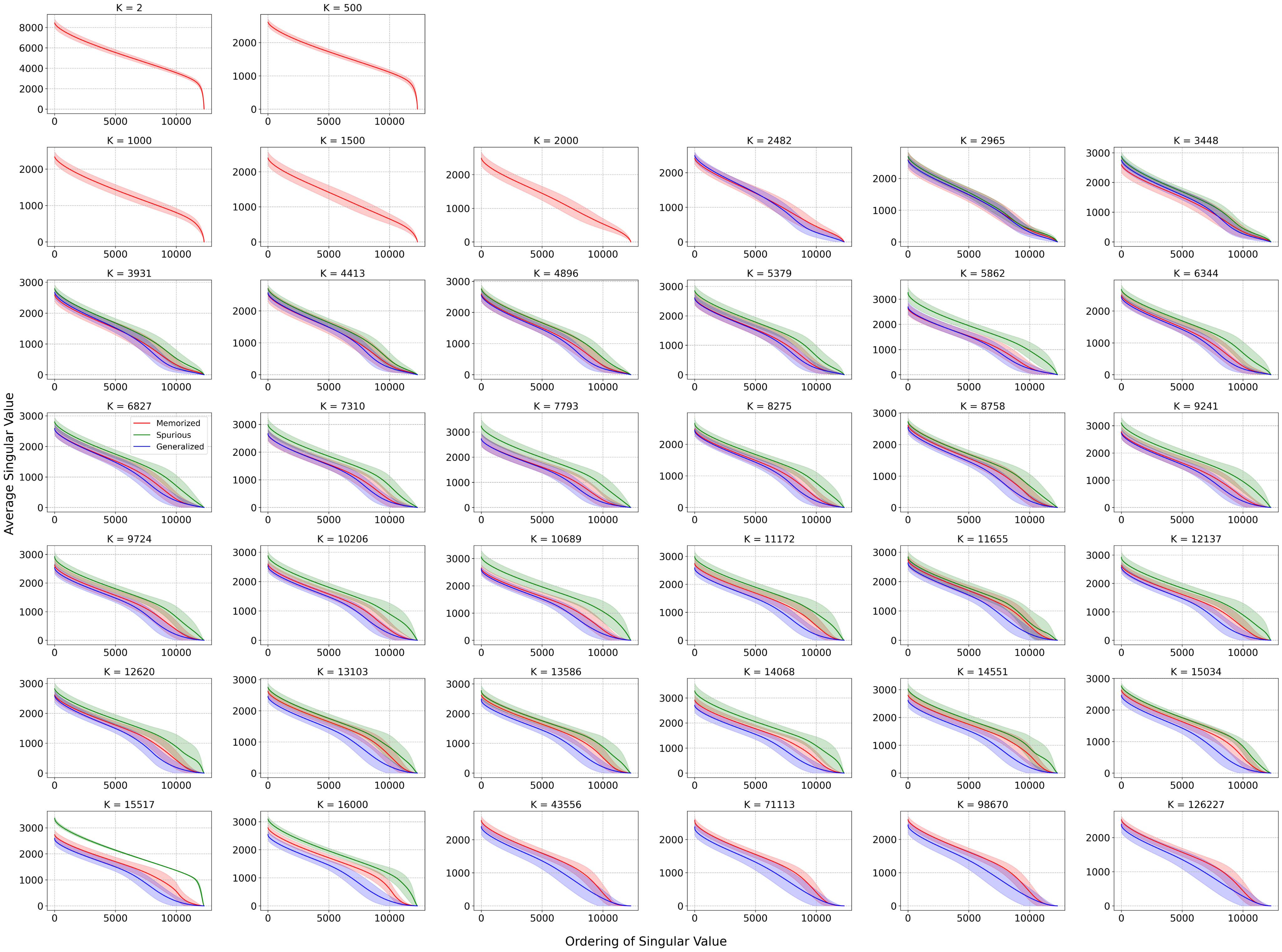}
    \caption{
    Average singular value spectra of the score's Jacobian for memorized, spurious, and generalized samples from diffusion models trained on the LSUN-CHURCH \cite{lsun} dataset, shown across 38 increasing training data sizes $K$. The plots reveal a clear geometric hierarchy in the intermediate training regime, where memorized samples consistently exhibit the highest curvature (largest singular values), followed by spurious, and finally generalized samples. As the model enters the full generalization phase at large $K$, the average spectra for all three sample types become very distinguishable. Shaded region is the standard deviation of the singular values of different samples of each category.
    }
    \label{fig:church-all-svs}
\end{figure}

\begin{figure}[h]
    \centering
    \includegraphics[keepaspectratio, width=1\linewidth]{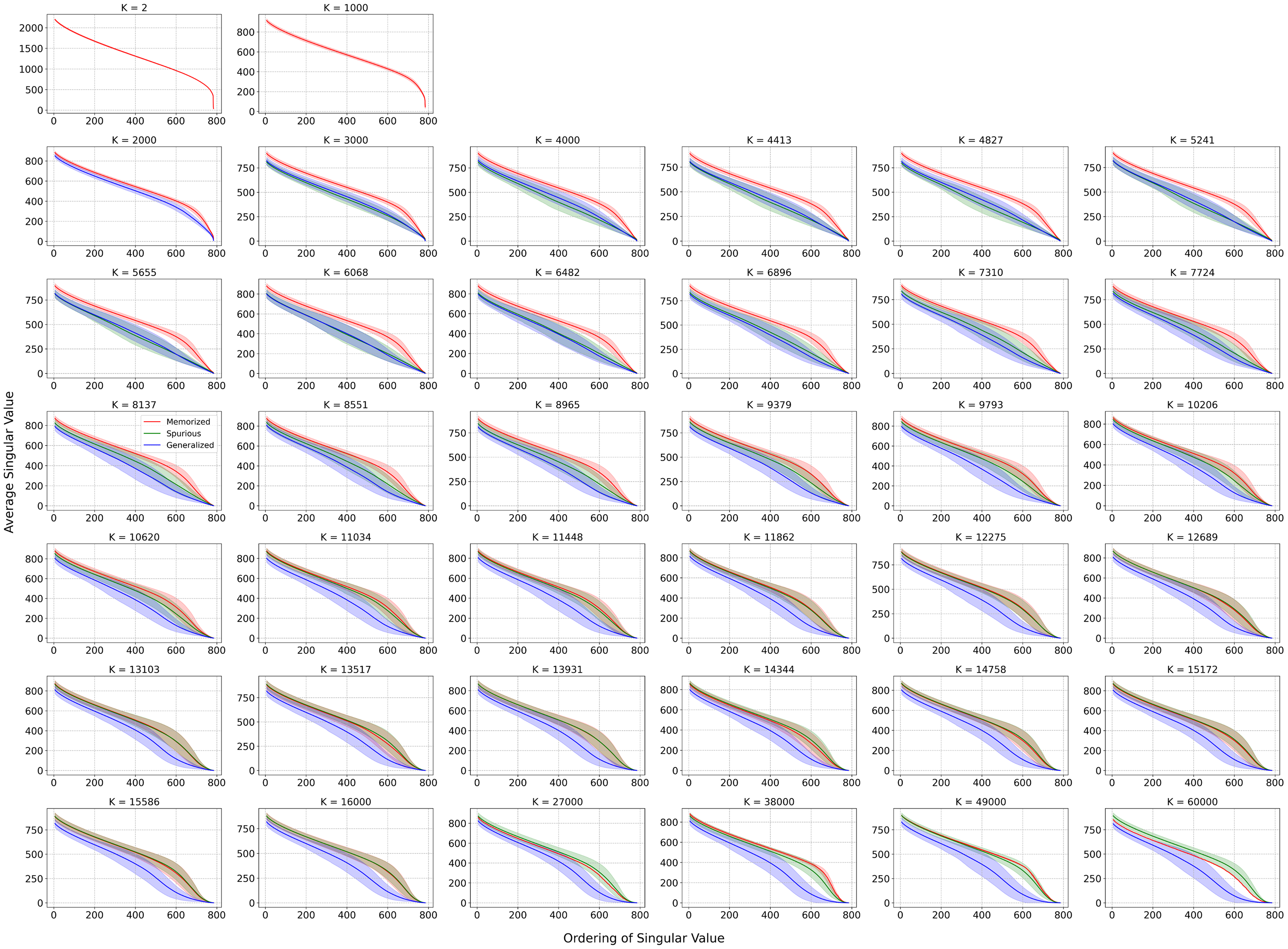}
    \caption{
    Average singular value spectra of the score's Jacobian for memorized, spurious, and generalized samples from diffusion models trained on the FASHION-MNIST \cite{fmnist} dataset, shown across 38 increasing training data sizes $K$. The plots reveal a clear geometric hierarchy in the intermediate training regime, where memorized samples consistently exhibit the highest curvature (largest singular values), followed by spurious, and finally generalized samples. As the model enters the full generalization phase at large $K$, the average spectra for all three sample types become very distinguishable. Shaded region is the standard deviation of the singular values of different samples of each category.
    }
    \label{fig:fmnist-all-svs}
\end{figure}

\begin{figure}[h]
    \centering
    \includegraphics[keepaspectratio, width=1\linewidth]{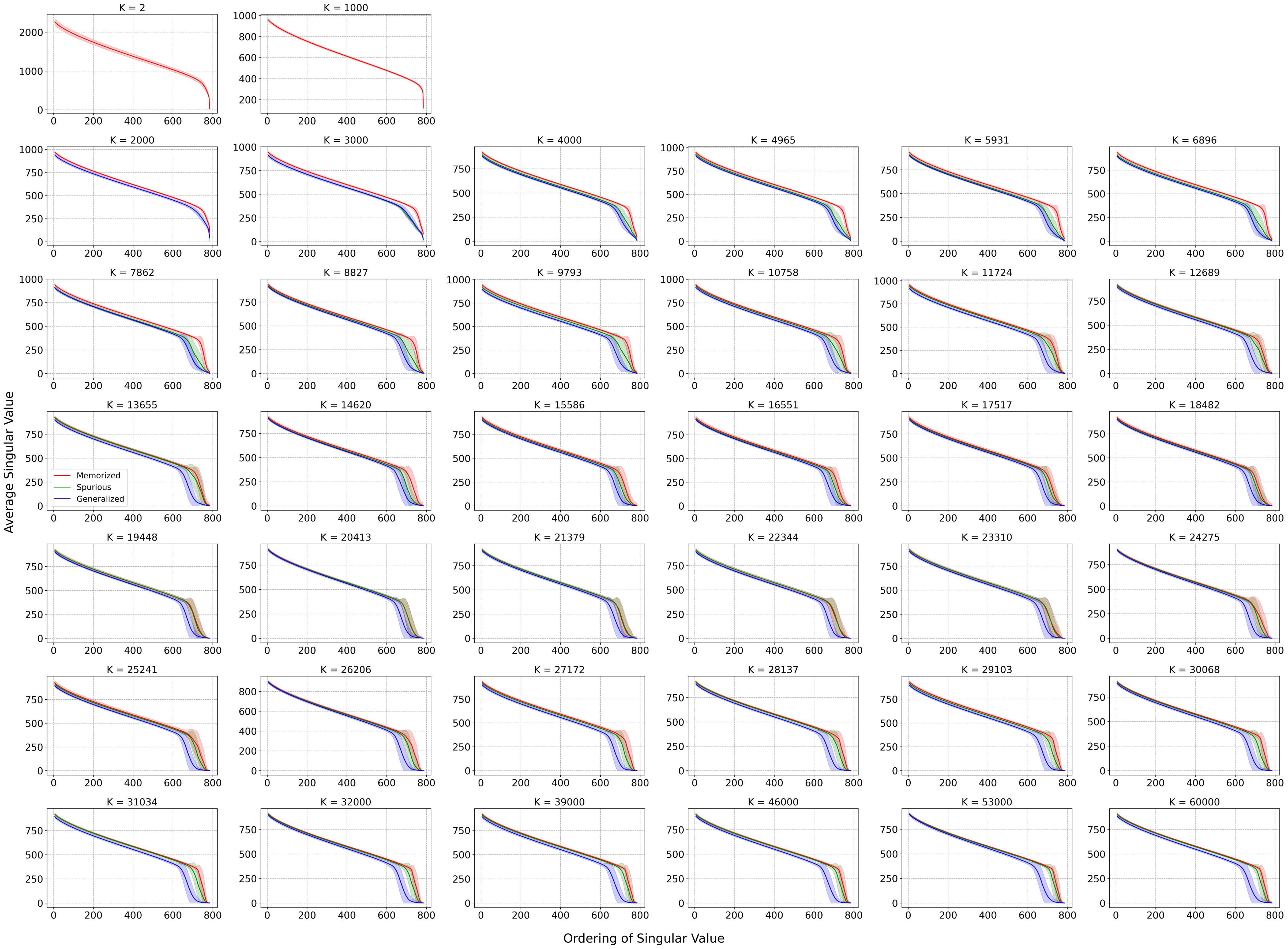}
    \caption{
    Average singular value spectra of the score's Jacobian for memorized, spurious, and generalized samples from diffusion models trained on the MNIST \cite{mnist} dataset, shown across 38 increasing training data sizes $K$. The plots reveal a clear geometric hierarchy in the intermediate training regime, where memorized samples consistently exhibit the highest curvature (largest singular values), followed by spurious, and finally generalized samples. As the model enters the full generalization phase at large $K$, the average spectra for all three sample types become very distinguishable. Shaded region is the standard deviation of the singular values of different samples of each category.
    }
    \label{fig:mnist-all-svs}
\end{figure}

\begin{figure}[h]
    \centering
    \includegraphics[keepaspectratio, width=1\linewidth]{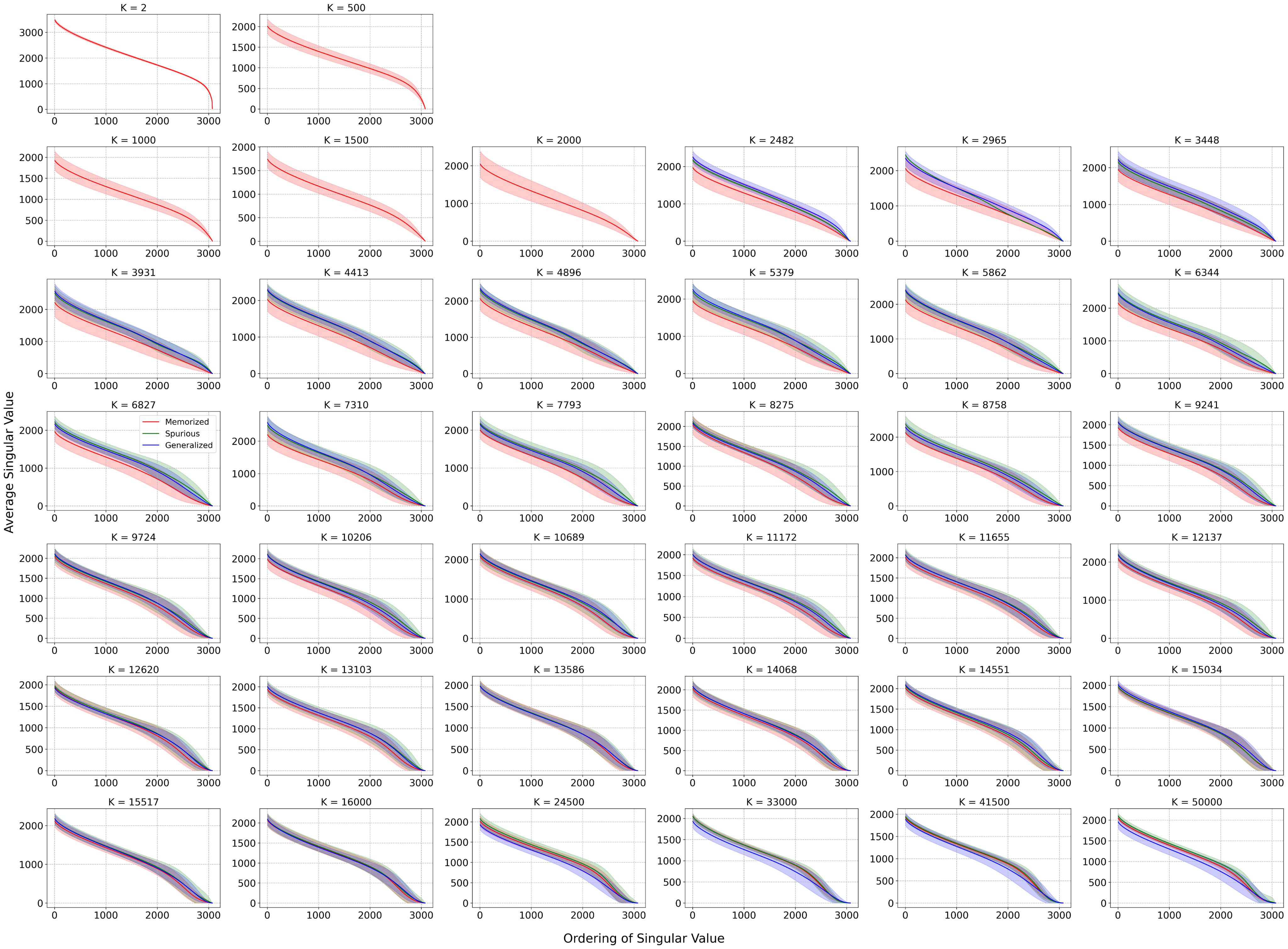}
    \caption{
    Average singular value spectra of the score's Jacobian for memorized, spurious, and generalized samples from diffusion models trained on the CIFAR10 \cite{cifar10} dataset, shown across 38 increasing training data sizes $K$. In contrast to the clear separation as seen in other datasets, the spectra for all three sample types remain heavily overlapped throughout the entire transition from memorization to generalization. However, as the model enters the full generalization phase at large $K$, the average spectra for all three sample types become less indistinguishable. Shaded region is the standard deviation of the singular values of different samples of each category.
    }
    \label{fig:cifar10-all-svs}
\end{figure}

Using the relationship $s_\theta(\rvx, t) = -\frac{\epsilon_\theta(\rvx, t)}{\sigma_t}$, the diffusing particles typically explore shells of a radius that concentrates on $\sigma_t$. Then, for a small perturbation $\rvp$ around a state $\rvx^*$ on the latent manifold $\mathcal{M}_t$, the score function is well approximated by linearization:
\begin{equation}
    s(\rvp, t) \approx J(\rvx^*, t) \rvp = \sum_j (\rvv_j \cdot \rvp) \lambda_j(\rvx^*, t) \rvv_j
\end{equation}
where $\rvv_j$ and $\lambda_j$ are the respective $j$-th eigenvector and eigenvalue of the negative Jacobian $-J(\rvx^*, t)$. Small perturbations, which aligned with the tangent space of $\mathcal{M}_t$, correspond to small eigenvalues; while orthogonal perturbations correspond to high eigenvalues, as the score tends to push the stochastic dynamics towards its fixed-points. Hence, based on these relationships, the spectrum of eigenvalues can provide detailed information regarding the local geometry of the stable latent set (\ref{eqn:stable-latent-set}). Moreover, as already indicated by \cite{stanczuk2022your}, one can estimate the dimensionality of the manifold from the location of a drop, or a sharp change, in the sorted spectrum of eigenvalues. The full range of the spectra results is shown in Figs.~(\ref{fig:church-all-svs})-(\ref{fig:cifar10-all-svs}). 

\subsection{Experimental Details}
Using Alg.~(\ref{alg:central-diff}), we computed the singular values of the Jacobian of the learnt score function $s_\theta$ given a small time $t_0$ for all datasets shown in Fig.~(5A) of the main text. For each sample type, we utilized a maximum of 512 samples and computed the singular values at $t_0 = 2$ for all datasets and training data sizes. The singular values are sorted from greatest to least, and the $\textsf{top-5}$ values are ignored to avoid the case of very large singular values skewing the visualizations. Meanwhile, for the spectra results of the training data points shown in the bottom row of Fig.~(\ref{fig:energy-curvature}A), we utilized $\min(K, 2048)$ number of training samples for all datasets and training data sizes. In this visualization, we utilized the rainbow color scheme and plotted the singular values for small training data sizes, starting from the purple color, to larger ones, ending at the red color.

\begin{figure}[h]
    \centering
    \includegraphics[keepaspectratio, width=1\linewidth]{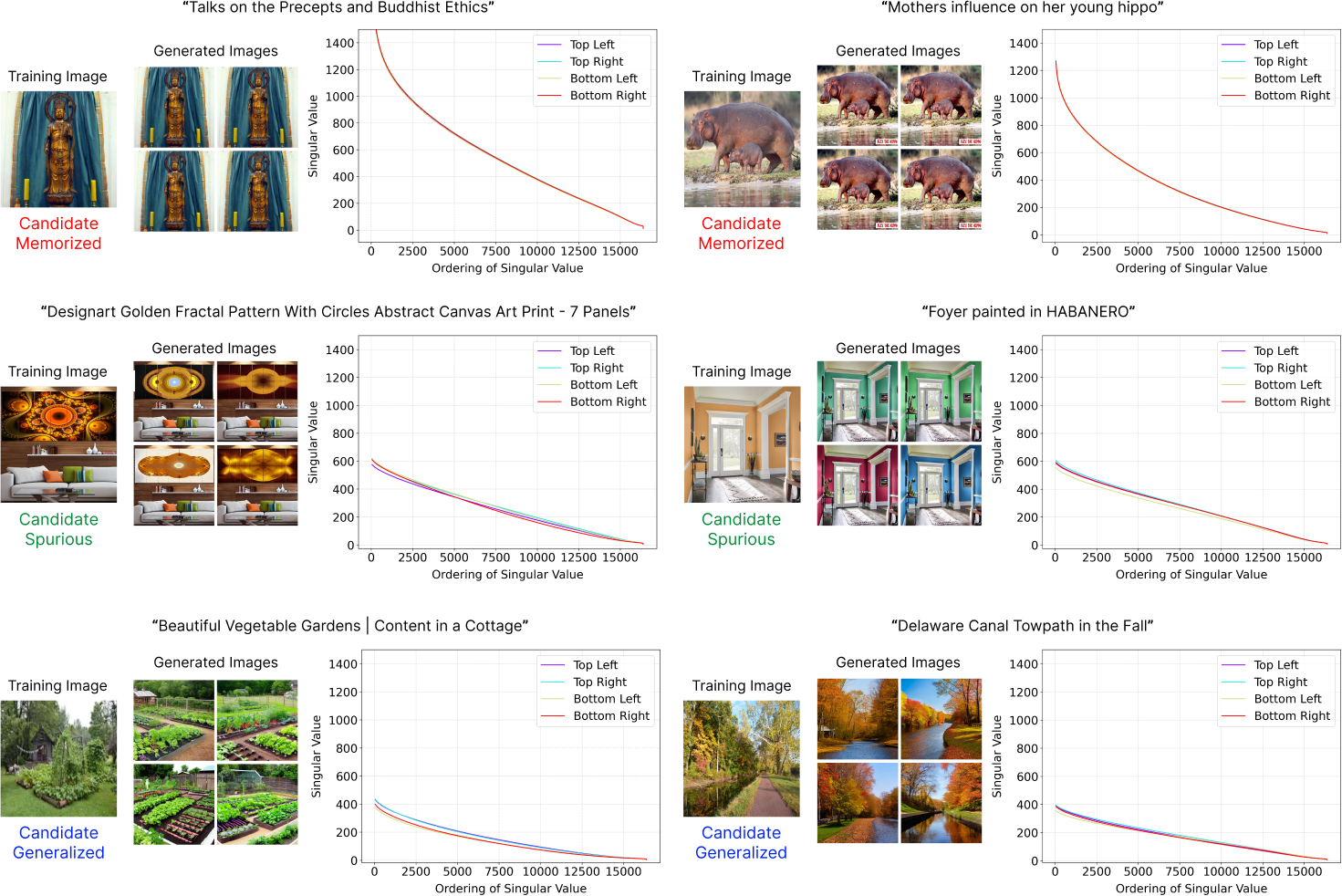}
    \caption{
    Additional examples of candidate memorized, spurious, and generalized samples from Stable Diffusion trained on LAION dataset \cite{schuhmann2022laion}, each corresponding to a distinct curvature signature. The candidate memorized sample has much larger singular values, while the candidate generalized sample has much smaller ones. The selected possible spurious samples have larger singular values than the candidate generalized sample, but much less in contrast to the candidate memorized sample. The common trait between candidate memorized and spurious samples is that both are stable attractors, as demonstrated by their repeated similar generations given different initial points (or noise vectors) and the conditioning on text-prompt. The y-axis is clipped at the value of 1500 to better contrast the shown examples' spectra.
    }
    \label{fig:stable-extra}
\end{figure}

Finally, regarding our experiment with Stable Diffusion \cite{rombach2022high}, following \cite{wen2024detecting, jeon2024understanding} we utilized its version 1.4 from the package Diffusers \cite{von-platen-etal-2022-diffusers}. The candidate memorized and spurious prompts we utilized are originally found by \cite{webster2023reproducible, webster2023duplication}. Additional examples are shown in Fig.~(\ref{fig:stable-extra}). Meanwhile, for candidate generalized samples, their prompts are obtained from LAION2B-Aesthetic \cite{schuhmann2022laion}. Instead of using Alg.~(\ref{alg:central-diff}), we opted for Alg.~(\ref{alg:non-central-diff}) and computed the singular values of the Jacobian of the score using $t_0 = 5$, due to the lack of computational resources. Please note that according to \cite{ventura2024manifolds} Alg.~(\ref{alg:central-diff}) produces a much smaller number of extreme singular values than the later algorithm. For generation of the samples, we utilized DDIM \cite{song2020denoising} sampler with different noise vectors initialized from a fixed seed. For the visualizations, we skipped the $\textsf{top-50}$ singular values and set the limit for the y-axis as 1500. 

\vspace{0.2in}

\begin{algorithm}[H]
\DontPrintSemicolon
\caption{Compute singular values of the Hessian of the energy using central difference for a pattern}
\label{alg:central-diff}
\SetKwProg{inputs}{Inputs}{}{end}
\SetKwProg{output}{Output}{}{end}
\SetKwProg{infer}{Compute matrix $\mathsf{M}$}{}{end}

\inputs{}{
Target sample $\rvx_0 \in \mathbb{R}^N$ \; \vspace{2.5pt} 
Time $t_0$\; \vspace{2.5pt}
Score model $s_\theta$ \; \vspace{2.5pt}
Forward process $\mathcal{F}$ \;
}
\BlankLine
$\mathsf{M} \gets \text{Empty}()$ \tcp*{\small Create an empty matrix}
\BlankLine
\infer{}{
\For{$i = 1, 2, \dots, 4N$}{\vspace{5pt}
    $\epsilon \sim \mathcal{N}(0, I)$  \tcp*{\small Sample noise} \;
    $\rvx^+ \gets \mathcal{F}(\rvx_0, \epsilon, t_0)$ \tcp*{\small Left perturbation} \; 
    $\rvx^- \gets \mathcal{F}(\rvx_0, -\epsilon, t_0)$\tcp*{\small Right perturbation} \; 
    $\mathsf{M}^{(i)} \gets \frac{s_\theta(\rvx^+, t_0) - s_\theta(\rvx^-, t_0)}{2}$ \tcp*{\small Add a new column to $\mathsf{M}$}
}
}
\output{}{
\vspace{5pt} 
$\{ s_i \}^{N}_{i = 1} \gets \text{SVD}(\mathsf{M})$ \tcp*{\small Compute singular values}
}
\end{algorithm}

\begin{algorithm}[H]
\DontPrintSemicolon
\caption{Compute singular values of the Hessian of the energy for a pattern}
\label{alg:non-central-diff}
\SetKwProg{inputs}{Inputs}{}{end}
\SetKwProg{output}{Output}{}{end}
\SetKwProg{infer}{Compute matrix $\mathsf{M}$}{}{end}

\inputs{}{
Target sample $\rvx_0 \in \mathbb{R}^N$ \; \vspace{2.5pt} 
Time $t_0$\; \vspace{2.5pt}
Score model $s_\theta$ \; \vspace{2.5pt}
Forward process $\mathcal{F}$ \;
}
\BlankLine
$\mathsf{M} \gets \text{Empty}()$ \tcp*{\small Create an empty matrix}
\BlankLine
\infer{}{
\For{$i = 1, 2, \dots, 4N$}{\vspace{5pt}
    $\epsilon \sim \mathcal{N}(0, I)$  \tcp*{\small Sample noise} \;
    $\rvx^+ \gets \mathcal{F}(\rvx_0, \epsilon, t_0)$ \tcp*{\small Perturbation} \; 
    $\mathsf{M}^{(i)} \gets s_\theta(\rvx^+, t_0)$\tcp*{\small Add a new column to $\mathsf{M}$}
}
}
\output{}{
\vspace{5pt} 
$\{ s_i \}^{N}_{i = 1} \gets \text{SVD}(\mathsf{M})$ \tcp*{\small Compute singular values}
}
\end{algorithm}

\clearpage 
\subsection{Curvature Results}

\clearpage 
\subsection{Additional Examples of Stable Diffusion}

\clearpage 
\section{Relative Energy Gap}
\label{sec:relative_energy_details}

\subsection{Overview}

In this section, we provide an additional experiment on the relative energy gap, which measures the gap between the energy depth of training data points and each of the three distinct sample types, using the method for approximating the relative potential from \cite{raya2024spontaneous}.\\

\noindent Consider the reverse SDE of the main text:
\begin{align}
    \mathrm{d} \rvx_t = [ \rvf(\rvx_t, t) - g(t)^2 \nabla_{\rvx_t} \log p_t (\rvx_t) ] \mathrm{d}t + g(t) \mathrm{d} \bar{\rvw}_t .
\end{align}
We can re-express the above equation with the potential function $u(x, t)$:
\begin{equation}
    -\nabla_{\rvx_t} u(\rvx_t, t) = \mathbf{f}(\rvx_t, t) - g(t)^2  \nabla_{\rvx_{t}}  \log p_t(\rvx_t) 
    \label{eqn:reverse-sde-reexpress}
\end{equation}
by using the drift term. By moving the minus sign and taking the integration of Eq.~(\ref{eqn:reverse-sde-reexpress}) with respect to $\rvx_t$, we have
\begin{equation}
    u(\rvx_t, t) = g(t)^2 \log p_t(\rvx_t) + \int^0_{\rvx_t} f(\rvz, t) d\rvz,
    \label{eqn:potential_energy_function_appendix}
\end{equation}
which is the potential function $u(\rvx_t, t)$ of the generative process. Substituting this potential function (\ref{eqn:potential_energy_function_appendix}) into the reverse SDE, we have the following generative process:
\begin{equation}
    \mathrm{d} \rvx_t = -\nabla_{\rvx_t} u(\rvx_t, t) \mathrm{d} t + g(t) \mathrm{d}\bar{\rvw}_t .
    \label{eqn:generative-process-potential}
\end{equation}

\noindent For our experimental setting, where we used the DDPM \cite{ho2020denoising} setting, we have the following variance preserving forward SDE:
\begin{equation}
    \mathrm{d} \rvx_t = -\frac{1}{2} \beta(t) \rvx_t \mathrm{d} t + \sqrt{\beta(t)} \mathrm{d} \rvw_t ,
\end{equation}
and the corresponding reverse (or generative) process SDE
\begin{equation}
    \mathrm{d} \rvx_t = \bigg [ -\frac{1}{2} \beta(t) \rvx_t - \beta(t)  \nabla_{\rvx_t}  \log p_t(\rvx_t) \bigg ] \mathrm{d} t + \sqrt{\beta(t)} \mathrm{d} \bar{\rvw}_t .
    \label{eqn:vp-reverse-sde}
\end{equation}
With Eq.~(\ref{eqn:generative-process-potential}), we have 
\begin{equation}
    \mathrm{d} \rvx_t = -\nabla_{\rvx_t} u(\rvx_t, t) \mathrm{d} t + \sqrt{\beta(t)} \mathrm{d} \bar{\rvw}_t    ,
\end{equation}
and by using Eq.~(\ref{eqn:reverse-sde-reexpress}), the potential for our experimental setting equates to
\begin{equation}
\begin{split}
    u(\rvx_t, t) &= g(t)^2 \log p_t(\rvx_t) + \int_{\rvx_t}^0 f(\rvz, t) d\rvz \\ 
    & = \beta(t) \log p_t (\rvx_t) + \frac{1}{2} \beta(t) \int_{\rvx_t}^0 \rvz_t \mathrm{d} \rvz \\ 
    &= \beta(t) \log p_t (\rvx_t) - \frac{1}{4} \beta(t) \rvx_t^2 . 
\end{split}  
\label{eqn:potential_function_vp}
\end{equation}

\begin{figure}[!t]
\centering
\includegraphics[keepaspectratio, width=0.9\textwidth, height=1.0\textheight]{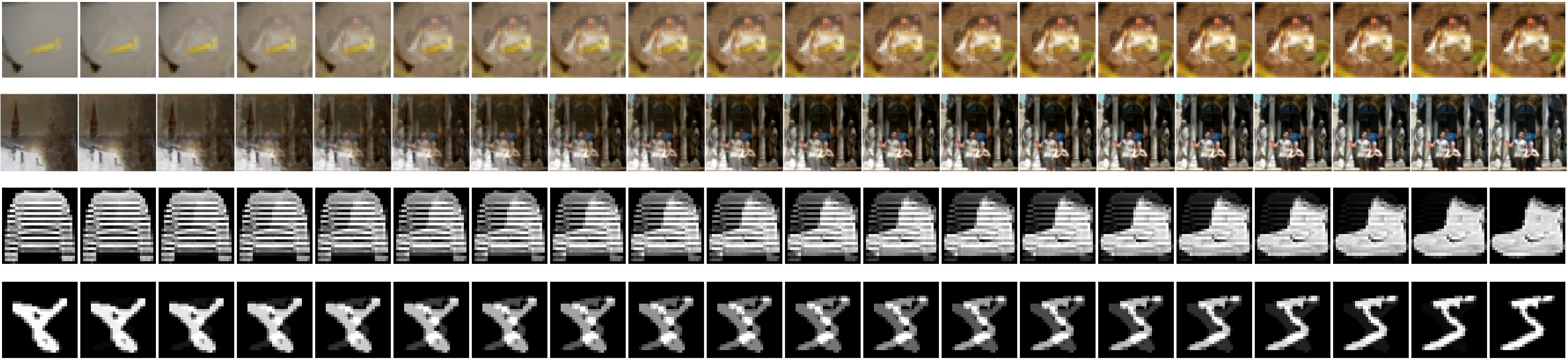}
\caption{Circular interpolation of a target image $\rvx_1$ and a reference image $\rvx_2$ over $\alpha \in \{0, \dots, \frac{\pi}{2}\}$ for different datasets.}

\label{fig:relative-energy-interpolation}

\end{figure}

Assume we have two images, $\rvx_1 (t)$ and $\rvx_2 (t)$, where $\rvx_1$ is a target image and $\rvx_2 (t)$ is a reference image at time $t$. Since we are interested in the energy of unperturbed images, let $\rvx_1 = \rvx_1(0)$ and $\rvx_2 = \rvx_2(0)$. As discussed in \cite{raya2024spontaneous}, the $N$-dimensional potential of a diffusion model can be projected down into 1-dimension via a generative trajectory between $\rvx_1$ and $\rvx_2$. We can obtain such a trajectory via circular interpolation between $\rvx_1$ and $\rvx_2$ following
\begin{equation}
     \tilde{\rvx}(\alpha) = \cos (\alpha) \rvx_1 + \sin (\alpha) \rvx_2 ,
     \label{eqn:circular_interpolation}
\end{equation}
and compute the relative potential (up to a constant) as a function of $\alpha$ by using 
\begin{equation}
    \tilde{u}(\alpha) = u(\tilde{\rvx} (\alpha)) = \int^{\pi / 2}_{0} \nabla u (\tilde{\rvx} (\alpha)) \cdot \rvv \, \mathrm{d} \alpha ,
    \label{eqn:relative-potential-int}
\end{equation}
where $\rvv = - \sin(a) \rvx_1 + \cos(a) \rvx_2$. See Fig.~(\ref{fig:relative-energy-interpolation}) for a visualization of this circular interpolation between two samples. For discrete times, we instead estimate the relative energy as 
\begin{equation}
    \tilde{u}(\bar{\alpha}_L) \approx \sum^{L}_{i = 1} \nabla u (\tilde{\rvx} (\bar{\alpha}_i)) \cdot \rvv \Delta \alpha 
    \label{eqn:relative-potential-discrete}
\end{equation}
given that $\bar{\alpha} = \{0, \dots, \frac{\pi}{2} \}$ and the potential $u(\cdot)$ is obtained from the above Eq.~(\ref{eqn:potential_function_vp}).

\subsection{Experimental Details}
In our experimentation, we utilized a maximum of the top 512 samples for each of three sample types, memorized, spurious, and generalized, to compute the relative energy following Eq.~(\ref{eqn:relative-potential-discrete}), at each training data size $K$. For training data points, we instead utilized $\min (K, 2048)$ number of samples. Note, with the exception of the training data set, the samples in the three sets are sorted from least to greatest, based on their distance corresponding to their respective detection metric (from the main text). We computed the relative energy of each set following Eq.~(\ref{alg:relative_energy}), which is strictly following the formulations detailed above. Note, we randomly selected and utilized an image from the entire training set as part of the relative energy calculation. 

To define the average relative energy gap, let $\tilde{u}_\text{target} \in \mathbb{R}^{M}$ be a set of $M$ relative potentials computed from using either a memorized, spurious, or generalized set and following Eq.~(\ref{eqn:relative-potential-discrete}). Please note that $M$ is capped at $512$ for the three sample types. Meanwhile, let $\tilde{u}_\text{data} \in \mathbb{R}^{M'}$ be the set of relative potentials computed from the training dataset at size $K$ where $M' = \min(K, 2048)$. The average relative energy gap is defined as 
\begin{equation}
\Delta (\tilde{u}_\text{target}, \tilde{u}_\text{data}) = \frac{1}{M} \sum^M_{i = 1} \big ( \tilde{u}_\text{target} \big )_i - \frac{1}{M'} \sum^{M'}_{j = 1} \big ( \tilde{u}_\text{data} \big )_j = \mu(\tilde{u}_\text{target}) - \mu(\tilde{u}_\text{data})
 \label{eqn:relative-energy-gap}
\end{equation}
Using Eq.~(\ref{eqn:relative-energy-gap}), we computed the average relative energy gap and recorded our results in Fig.~(\ref{fig:all-relative-energy}). The shaded regions reported in this figure represent the standard deviations computed from using $\tilde{u}_\text{target}$, and the average relative energy gap as the mean value. Since at some training data size $K$, there might be very few samples in one of the three sets, we opted not to include the relative energy gap of such sets, partly due to their large variance. Similar to the basin of attraction experiment, we determine whether a set is too small if the number of its samples is less than $0.1\%$ of its corresponding synthetic set. Finally, for our interpolation values $\bar{\alpha} = \{0, \dots, \frac{\pi}{2}\}$ used for Eq.~(\ref{eqn:relative-potential-discrete}), we generate $20$ points starting from 0 to $\frac{\pi}{2}$ using linear spacing. 

\subsection{Discussion}

\begin{figure}[h]
\vspace{-5mm}
\centering  
\setlength{\abovecaptionskip}{5pt}
\includegraphics[keepaspectratio, width=0.8\textwidth, height=1\textheight]{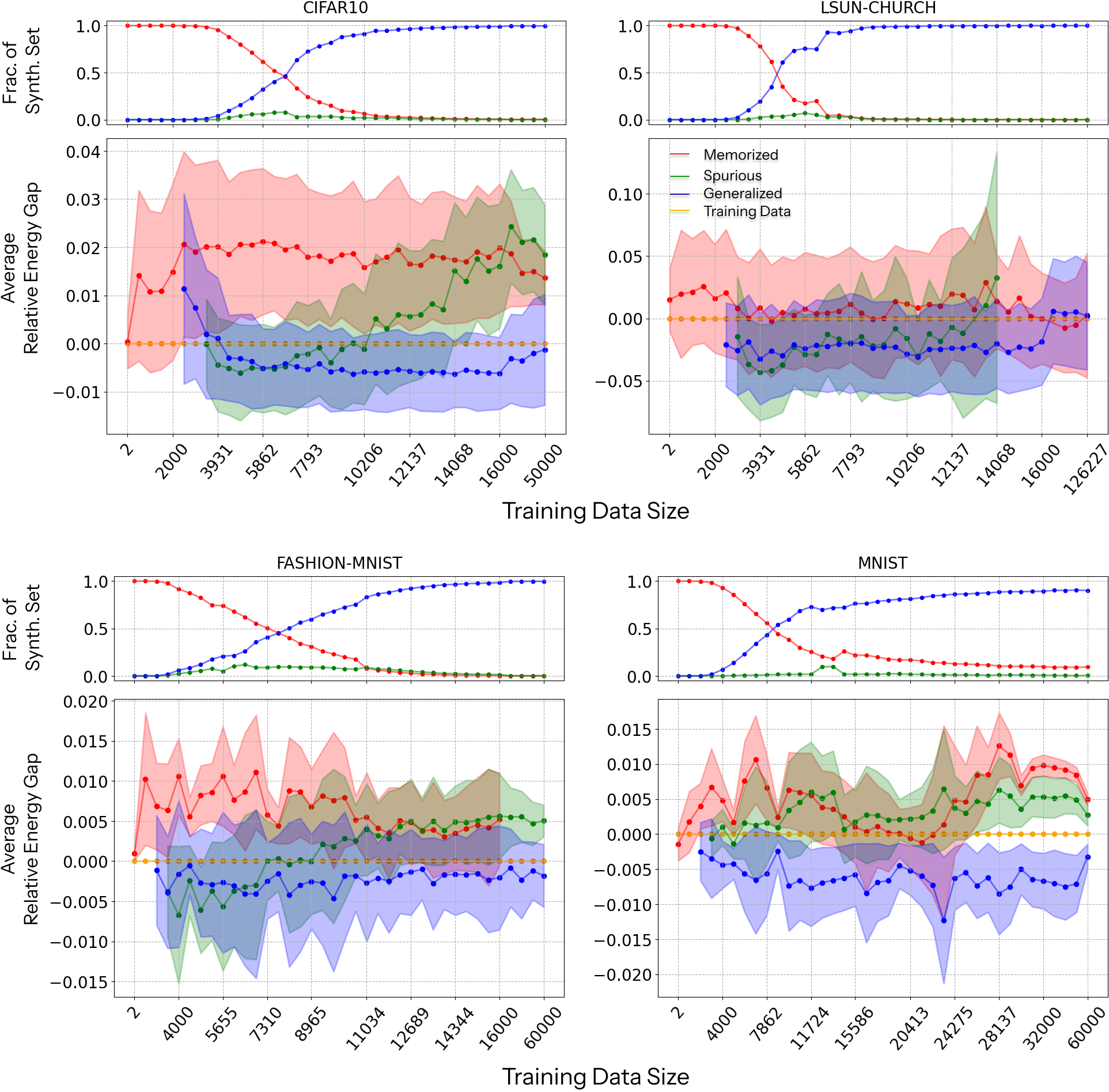}
\caption{
Average energy gap of each sample type and the training samples for the four datasets as the training data size $K$ grows, computed using Eq.~(\ref{eqn:relative-potential-discrete}). Gaps are measured relative to training samples, whose energy is set to zero since there is no gap to itself. A maximum of 512 samples per type is used at each $K$, or the full set is used instead if there are fewer than 512 points. Shaded regions show standard deviation of the gap values. With the exception of LSUN-CHURCH, the relative energy of the memorized samples matches that of the training data samples at $K = 2$, but quickly diverges as $K$ increases. Similarly, spurious samples' relative potential resembles that of the training data samples during the intermediate phase of the memorization-generalization phase. However, the potential quickly diverges as the later stages of generalization are initiated and consequently, the gap between generalized and training data samples' relative energy significantly decreases. 
}
\label{fig:all-relative-energy}
\end{figure}

\begin{figure}[h]
\vspace{-5mm}
\centering  
\setlength{\belowcaptionskip}{0pt} 
\includegraphics[keepaspectratio, width=0.8\textwidth, height=1\textheight]{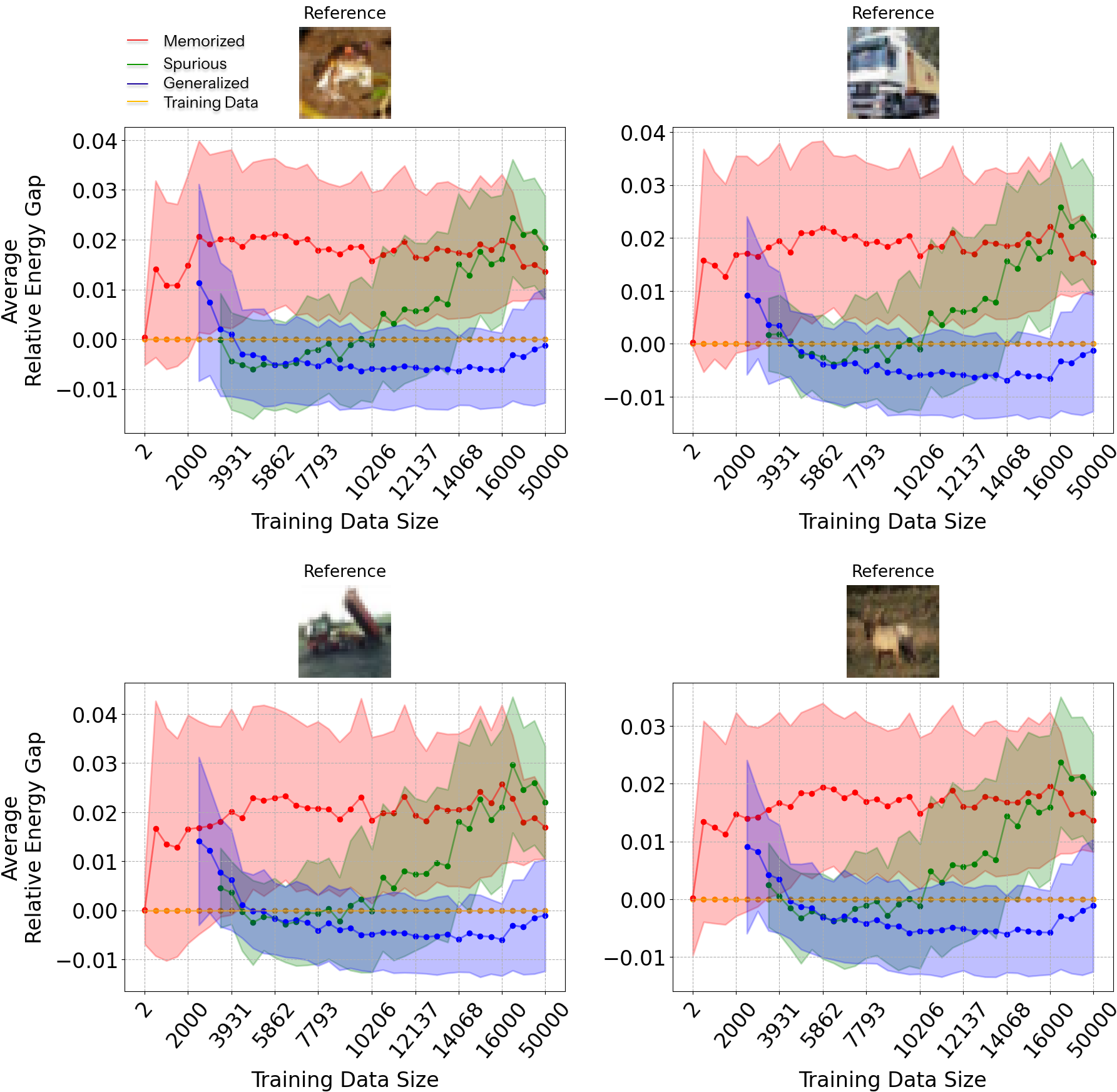}
\caption{
Average energy gap of each sample type and the training samples of CIFAR10 \cite{cifar10} as the training data size $K$ grows, computed using Eq.~(\ref{eqn:relative-potential-discrete}) and four different reference images. Gaps are measured relative to training samples, whose energy is set to zero since there is no gap to itself. A maximum of 512 samples per type is used at each $K$, or the full set is used instead if there are fewer than 512 points. Shaded regions show standard deviation of the gap values. The relative energy gap values are slightly different for distinct reference images but the general trend remains. The gap for memorized samples is initially close to zero and quickly diverges as the training data size $K$ increases. Similarly, spurious samples' relative potential resembles that of the training data samples during the intermediate phase of the memorization-generalization phase. However, the potential quickly diverges as the later stages of generalization are initiated and consequently, the gap between generalized and training data samples' relative energy significantly decreases. 
}
\label{fig:cifar10-top4-rel-energy}
\end{figure}

\begin{figure}[h]
\centering  
\setlength{\belowcaptionskip}{2pt} 
\includegraphics[keepaspectratio, width=0.89\textwidth, height=1\textheight]{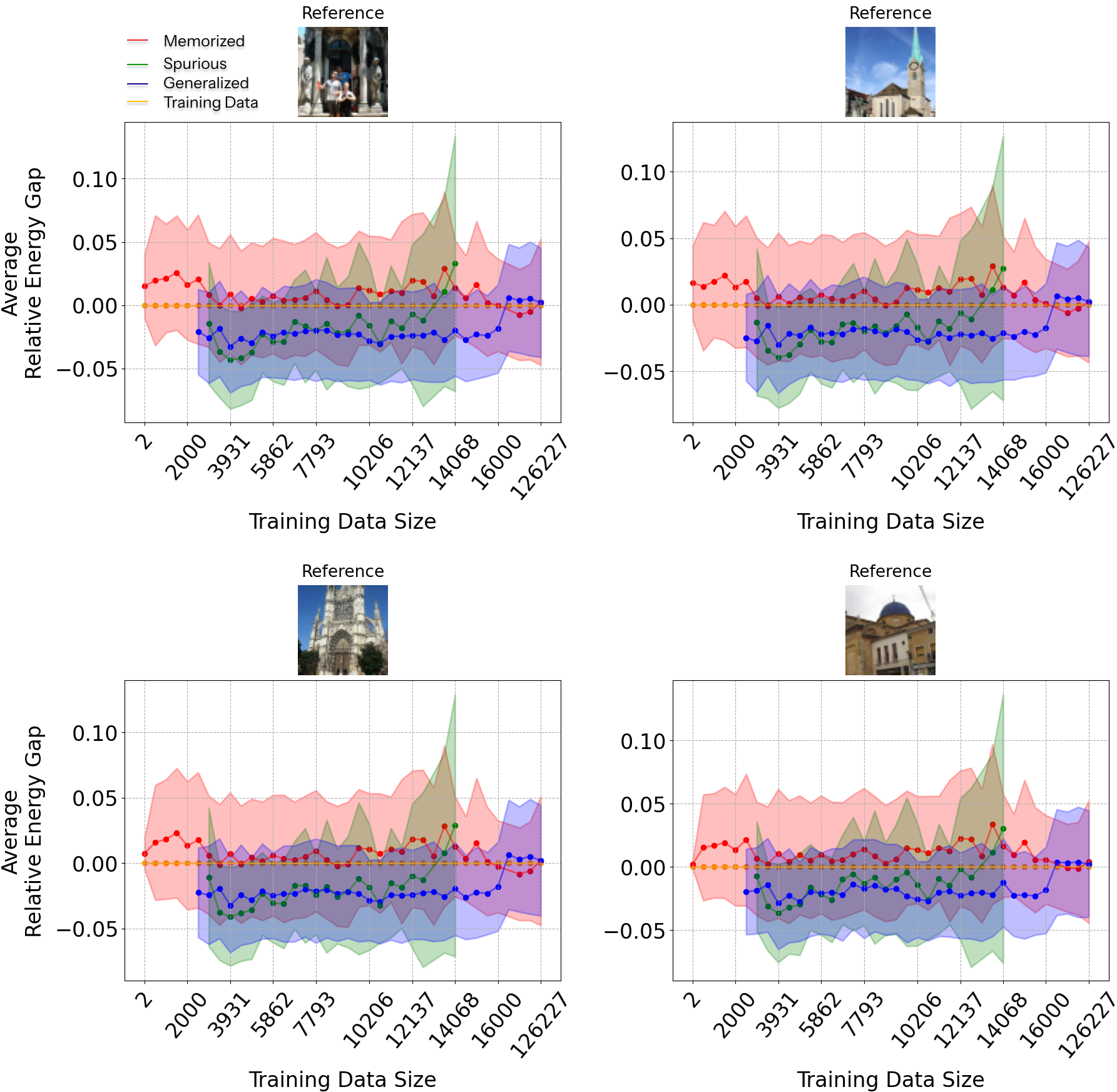}
\caption{
Average energy gap of each sample type and the training samples of LSUN-CHURCH \cite{lsun} as the training data size $K$ grows, computed using Eq.~(\ref{eqn:relative-potential-discrete}) and four different reference images. Gaps are measured relative to training samples, whose energy is set to zero since there is no gap to itself. A maximum of 512 samples per type is used at each $K$, or the full set is used instead if there are fewer than 512 points. Shaded regions show standard deviation of the gap values. The relative energy gap values are slightly different for distinct reference images but the general trend remains. The gap for memorized samples is initially close to zero and quickly diverges as the training data size $K$ increases. Similarly, spurious samples' relative potential resembles that of the training data samples during the intermediate phase of the memorization-generalization phase. However, the potential quickly diverges as the later stages of generalization are initiated and consequently, the gap between generalized and training data samples' relative energy significantly decreases. 
}
\label{fig:church-top4-rel-energy}
\end{figure}

\begin{figure}[h]
\centering  
\setlength{\belowcaptionskip}{2pt} 
\includegraphics[keepaspectratio, width=0.89\textwidth, height=1\textheight]{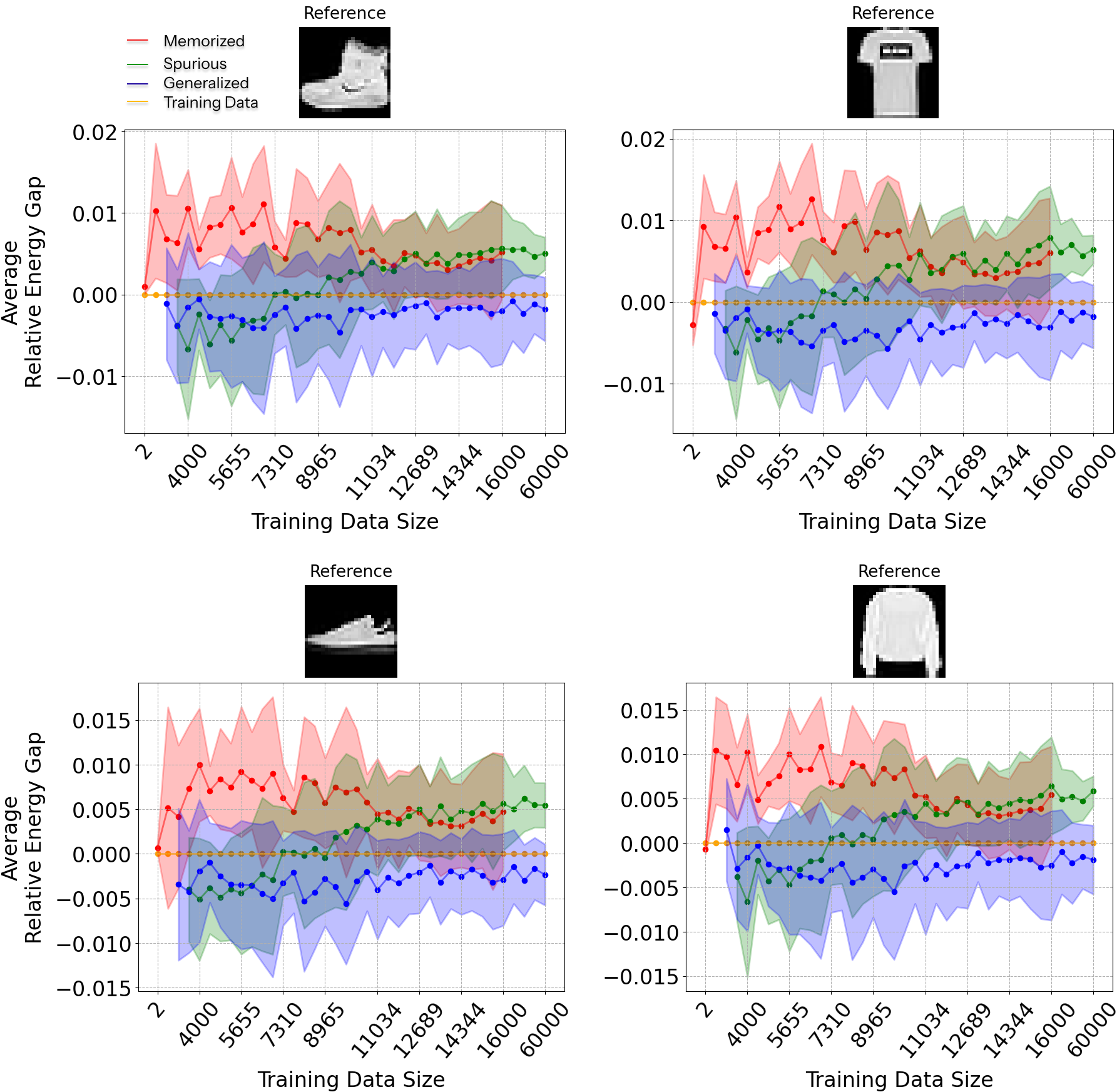}
\caption{
Average energy gap of each sample type and the training samples of FASHION-MNIST \cite{fmnist}  the training data size $K$ grows, computed using Eq.~(\ref{eqn:relative-potential-discrete}) and four different reference images. Gaps are measured relative to training samples, whose energy is set to zero since there is no gap to itself. A maximum of 512 samples per type is used at each $K$, or the full set is used instead if there are fewer than 512 points. Shaded regions show standard deviation of the gap values. The relative energy gap values are slightly different for distinct reference images but the general trend remains. The gap for memorized samples is initially close to zero (for three of the reference images) and quickly diverges as the training data size $K$ increases. Similarly, spurious samples' relative potential resembles that of the training data samples during the intermediate phase of the memorization-generalization phase. However, the potential quickly diverges as the later stages of generalization are initiated and consequently, the gap between generalized and training data samples' relative energy significantly decreases. 
}
\label{fig:fmnist-top4-rel-energy}
\end{figure}

\begin{figure}[h]
\centering  
\setlength{\belowcaptionskip}{2pt} 
\includegraphics[keepaspectratio, width=0.89\textwidth, height=1\textheight]{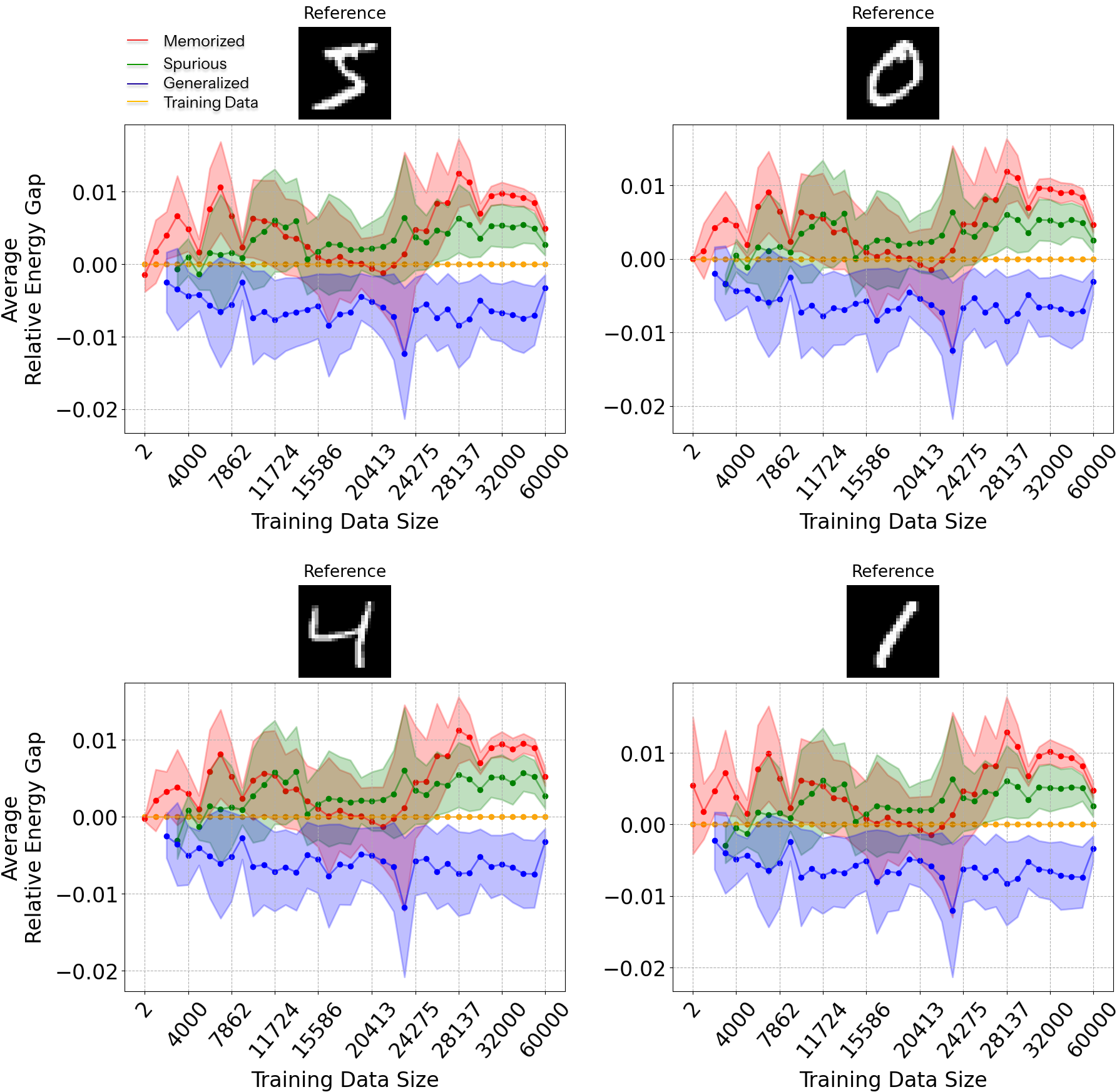}
\caption{
Average energy gap of each sample type and the training samples of MNIST \cite{mnist} as the training data size $K$ grows, computed using Eq.~(\ref{eqn:relative-potential-discrete}) and four different reference images. Gaps are measured relative to training samples, whose energy is set to zero since there is no gap to itself. A maximum of 512 samples per type is used at each $K$, or the full set is used instead if there are fewer than 512 points. Shaded regions show standard deviation of the gap values. The relative energy gap values are slightly different for distinct reference images but the general trend remains. The gap for memorized samples is initially close to zero (for three of the reference images) and quickly diverges as the training data size $K$ increases. Similarly, spurious samples' relative potential resembles that of the training data samples during the intermediate phase of the memorization-generalization phase. However, the potential quickly diverges as the later stages of generalization are initiated and consequently, the gap between generalized and training data samples' relative energy significantly decreases. 
}
\label{fig:mnist-top4-rel-energy}
\end{figure}

Since the relative energy computation requires a reference image, we also studied the effects of having different reference images on the relative energy gap in Figs.~(\ref{fig:cifar10-top4-rel-energy})-(\ref{fig:mnist-top4-rel-energy}). Overall, the results for the relative energy gap across the memorization-generalization transition for all datasets are shown in Fig.~(\ref{fig:all-relative-energy}). As the training data size $K$ increases, the gap for memorized samples is initially close to zero and quickly diverges. Spurious samples' relative potential resembles that of the training data samples during the intermediate phase of the memorization-generalization phase, where their potential is the lowest among the three sample types. However, the potential of spurious states quickly diverges as the later stages of generalization are initiated, and consequently, the gap between generalized and training data samples' relative energies became the most narrowed among the three distinct sample types -- further emphasizing the changing in the energy landscape as the DM undergoes its transition from memorization to generalization as $K$ increases.

\clearpage 
\begin{algorithm}[H]
\DontPrintSemicolon
\caption{Compute the discretized relative energy between a target and a reference sample}
\label{alg:relative_energy}

\SetKwProg{inputs}{Inputs}{}{end}
\SetKwProg{output}{Output}{}{end}
\SetKwProg{infer}{Compute $\tilde{u}(\bar{\alpha}_L)$}{}{end}

\inputs{}{
Target pattern $\rvx_1 \in \mathbb{R}^N$ \; \vspace{2.5pt} 
Reference pattern $\rvx_2 \in \mathbb{R}^N$\; \vspace{2.5pt}
Score model $s_\theta$\; \vspace{2.5pt}
Interpolation values $\bar{\alpha}$\; \vspace{2.5pt}
Initial variance $\beta_0$ 
}
\BlankLine
\textbf{Assumption:} $\rvx_1$ and $\rvx_2$ are samples at $t = 0$\;\;
\BlankLine
\infer{}{
\vspace{5pt}
$\bar{\alpha} \gets \{ 0, \cdots, \frac{\pi}{2} \}$ \tcp*{\small Linear Spacing from 0 to $\frac{\pi}{2}$} \vspace{5pt}

$\Delta \alpha \gets \bar{\alpha}_1 - \bar{\alpha}_0$ \; \vspace{5pt} 

$\tilde{u}(\bar{\alpha}_L) \gets 0$ \; \vspace{8pt}

\For{$\alpha \in \bar{\alpha}$}{\vspace{5pt}
    $\tilde{\rvx} \gets \cos (\alpha) \rvx_1 + \sin (\alpha) \rvx_2 $\; \vspace{5pt}
    $\tilde{\rvv} \gets - \sin(a) \rvx_1 + \cos(a) \rvx_2 $ \; \vspace{5pt}
    $\nabla_{\tilde{\rvx}} \log p_0(\tilde{\rvx}) \gets -s_\theta(\tilde{\rvx}, 0) / \sqrt{\beta_0}$\; \vspace{8pt}
    $\nabla_{\tilde{\rvx}} u(\tilde{\rvx}) \cdot \rvv \, \Delta \alpha \gets \big( -\frac{1}{2} \beta_0 \tilde{\rvx} - \beta_0 \nabla_{\tilde{\rvx}} \log p_0(\tilde{\rvx}) \big) \cdot \rvv \, \Delta \alpha$  \tcp*{\small Eq.~(\ref{eqn:vp-reverse-sde})}\; 
    $\tilde{u}(\alpha) \gets \sum^N_{j = 1} \bigg ( \nabla_{\tilde{\rvx}} u(\tilde{\rvx}) \cdot \rvv \, \Delta \alpha \bigg )_j$ \tcp*{\small Sum over $N$} \;
    $\tilde{u}(\bar{\alpha}_L) \gets \tilde{u}(\bar{\alpha}_L) + \tilde{u}(\alpha)$ \tcp*{\small Eq.~(\ref{eqn:relative-potential-discrete})}
}
}
\output{}{
\vspace{5pt} 
Relative energy $\tilde{u}(\bar{\alpha}_L)$}
\end{algorithm}
\clearpage 

\clearpage 
\subsection{Average Relative Energy Gap Results}
\clearpage

\subsection{Average Relative Energy Gap Results with Different References}

\clearpage 
\section{Hardware Details}
The training of the models detailed in Table~(\ref{tab:models}) was done using NVIDIA Tesla V100 GPUs, where a single GPU was used to train each model. Each GPU has 32GB of memory and is linked with Power9 processors, clocking at 3.15 GHz maximum. Each CIFAR10 model took roughly 18, 20, and 22 hours to train for different initial latent dimensions, e.g., 64, 96, 128, respectively. Meanwhile, each LSUN-CHURCH model requires roughly 48, 56, and 103 hours of training with respect to the model's initial latent dimension. In contrast, FASHION-MNIST and MNIST models require approximately 17 hours to be fully trained (per model). Please note that these durations are with respect to the required training iterations for each model and their corresponding dataset detailed in Table~(\ref{tab:models}). Also, note that we utilized the same hardware for our analyses.
\clearpage 

\end{document}